\documentclass{article}
\usepackage{preprint,times}

\usepackage{url}
\usepackage{amsmath,amssymb,amsthm,mathtools,bm}
\usepackage{graphicx,subcaption,booktabs,tabularx,array,wrapfig,float,placeins}
\usepackage{enumitem}
\usepackage[dvipsnames]{xcolor}
\usepackage[most]{tcolorbox}
\usepackage{varwidth}
\newtcolorbox{takeaway}[1][]{
  enhanced jigsaw,breakable,sharp corners,boxrule=1.1pt,
  colback=ForestGreen!10!white,colframe=ForestGreen!80!black,
  coltitle=black,fonttitle=\bfseries,fontupper=\normalfont,
  attach boxed title to top left={yshift=-2mm,xshift=2mm},
  varwidth boxed title=0.86\linewidth,
  boxed title style={colback=ForestGreen!48!white,boxrule=1.1pt},
  title=Takeaway,#1,
}
\usepackage{tikz}
\usetikzlibrary{arrows.meta,positioning,calc,decorations.pathmorphing}
\definecolor{paperblue}{RGB}{31,78,121}
\usepackage{hyperref}
\usepackage{etoc}
\hypersetup{hidelinks}
\usepackage[nameinlink,noabbrev]{cleveref}
\numberwithin{equation}{section}

\crefname{section}{Sec.}{Secs.}\Crefname{section}{Sec.}{Secs.}
\crefname{subsection}{Sec.}{Secs.}\Crefname{subsection}{Sec.}{Secs.}
\crefname{subsubsection}{Sec.}{Secs.}\Crefname{subsubsection}{Sec.}{Secs.}
\crefname{appendix}{App.}{Apps.}\Crefname{appendix}{App.}{Apps.}
\crefname{subappendix}{App.}{Apps.}\Crefname{subappendix}{App.}{Apps.}
\crefname{subsubappendix}{App.}{Apps.}\Crefname{subsubappendix}{App.}{Apps.}
\crefname{figure}{Fig.}{Figs.}\Crefname{figure}{Fig.}{Figs.}
\crefname{table}{Tab.}{Tabs.}\Crefname{table}{Tab.}{Tabs.}
\crefname{equation}{Eq.}{Eqs.}\Crefname{equation}{Eq.}{Eqs.}

\theoremstyle{definition}
\newtheorem{definitionstatement}{Definition}[section]
\newtheorem{assumptionstatement}[definitionstatement]{Assumption}
\theoremstyle{plain}
\newtheorem{propositionstatement}[definitionstatement]{Proposition}
\newtheorem{lemmastatement}[definitionstatement]{Lemma}
\newtheorem{corollarystatement}[definitionstatement]{Corollary}
\newtheorem{theoremstatement}[definitionstatement]{Theorem}
\theoremstyle{remark}
\newtheorem{remarkstatement}[definitionstatement]{Remark}
\newenvironment{definition}[2]{\begin{definitionstatement}[#1]\label[definition]{def:#2}}{\end{definitionstatement}}
\newenvironment{assumption}[2]{\begin{assumptionstatement}[#1]\label[assumption]{ass:#2}}{\end{assumptionstatement}}
\newenvironment{proposition}[2]{\begin{propositionstatement}[#1]\label[proposition]{prop:#2}}{\end{propositionstatement}}
\newenvironment{lemma}[2]{\begin{lemmastatement}[#1]\label[lemma]{lem:#2}}{\end{lemmastatement}}
\newenvironment{theorem}[2]{\begin{theoremstatement}[#1]\label[theorem]{thm:#2}}{\end{theoremstatement}}
\newenvironment{corollary}[2]{\begin{corollarystatement}[#1]\label[corollary]{cor:#2}}{\end{corollarystatement}}

\crefname{definition}{definition}{definitions}\Crefname{definition}{Definition}{Definitions}
\crefname{assumption}{assumption}{assumptions}\Crefname{assumption}{Assumption}{Assumptions}
\crefname{proposition}{proposition}{propositions}\Crefname{proposition}{Proposition}{Propositions}
\crefname{lemma}{lemma}{lemmas}\Crefname{lemma}{Lemma}{Lemmas}
\crefname{theorem}{theorem}{theorems}\Crefname{theorem}{Theorem}{Theorems}
\crefname{corollary}{corollary}{corollaries}\Crefname{corollary}{Corollary}{Corollaries}
\crefname{remark}{remark}{remarks}\Crefname{remark}{Remark}{Remarks}

\providecommand{\E}{\mathbb{E}}

\newcommand{\norm}[1]{\left\lVert #1 \right\rVert}
\newcommand{\ip}[2]{\left\langle #1,#2\right\rangle}
\newcommand{\dd}{\mathrm{d}}
\newcommand{\tw}{\tau_w}
\newcommand{\lr}{\eta}
\newcommand{\lam}{\lambda}
\newcommand{\eps}{\varepsilon}
\newcommand{\one}{\mathbf{1}}
\newcommand{\dphi}{\Delta\phi}
\newcommand{\qtot}{{q^{\mathrm{tot}}}}
\newcommand{\qperp}{{q^{\perp}}}
\newcommand{\qpar}{{q^{\parallel}}}
\newcommand{\qbarR}{q_{\bar R}}
\newcommand{\bx}{\bm{x}}
\newcommand{\bg}{\bm{g}}
\newcommand{\bv}{\bm{v}}
\newcommand{\bxi}{\bm{\xi}}
\newcommand{\bI}{\bm{I}}

\newcommand{\bmom}{\bm{m}}
\newcommand{\bu}{\bm{u}}
\newcommand{\xhat}{\hat{\bx}}
\newcommand{\bzero}{\bm{0}}
\newcommand{\ba}{\bm{a}}
\newcommand{\bz}{\bm{z}}
\newcommand{\bw}{\bm{w}}
\newcommand{\bTheta}{\bm{\Theta}}
\newcommand{\gtilde}{\tilde{\bm g}}
\newcommand{\Sph}{\mathbb{S}}

\newcommand{\thetacap}{\theta_{\mathrm{cap}}}

\title{How Multi-Timescale Memory Cools \mbox{Optimization}: Theory and Evidence from \mbox{\textsc{AdEMAMix}}}

\author{Jacopo Graldi\textsuperscript{1}\thanks{Correspondence to \href{mailto:jacopo.graldi@uzh.ch}{\texttt{jacopo.graldi@uzh.ch}}.}\hspace{.45em}\thanks{This work was done in part while the authors were visiting the Simons Institute for the Theory of \mbox{Computing, UC Berkeley.}}\quad
  Antonio Orvieto\textsuperscript{2}\quad
  Anastasia Koloskova\textsuperscript{1}\footnotemark[2]\\
  \normalfont\textsuperscript{1}\,Department of Mathematical Modeling and Machine Learning, University of Zurich\\
  \normalfont\textsuperscript{2}\,ELLIS Institute T\"ubingen, MPI for Intelligent Systems, T\"ubingen AI Center}
\hypersetup{pdfauthor={Jacopo Graldi, Antonio Orvieto, Anastasia Koloskova}}
\begin{document}

\maketitle
\etocdepthtag.toc{main}
\vspace{-1.5\baselineskip}
\begin{abstract}
Augmenting \textsc{AdamW} with multiple momentum timescales improves large language model pretraining, yet the mechanism driving these gains remains poorly understood. We identify an interaction between multi-timescale optimizer memory and the geometry of normalized networks: memory changes parameter norms and thereby controls how fast parameters rotate. Starting from exact norm dynamics under decoupled weight decay, we show that the parameter norm depends on the temporal correlation of the updates, with weight decay controlling how long these correlations influence optimization. We show how an independently weighted slow momentum can substantially increase the norm of parameters, while contributing negligible power to the instantaneous update, therefore acting as an \textit{endogenous} annealing schedule. We study this mechanism in \textsc{AdEMAMix} \citep{ademamix2025} for language-model pretraining. Controlled interventions show that this scheduling mechanism has clear practical consequences and is an important factor behind \textsc{AdEMAMix}'s advantage over \textsc{AdamW}.
\end{abstract}

\enlargethispage{\baselineskip}

\section{Introduction}
\label{sec:introduction}

The rapid scaling of large language models (LLMs) has increasingly relied on practical optimizer heuristics, while a principled understanding of the underlying learning mechanics \citep{simon2026learningmechanics} remains an open challenge.
Multi-timescale optimizer memory is a recent example.
While \textsc{AdamW} \citep{loshchilovHutter2019adamw} has long served as the standard optimizer, recent work demonstrates that augmenting it with multiple memory timescales yields significant, consistent gains across large-scale pretraining regimes \citep{ademamix2025, semenov2025benchmark,hernandezCano2026apertus,ferbach2026adana,everettQiu2026}.
Most notably, \textsc{AdEMAMix} \citep{ademamix2025} equips \textsc{AdamW} with a second, much slower independently weighted exponential moving average (EMA) of past gradients whose timescale and strength evolve over training.
Despite recent work relating \textsc{AdEMAMix} to accelerated stochastic methods \citep{morwani2025connections, ferbach2026adana}, these perspectives do not explain how its multi-timescale memory alters the training dynamics of modern neural networks. Why such a slow momentum channel consistently benefits large-scale pretraining is therefore an open puzzle.

\begin{figure}[t]
\centering
\begin{subfigure}[t]{.32\linewidth}\includegraphics[width=\linewidth]{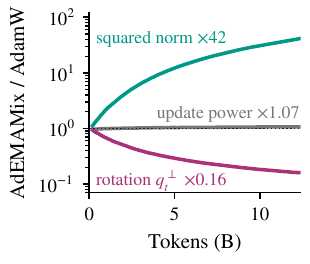}\caption{}\end{subfigure}\hfill
\begin{subfigure}[t]{.32\linewidth}\includegraphics[width=\linewidth]{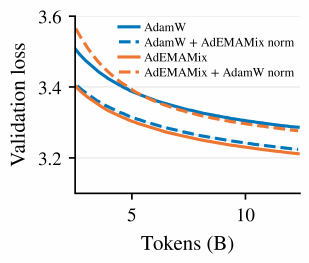}\caption{}\end{subfigure}\hfill
\begin{subfigure}[t]{.32\linewidth}\includegraphics[width=\linewidth]{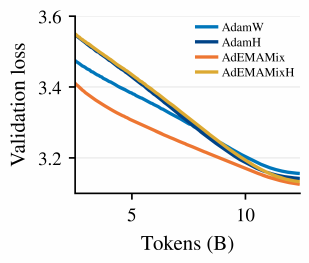}\caption{}\end{subfigure}
\caption{Same learning rate and weight decay, different memory, in 124M Llama pretraining. (a) \textsc{AdEMAMix}/\textsc{AdamW} ratios at the same constant LR: update power stays within $10\%$, while the squared parameter norm grows $42\times$ and tangential rotation slows about sixfold. (b) Validation loss for the native and transplanted runs at their selected constant rates. (c) Validation loss under cosine LR with a free radius and with \textsc{Hyperball} extension (\textsc{AdamH}, \textsc{AdEMAMixH}): fixing the radius helps \textsc{AdamW} but hurts \textsc{AdEMAMix}, narrowing its advantage over AdamH. In (b,c), curves and bands show the mean and one sample standard deviation over three seeds.}
\label{fig:teaser}
\end{figure}

In this work, we show that a major driver of \textsc{AdEMAMix}'s success is an indirect, geometric mechanism: \emph{the interaction between optimizer memory and parameter scale invariance}. Modern language models rely heavily on normalization layers, making parameter groups exactly or approximately scale-invariant: the loss depends only on parameter directions rather than also on their norms \citep{vanLaarhoven2017,liArora2020}. However, the norm remains dynamically critical: it interacts with the learning rate (LR) and with optimizer updates to regulate the parameters' rotation speed during training \citep{hoffer2018norm,wan2021smd,kosson2024rotational}.
We study how an optimizer's memory configuration governs this radial norm trajectory and, in turn, endogenously regulates the parameter rotational dynamics:

\begin{center}
\emph{How does multi-timescale optimizer memory shape the parameter norm, and how does this transform learning-rate schedules into endogenous rotational schedules?}
\end{center}

Prior work establishes that momentum can alter scale-invariant norms and rotational dynamics \citep{heo2021adamp,wan2021smd,kosson2024rotational}.
Our focus is on how the \emph{temporal structure and timescales} of optimizer memory determine this norm trajectory. The key mechanism is the separation between what an optimizer momentum buffer contributes to the update applied now versus what its correlated history leaves behind in the parameter norm. Starting from the exact norm dynamics under decoupled weight decay (WD), we show temporal correlations of the update govern the norm at equilibrium, with WD weighting correlations between updates at different steps.
This separates \textit{instantaneous update power} from the contribution of \textit{temporal correlation} to the norm response: a memory channel can contribute little instantaneous power while its correlated history exerts a much larger accumulated effect on the norm, thereby dampening future rotation. Multi-timescale memory strongly amplifies this separation. In particular, an independently weighted slow channel can make this correlation contribution large without noticeably increasing instantaneous update power, accumulating a disproportionate norm contribution that anneals rotational motion. \textsc{AdEMAMix} provides a clear instance of this mechanism: in our language-model runs, its slow, independently weighted, and scheduled momentum branch contributes only a few percent of instantaneous update power, yet its correlated history has a disproportionately persistent effect. The parameter norm in scale-invariant groups becomes roughly six times larger than \textsc{AdamW}'s at the same LR and WD, while rotational motion is reduced by about the same factor (\Cref{fig:teaser}a). The norm trajectory induced by optimizer memory is therefore not merely a geometric side effect: it transforms the LR schedule into an optimizer-specific rotational schedule, making optimizer memory an integral part of the scheduling problem itself.

To quantify the optimization consequences of this transformation, we evaluate a finite-horizon last-iterate bound on the resulting rotational schedules \citep{graldi2026rotational}, building on the observation of \citet{schaipp2025surprising} that convex last-iterate bounds capture several empirical properties of LR scheduling.
Holding the LR schedule fixed (e.g., cosine \citep{loshchilovHutter2017} or a warmup-stable-decay (WSD) schedule \citep{hu2024minicpm,haegele2024scaling}), changing optimizer memory can materially alter the resulting rotational schedule and the value of the bound. We compare the rotational profiles predicted by our memory model, and find that the bound is smaller for the cooling profile generated by slow memory, predicting a useful scheduling contribution to \textsc{AdEMAMix}'s performance.

We next ask whether this channel contributes to \textsc{AdEMAMix}'s advantage in LLM pretraining. Recent work shows that matching effective or angular step sizes over training can reproduce aspects of the loss trajectories of different optimizers or hyperparameter settings \citep{liu2026elr,yuan2026hypertransfer,xiao2026hyperballfreelunch} and that this can be advantageous in terms of performance \citep{wen2026hyperball,haegele2026md,huebler2026angularmuown}. This motivates intervening on the norms that govern those step sizes.
Removing the radial component of the update suppresses \textsc{AdEMAMix}'s norm growth. Giving \textsc{AdamW} the \textsc{AdEMAMix} norm trajectory recovers most of the loss gap and shifts its preferred LR. The reverse transplant, instead, largely removes \textsc{AdEMAMix}'s advantage despite full LR retuning (\Cref{fig:teaser}b). Applying \textsc{Hyperball} \citep{wen2026hyperball} improves \textsc{AdamW} and worsens \textsc{AdEMAMix}, leaving about a third of the original gap under cosine LR (\Cref{fig:teaser}c). When we switch off slow memory mid-training, its effect on relative updates weakens as the excess norm decays.
Together, these interventions indicate that norm-induced rotational scheduling accounts for a large part of \textsc{AdEMAMix}'s advantage over \textsc{AdamW}.

More broadly, our results show that the LR, WD, and optimizer memory cannot be understood independently for scale-invariant parameters: together they generate the rotational schedule actually executed during training. We therefore recognize that this realized rotational schedule, rather than any individual hyperparameter or schedule in isolation, is a natural object for understanding and comparing optimizer dynamics and performance.

\paragraph*{Our Contributions}
Our main contributions are:
\begin{itemize}
    \item \textbf{Norm dynamics from optimizer memory.}
    We develop a framework that expresses radial dynamics through the temporal structure of the update, separating instantaneous power from the accumulated contribution of correlated updates.

    \item \textbf{Rotational slowdown from multi-timescale memory.}
    This decomposition exposes how different memory architectures, and in particular independently weighted multiple timescales, can generate outsized radial effects while contributing little instantaneous power, thereby increasing the parameter norm and annealing future rotational motion.

    \item \textbf{Rotational scheduling and empirical validation.}
    We connect these dynamics to finite-horizon rotational scheduling and show through controlled LLM interventions that the norm channel explains a large part of \textsc{AdEMAMix}'s advantage over \textsc{AdamW}.
\end{itemize}
\Cref{sec:related-work} discusses related work on multi-timescale optimizer memory, scale-invariant geometry, and radial dynamics.

\section{Background: \textsc{AdEMAMix} and Scale-Invariant Dynamics}
\label{sec:background}

\subsection{\textsc{AdEMAMix}}
\label{sec:ademamix}

We consider updates with decoupled weight decay,
\begin{equation}
    \bx_{t+1}=\delta_t\bx_t-\lr_t\bu_t,
    \qquad \delta_t:=1-\lr_t\lam_t\in(0,1],
    \label{eq:param_update}
\end{equation}
where $\bx_t\in\mathbb R^d$, $\lr_t>0$ is the learning rate, $\lam_t\geq0$ is the weight-decay coefficient, and $\bu_t$ is the optimizer update.

\begin{definition}{\textsc{AdEMAMix} Update}{ademamix-update}
Given the current gradient $\bg_t$, \textsc{AdEMAMix} \citep{ademamix2025} combines fast and slow momentum buffers with a shared second-moment scale:
\begin{equation}
\begin{aligned}
    \bmom^{(j)}_{t+1}&=\beta^{(j)}_t\bmom^{(j)}_t+(1-\beta^{(j)}_t)\bg_t,
    \qquad j\in\{f,s\},\\
    \bv_{t+1}&=\beta_2\bv_t+(1-\beta_2)\bg_t^{\,2},
    \qquad
    \bu_t=\frac{\bmom^{(f)}_{t+1}+\alpha_t\bmom^{(s)}_{t+1}}{\sqrt{\bv_{t+1}}+\eps}.
\end{aligned}
    \label{eq:mixed_momentum}
\end{equation}
We have that $0\leq\beta^{(f)}_t=\beta^{(f)},\beta^{(s)}_t,\beta_2<1$, $\alpha_t\geq0$, and $\eps>0$. \textsc{AdamW} \citep{loshchilovHutter2019adamw} is the case $\alpha_t=0$. $\alpha_t$ and $\beta^{(s)}_t$ are warmed up through training (\Cref{app:llm-protocol}).
For simplicity we ignore bias correction.
\end{definition}

\subsection{Scale Invariance, Norms, and Rotational Motion}
\label{sec:si-geometry}

A nonzero parameter group is scale-invariant if $L(\kappa\bx)=L(\bx)$ for $\kappa>0$, so the loss depends only on its unit direction $\xhat:=\bx/\norm{\bx}$. Differentiating this identity in $\kappa$ and in $\bx$, respectively, gives
\begin{equation*}
    \ip{\nabla L(\bx)}{\bx}=0,
    \qquad
    \nabla L(\kappa\bx)=\kappa^{-1}\nabla L(\bx).
\end{equation*}
Thus the gradient is tangent to the weight sphere and scales inversely with radius \citep{kosson2024rotational,arora2019}. Equivalently, $\nabla L(\bx)=\norm{\bx}^{-1}\nabla_{\mathbb S}L(\xhat)$, where $\nabla_{\mathbb S}L(\xhat):=(\bI-\xhat\xhat^\top)\nabla L(\xhat)$ is the spherical gradient; \Cref{app:geometry} gives the derivations.

\noindent\begin{minipage}[t]{0.66\linewidth}
\vspace{0pt}
Directional motion depends on the update's radial and tangential components, $u_{t,\parallel}:=\ip{\bu_t}{\xhat_t}$ and $\bu_{t,\perp}:=\bu_t-u_{t,\parallel}\xhat_t$. Dividing their displacements by the radius gives
\begin{equation*}
    \qpar_t:=\frac{\lr_tu_{t,\parallel}}{\norm{\bx_t}},
    \qquad
    \qperp_t:=\frac{\lr_t\norm{\bu_{t,\perp}}}{\norm{\bx_t}}.
\end{equation*}
The total relative update is $\qtot_t:=\lr_t\norm{\bu_t}/\norm{\bx_t}$.
The turning angle $\dphi_t:=\angle(\bx_t,\bx_{t+1})$ satisfies
$\dphi_t=\arctan\bigl(\qperp_t/(\delta_t-\qpar_t)\bigr)$ when $\delta_t-\qpar_t>0$.
For small relative updates and small weight decay, we can approximate $\qperp_t\simeq\dphi_t$ \citep{kosson2024rotational}; \Cref{app:geometry-rotation} gives the exact relation and approximation.
\end{minipage}\hfill
\begin{minipage}[t]{0.31\linewidth}
  \vspace{0pt}
  \centering
  \includegraphics[width=\linewidth]{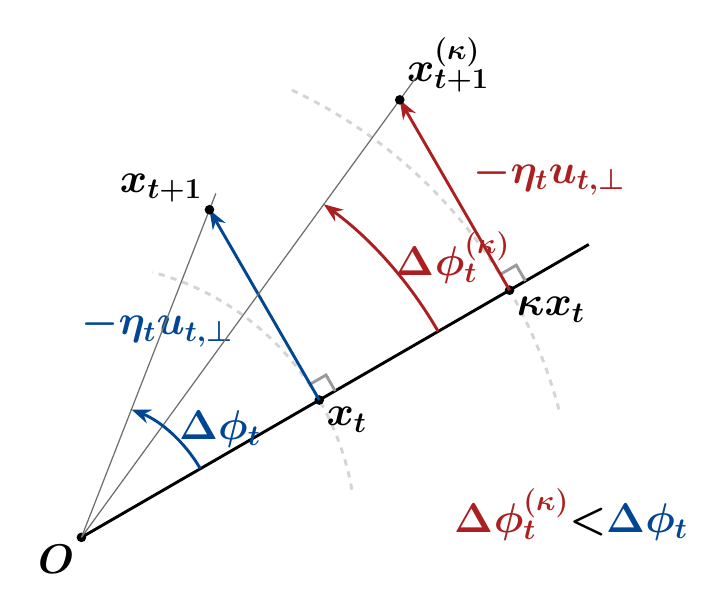}
  \captionsetup{hypcap=false}
  \captionof{figure}{Equal tangential updates produce smaller rotations at larger radii.}
  \label{fig:two-rate-geometry-main}
\end{minipage}

\begin{takeaway}
For a fixed tangential displacement, a larger parameter norm gives a smaller rotation.
\end{takeaway}

\section{The Empirical Puzzle}
\label{sec:puzzle}

\textsc{AdEMAMix} generates roughly sixfold larger parameter norms and fivefold slower rotation than \textsc{AdamW}, although the slow buffer's own contribution to instantaneous update power is negligible. We observe this contrast in 124M-parameter Llama pretraining on FineWeb, with the same learning-rate schedule, peak rate $8\cdot10^{-4}$, and weight decay for both optimizers.

\Cref{fig:teaser}a shows the norm increase under a constant learning rate; \Cref{fig:puzzle} in \Cref{app:llm-measurements} shows the norm and rotation trajectories under both constant and cosine rates. Here and below, the matrix norm is the aggregate $\norm{\bx_t}:=(\sum_m\norm{\bx_{m,t}}_F^2)^{1/2}$ over the hidden matrices that are weight-decayed, and specified in \Cref{app:llm-protocol}. The larger radius accompanies much smaller tangential relative updates.

To compare this effect with instantaneous power, let $\bu_t^{(f)}$ and $\bu_t^{(s)}$ denote the preconditioned fast and slow buffers. Then $\bu_t=\bu_t^{(f)}+\alpha_t\bu_t^{(s)}$ and
\begin{equation}
  \norm{\bu_t}^2=\norm{\bu_t^{(f)}}^2
  +2\alpha_t\ip{\bu_t^{(f)}}{\bu_t^{(s)}}
  +\alpha_t^2\norm{\bu_t^{(s)}}^2.
  \label{eq:present-power-split}
\end{equation}
Over the final $30\%$ of training, the slow-buffer power and its overlap with the fast buffer together account for only about $5\%$ of total power under constant LR and $6\%$ under cosine (\Cref{tab:survived-shares}). \textit{How can this small instantaneous contribution accompany such a large norm increase and rotational slowdown?}

\section{Memory-Induced Norm Cooling}
\label{sec:mechanism}

\subsection{Temporal Correlations Control the Parameter Norm}
\label{sec:correlation-law}

To gain a better understanding of \textsc{AdEMAMix}, we start by expanding the parameter norm recursion.

\begin{lemma}{Squared-Norm Growth Identity}{norm-growth}
Every update in \eqref{eq:param_update} satisfies
\begin{equation}
    \norm{\bx_{t+1}}^2
    =\delta_t^2\norm{\bx_t}^2
    -2\delta_t\lr_t\ip{\bx_t}{\bu_t}
    +\lr_t^2\norm{\bu_t}^2.
    \label{eq:nqm-norm-growth-compact}
\end{equation}
\end{lemma}
The first term contracts the previous squared radius thanks to WD. The second is a signed radial contribution: alignment of $\bu_t$ against $\bx_t$ drives outward motion. The last term is instantaneous displacement power. Without decay, and for heavy-ball momentum on tangent gradients, the radial-work term equals the accumulated past-update power through which momentum grows scale-invariant norms in \citet[Lemma~2.2]{heo2021adamp}.

Earlier updates enter the parameter with a minus sign. When a later update aligns with one of them, its applied displacement reinforces what remains of the earlier displacement. Positive correlations across the update history can therefore make $\ip{\bx_t}{\bu_t}$ negative, producing outward radial work on average. Weight decay contracts those earlier displacements at every intervening step. It consequently determines how long correlations with past updates can influence the current norm, as captured by the weighted sum below.

\begin{definition}{Update Correlations and Their Norm Contribution}{update-correlation}
For updates $(\bu_t)_{t\in\mathbb Z}$ with finite second moments, constant mean, and
$C_u(h):=\E\ip{\bu_t}{\bu_{t-h}}$ independent of $t$, define, for $0<\delta<1$,
\begin{equation}
    \mathcal C_\delta[\bu]:=C_u(0)+2\sum_{h\geq1}\delta^h C_u(h).
    \label{eq:applied-update-correlation-functional}
\end{equation}
This combines instantaneous update power $C_u(0)=\E\norm{\bu_t}^2$ with correlations between updates $h$ steps apart, weighted by $\delta$.
\end{definition}

\begin{theorem}{Stationary Norm and Relative-Update Scale}{equilibrium-norm}
Let the updates satisfy \Cref{def:update-correlation} and the parameter obey \eqref{eq:param_update} for all $t\in\mathbb Z$, with constant $\lr>0$ and $0<\delta<1$. Assume radius stationarity: $\E\norm{\bx_t}^2=\bar R^2<\infty$ is independent of $t$. For $C_u(0)>0$, define
$\qbarR:=\lr\sqrt{\E\norm{\bu_t}^2}/\bar R$. Then
\begin{equation}
    \bar R^2=\frac{\lr^2\mathcal C_\delta[\bu]}{1-\delta^2},
    \qquad
    \qbarR^2=(1-\delta^2)\frac{C_u(0)}{\mathcal C_\delta[\bu]}.
    \label{eq:nqm-norm-equilibrium-colored}
\end{equation}
\end{theorem}
Positive correlations between updates $\mathcal C_\delta[\bu]$ can therefore increase $\bar R$ and decrease $\qbarR$ even when $C_u(0)$ changes little. The proof is in \Cref{app:proofs-exact}. Under the isotropic Gaussian-gradient model of \citet{wen2026hyperball}, where the update correlations have closed forms, our radius identity reduces to their equilibrium \citep[Eq.~39]{wen2026hyperball}.

\subsection{Optimizer Memory Generates Temporal Correlation}
\label{sec:memory-correlation}
The norm identity relates the radius to correlations of the updates applied to the parameters. We next ask how optimizer memory generates these correlations. A momentum buffer reuses each input across several steps, so its outputs can be correlated even when the elements entering its buffer are uncorrelated. We represent this reuse by writing the modeled update as a weighted sum of past inputs:
\begin{equation}
 \widehat{\bu}_t=\sum_{k\ge0}e_k\ba_{t-k}.
 \label{eq:memory-model-main}
\end{equation}
The coefficient $e_k$ weights an input $k$ steps old. This form uses constant coefficients and an infinite history. The following choices recover familiar optimizers:
\begin{center}
\small
\renewcommand{\arraystretch}{1.15}
\begin{tabular}{@{}lll@{}}
\toprule
Optimizer & Input $\ba_t$ & Weights $e_k$ \\
\midrule
Gradient descent / SGD & Full / stochastic gradient & $e_k^{\mathrm{GD}}=e_k^{\mathrm{SGD}}=\one_{k=0}$ \\
Momentum SGD & Stochastic gradient & $e_k^{\mathrm{SGDm}}=(1-\beta)\beta^k$ \\
\textsc{AdamW} & $\bg_t/(\sqrt{\bv_{t+1}}+\eps)$ & $e_k^{\mathrm{AdamW}}=(1-\beta^{(f)})(\beta^{(f)})^k$ \\
\textsc{AdEMAMix} & $\bg_t/(\sqrt{\bv_{t+1}}+\eps)$ & $e_k^{\mathrm{AMix}}=e_k^{\mathrm{AdamW}}+\alpha(1-\beta^{(s)})(\beta^{(s)})^k$ \\
\bottomrule
\end{tabular}
\end{center}
For the raw-gradient methods, the representation is exact, using the available history for zero-start momentum. For \textsc{AdamW} and \textsc{AdEMAMix}, it fixes each input's divisor at the time that input enters the buffer, whereas the optimizer divides by the current divisor. This approximation is reminiscent of the common slowly-varying preconditioner approximation (e.g. \citep{kingma2014adam}). \Cref{app:llm-preconditioner} checks this approximation against the accumulated updates in our LLM runs.

So far, we have allowed arbitrary update correlations. To study the effect of optimizer memory, we now follow the noise-dominated random-walk analysis of \citet{kosson2024rotational}, which gives useful predictions of equilibrium weight norms and rotation rates in trained networks. We assume zero-mean inputs that are uncorrelated across steps, so that correlations in the modeled updates arise from the optimizer's reuse of past inputs.

\begin{assumption}{Zero-Mean, Uncorrelated Inputs}{white-noise}
The inputs satisfy $\E\ba_t=\bzero$, $\E\norm{\ba_t}^2=C_a(0)<\infty$ for every $t$, and $\E\ip{\ba_t}{\ba_r}=0$ for $t\ne r$.
\end{assumption}

\begin{definition}{Power and Update Correlations}{survival-weighted-forcing}
For a real, absolutely summable kernel $e$, define
\begin{equation}
 \Gamma_e(h):=\sum_{n\ge0}e_ne_{n+h},\qquad
 P[e]:=\Gamma_e(0),\qquad
 F_\delta[e]:=P[e]+2\sum_{h\ge1}\delta^h\Gamma_e(h),
 \label{eq:kernel-power-forcing}
\end{equation}
with $0\leq\delta\leq1$. Here $P[e]$ measures current power; $F_\delta[e]$ also includes the temporal correlations weighted by weight decay.
\end{definition}
Under \Cref{ass:white-noise}, $C_{\widehat u}(h)=C_a(0)\Gamma_e(h)$, so modeled update power is $C_a(0)P[e]$ and its norm contribution is $\mathcal C_\delta[\widehat{\bu}]=C_a(0)F_\delta[e]$.

\begin{proposition}{Stationary Norm Under Uncorrelated Inputs}{white-equilibrium}
Under \Cref{ass:white-noise}, let $e$ be absolutely summable and let \eqref{eq:param_update} be driven by $\widehat{\bu}_t$. Assume radius stationarity, $\E\norm{\bx_t}^2=\bar R^2<\infty$ independent of $t$, with constant $\lr>0$ and $0<\delta<1$. Then
\begin{equation}
    \bar R^2=\frac{\lr^2 C_a(0)F_\delta[e]}{1-\delta^2},
    \qquad
    \qbarR^2=(1-\delta^2)\frac{P[e]}{F_\delta[e]}
    \quad\bigl(C_a(0)P[e]>0\bigr).
    \label{eq:white-radius-ratio-main}
\end{equation}
\end{proposition}
The proof is in \Cref{app:proofs-represented}; \Cref{app:proofs-nonzero-mean} gives the corresponding formulas for a constant nonzero input mean.

Radius comparisons require specifying how input power changes between optimizers. For \textsc{AdamW} and \textsc{AdEMAMix}, the normalization of preconditioning cancels a common gradient rescaling when the divisor scales with it, motivating approximately radius-independent input power \citep{kosson2024rotational}. The live-preconditioned multilayer perceptron (MLP) in \Cref{app:controlled-mlp} illustrates the normalized-input model's norm responses.

\begin{corollary}{One Unit-Gain EMA}{single-buffer-cancellation}
For $e_k^{\mathrm{AdamW}}=(1-\beta)\beta^k$, $0\leq\beta<1$, and $0\leq\delta\leq1$,
\begin{equation}
 P[e^{\mathrm{AdamW}}]=\frac{1-\beta}{1+\beta},\quad
 F_\delta[e^{\mathrm{AdamW}}]=\frac{1-\beta}{1+\beta}\frac{1+\beta\delta}{1-\beta\delta},\quad
 \frac{P[e^{\mathrm{AdamW}}]}{F_\delta[e^{\mathrm{AdamW}}]}=\frac{1-\beta\delta}{1+\beta\delta}.
 \label{eq:nqm-norm-single-buffer-forcing}
\end{equation}
For $0<\delta<1$, increasing $\beta$ decreases both $P[e^{\mathrm{AdamW}}]$ and $F_\delta[e^{\mathrm{AdamW}}]$, with $\partial_\beta\log P<\partial_\beta\log F_\delta<0$, i.e. with the former decreasing faster than the latter.
\end{corollary}
At fixed LR, decay, and input power, longer one-buffer memory thus lowers both $\bar R$ and $\qbarR$. As $\delta\to1^-$, $F_\delta[e]$ tends to $(\sum_k e_k)^2$, which equals one for every unit-gain EMA; this limit recovers the total-update-contribution construction of \citet[Secs.~3.1--3.2]{kosson2024rotational}, which assigns each input's full future contribution to the step at which it enters. For short memory relative to weight decay, $1-\delta\ll1-\beta$, $F_\delta\approx1$ recovers their $\beta$-independent \textsc{AdamW} radius. Instead, for longer memory, $1-\delta\approx1-\beta$, $F_\delta[e]$ lies below $(\sum_k e_k)^2$ for every nonnegative $e$. At $\delta=1$ and matched update power, $F_1/P=(1+\beta)/(1-\beta)$ is the asymptotic norm-growth factor of momentum without decay \citep[Cor.~2.3]{heo2021adamp}; decay tempers it to $(1+\beta\delta)/(1-\beta\delta)$. Derivatives and the gain calculation are in \Cref{app:proofs-single}.

\subsection{The Two-Timescale \textsc{AdEMAMix} Channel}
\label{sec:lever}

The weights $e^{\mathrm{AMix}}$ add slow memory while preserving the fast buffer. With a slower second buffer, increasing $\alpha$ can raise $F_\delta[e^{\mathrm{AMix}}]$ proportionally faster than $P[e^{\mathrm{AMix}}]$.

\begin{theorem}{Two-Channel Power and Temporal Correlation}{ademamix-lever}
Fix $0<\delta<1$, $0\leq\beta^{(f)}<\beta^{(s)}<1$, and $\alpha\geq0$. Both $P[e^{\mathrm{AMix}}]$ and $F_\delta[e^{\mathrm{AMix}}]$ increase strictly with $\alpha$, while
\begin{equation}
    \frac{\dd}{\dd\alpha}\frac{P[e^{\mathrm{AMix}}]}{F_\delta[e^{\mathrm{AMix}}]}<0.
    \label{eq:nqm-lever-derivative}
\end{equation}
\end{theorem}
The exact formulas and proof are in \Cref{app:proofs-two}. Under \Cref{prop:white-equilibrium}, at matched input power, LR, and decay, \textit{the modeled radius increases while the relative-update scale decreases}.
These conclusions also hold with a fixed nonzero input mean and temporally uncorrelated residuals of fixed positive variance (\Cref{cor:two-channel-nonzero-mean}).

\paragraph{Example: Little Power, Substantial Cooling.}
For finite fixed $\alpha$, consider $1-\delta\ll1-\beta^{(s)}\ll1-\beta^{(f)}$: slow memory is much longer than fast memory, but both are short relative to weight decay. The pairwise formulas give
\begin{equation*}
 P[e^{\mathrm{AMix}}]\approx P[e^{\mathrm{AdamW}}]
       +\left(2\alpha+\frac{\alpha^2}{2}\right)(1-\beta^{(s)}),
 \qquad F_\delta[e^{\mathrm{AMix}}]\approx(1+\alpha)^2.
\end{equation*}
The slow buffer spreads its power across many steps, while weak decay preserves its temporal correlations. If $(2\alpha+\alpha^2/2)(1-\beta^{(s)})\ll P[e^{\mathrm{AdamW}}]$, current power barely changes, yet $\bar R$ grows by approximately $1+\alpha$ and $\qbarR$ falls by the same factor.
In the LLM recordings, the slow branch's own term contributes only a few percent of instantaneous update power, yet roughly four fifths of the squared accumulated update displacement after weight decay (\Cref{tab:survived-shares}).

The effect of slow memory is non-monotonic. If $\alpha$ is fixed and the slow memory grows much longer than the weight-decay timescale, $\beta^{(s)}\to1^-$, then $P[e^{\mathrm{AMix}}]/F_\delta[e^{\mathrm{AMix}}]$ returns to its \textsc{AdamW} value: weight decay shrinks old updates before the slow buffer reuses them (\Cref{app:proofs-limits}). Our LLM runs are near this regime: weight decay removes about half of the norm increase that slow memory would otherwise give.

\begin{figure}[t]
\centering
\begin{subfigure}[t]{.3\linewidth}\includegraphics[width=\linewidth]{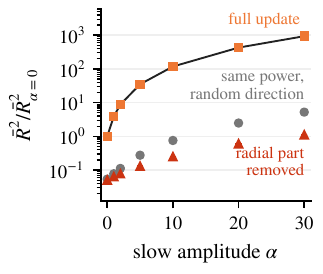}\caption{}\end{subfigure}\hfill
\begin{subfigure}[t]{.3\linewidth}\includegraphics[width=\linewidth]{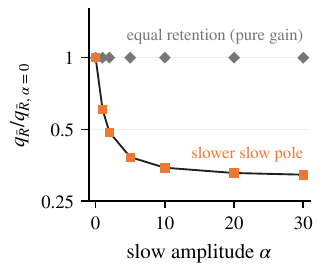}\caption{}\end{subfigure}\hfill
\begin{subfigure}[t]{.3\linewidth}\includegraphics[width=\linewidth]{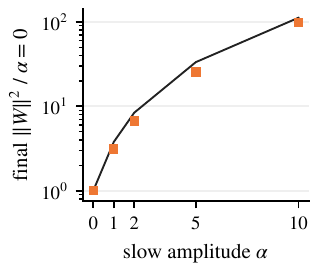}\caption{}\end{subfigure}
\caption{The norm-cooling predictions in controlled settings; markers are measurements, black lines the model, and all quantities are relative to $\alpha=0$. (a) Tangent-noise random walk with $\beta^{(s)}=0.99$: squared radius for the full update, for the same update power with randomized directions, and for the update with its radial part removed. (b) Relative-update scale $\qbarR$ for the same sweep, and for an equal-retention control $\beta^{(s)}=\beta^{(f)}$ that changes only gain. (c) Final squared norm of the scale-invariant MLP hidden matrices (geometric mean over matrices, five seeds) against the finite-history model; the model's scale is calibrated at $\alpha=0$ and cancels in the ratio.}
\label{fig:sphere}
\end{figure}

\Cref{fig:sphere} simulates these predictions. In a tangent-noise random walk, Gaussian inputs are projected onto the current weight sphere, and the parameter radius evolves under momentum and decay (\Cref{app:controlled-sphere}). Raising $\alpha$ grows the radius as predicted, while the same update power with randomized directions, or the update without its radial part, grows it far less (\Cref{fig:sphere}a). In \Cref{fig:sphere}b, $\qbarR$ falls with $\alpha$. When $\beta^{(s)}=\beta^{(f)}$, adding the second buffer simply multiplies the update by $1+\alpha$, leaving $\qbarR$ unchanged. In Llama pretraining, projecting out the radial component of the \textsc{AdEMAMix} update also suppresses most of the norm growth (\Cref{app:llm-interventions}).

\begin{takeaway}
In the stationary uncorrelated-input model, slow memory can raise the parameter norm substantially with little added update power at matched input power, learning rate, and decay. The larger norm reduces rotational motion.
\end{takeaway}

\subsection{Finite-Time Radius and Rotational Schedules}
\label{sec:forward}

For scheduled training, the finite model gives a radius trajectory and a time-varying update-power factor. Write $\widehat{\bu}_t=\sum_{i=0}^tE_{t,i}\ba_i$, where $E_{t,i}$ is the weight assigned to input $i$ at step $t$. Under \Cref{ass:white-noise} with fixed input power, let $\bx_t^{\mathrm{rep}}$ follow \eqref{eq:param_update} driven by $\widehat{\bu}_t$, and define $R_t:=\sqrt{\E\norm{\bx_t^{\mathrm{rep}}}^2}$ and $P_t:=\sum_{i=0}^tE_{t,i}^2$. The radius accumulates past updates, while $P_t$ sets current power.

To isolate this scalar scheduling effect, supply $R_t$ to a fresh tangent proposal scaled by $\sqrt{P_t}$. Its rotational coefficient is
\begin{equation}
 s_t=\frac{\lr_t\sqrt{P_t}}{\delta_tR_t}.
 \label{eq:transferred-schedule}
\end{equation}
Thus radius growth can lower the coefficient even at constant LR. The finite radius recursion and proposal derivation are in \Cref{app:scheduling-forward}.

The modeled coefficients capture the overall cooling trend in the measured LLM angles (\Cref{fig:generated} in \Cref{app:scheduling-curve-settings}). \Cref{sec:llm-interventions} tests the same model against a controlled intervention.

\subsection{Endogenous Rotational Scheduling}
\label{sec:scheduling}
\label{sec:when-help}
\noindent\begin{minipage}[t]{.6\linewidth}

Slow memory cools rotation even at constant LR. To assess the resulting schedule, we use the last-iterate bound of \citet{graldi2026rotational} as a surrogate criterion for the modeled rotational profiles. It separates the overall scale of the rotational coefficients from their distribution over time and gives useful predictions of schedule performance in LLM pretraining \citep[Sec.~4]{graldi2026rotational}.

For the $N$ coefficients in the evaluation window, write $\omega_t:=s_t/\sum_{j=1}^N s_j$. The schedule-dependent term in the bound is
\begin{equation}
 C_N(\omega)=\omega_N+
 \sum_{t=1}^{N-1}\frac{\omega_t^2}{\sum_{j=t+1}^N\omega_j}.
 \label{eq:scheduling-bound-main}
\end{equation}

Each term in the sum divides a squared coefficient by the sum of the coefficients that follow it. A large step is therefore more costly when little motion remains later in training. After optimizing the overall scale, the bound is proportional to $\sqrt{C_N(\omega)}$ \citep[Prop.~2.4]{graldi2026rotational}. \Cref{app:scheduling-criterion} gives the full bound and its assumptions.

At the same constant LR, the modeled \textsc{AdEMAMix} schedule keeps cooling as the \textsc{AdamW} profile flattens (\Cref{fig:scheduling-application}a). The bound ends lower for \textsc{AdEMAMix} (panel b), consistent with its advantage in LLM runs with constant LR (\Cref{fig:teaser}b and \Cref{fig:schedule-benefit}).
\end{minipage}\hfill
\begin{minipage}[t]{.37\linewidth}
\vspace{-1.2cm}
\captionsetup{type=figure}
\centering
\includegraphics[width=.8\linewidth]{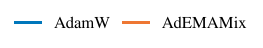}\par
\begin{subfigure}[t]{.9\linewidth}
\includegraphics[width=\linewidth]{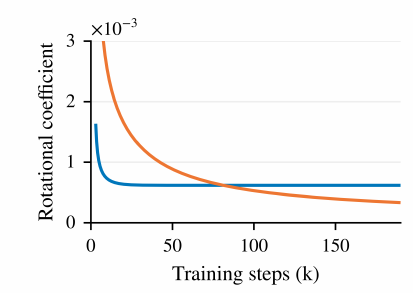}\caption{}
\end{subfigure}\par
\begin{subfigure}[t]{.9\linewidth}
\includegraphics[width=\linewidth]{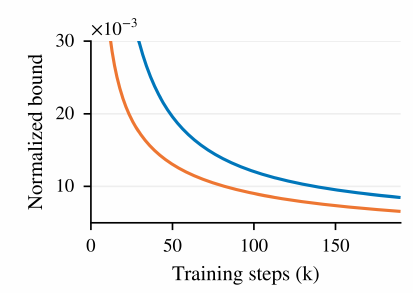}\caption{}
\end{subfigure}
\captionsetup{skip=4pt}
\caption{Schedules at constant LR. (a) Rescaled rotational coefficients. (b) Normalized bounds. Each profile uses its final-horizon optimal scale (\Cref{app:scheduling-application}).}
\label{fig:scheduling-application}
\end{minipage}

\section{Practical Consequences in LLM Pretraining}
\label{sec:llm}
\label{sec:llm-schedule}

How much of \textsc{AdEMAMix}'s advantage comes from the rotational schedule induced by the norms? We test this by intervening with controlled experiments on the norms and switching off the slow memory in 124M-parameter Llama pretraining on FineWeb (\Cref{app:llm-experiments}). Unless indicated otherwise, measurements show means over three seeds, with bands and error bars denoting one sample standard deviation.

\subsection{Intervening on the Norm Trajectory}
\label{sec:llm-operating-point}
\textsc{AdEMAMix} reaches a lower validation loss than \textsc{AdamW} when each optimizer uses its own tuned learning rate, held constant after warmup (\Cref{fig:teaser}b). The native runs differ in both optimizer memory and norm trajectory. In this intervention, we swap their recorded norm trajectories in both directions: \textsc{AdamW} receives the norms from \textsc{AdEMAMix} training, and \textsc{AdEMAMix} receives those from \textsc{AdamW} training. After each step, we rescale each targeted matrix to its donor's norm, retaining the recipient's post-step direction and buffers (\Cref{app:llm-interventions}). We tune the learning rate separately for each configuration (\Cref{fig:lr-ladder}).
We find that with \textsc{AdEMAMix}'s norms, \textsc{AdamW} recovers most of the final-loss gap. With \textsc{AdamW}'s norms, \textsc{AdEMAMix} loses most of its advantage (\Cref{fig:teaser}b).

Next, we ask how much does slow memory help when it cannot build up a larger radius? For this intervention we apply \textsc{Hyperball} \citep{wen2026hyperball} to both optimizers: acting as a wrapper on a base optimizer, it fixes each targeted matrix at its initial radius and normalizes the proposed update, so the prescribed learning rate sets the proposal's size relative to the radius. Each base optimizer still uses its momentum buffers and update rule to determine the update direction. This method, and closely related ones \citep{haegele2026md,huebler2026angularmuown}, have been shown to consistently improve the performance of base optimizers such as Adam and Muon in practice. We therefore expect it to improve \textsc{AdamW} as well in our setup, and we want to observe whether it also improves \textsc{AdEMAMix} when the slow memory cannot build up a larger radius.

Our experiments indeed show that \textsc{Hyperball} lowers \textsc{AdamW}'s final validation loss, but, very interestingly, it \emph{raises} \textsc{AdEMAMix}'s loss (\Cref{fig:teaser}c). With the wrapper, the slow-memory advantage shrinks from about $0.03$ to $0.01$ in validation loss, and the sweep favors much smaller slow amplitudes (\Cref{fig:hyperball-amplitude}).
The two interventions indicate that much of the benefit of slow memory comes from the rotational schedule it produces through norm growth.

\begin{figure}[htbp]
\centering
\includegraphics[width=0.8\linewidth]{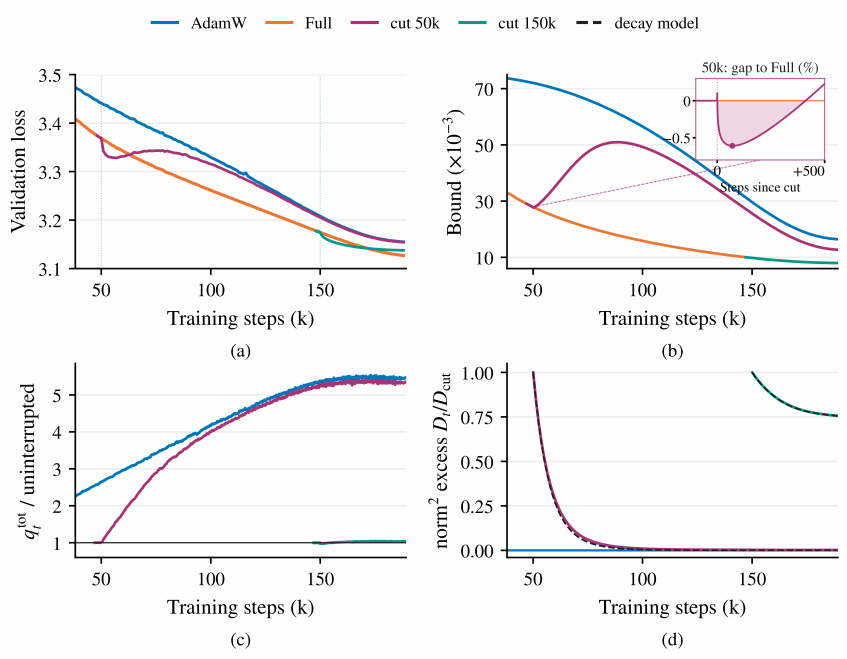}
\caption{Removing slow memory at 50k and 150k updates under cosine LR with peak $8\cdot10^{-4}$ (individual runs). Full denotes uninterrupted \textsc{AdEMAMix}. (a) Validation loss. (b) Normalized bound on modeled rotational coefficients. The inset shows the percentage gap to Full from 100 steps before to 500 steps after the 50k switch. (c) Measured total relative update divided by that of Full. (d) Squared-norm excess over \textsc{AdamW}, normalized at the switch, with weight-decay survival (dashed).}
\label{fig:switch}
\end{figure}

\subsection{Switching Off Slow Memory}
\label{sec:llm-interventions}

Finally, we analyze a peculiar pattern that emerges when we switch off \textsc{AdEMAMix}'s slow memory at a given step during training. Indeed, the validation loss briefly drops and then rebounds, as shown in \Cref{fig:switch}a.
\citet[App.~C.1.3, Fig.~11]{ademamix2025} observed this pattern but left the explanation for this behavior open. In our model, setting $\alpha_t=0$ immediately reduces update power while the weights retain the norm accumulated before the switch. The smaller updates therefore rotate the weights less, much as they would after a learning-rate drop. Once the slow buffer stops supplying correlated updates, weight decay gradually erodes the excess norm, which therefore returns to its baseline (i.e. AdamW) value with a timescale that our model predicts accurately (\Cref{app:scheduling-switch}). As the radius shrinks, the total relative update $\qtot_t$ rises again (\Cref{fig:switch}c), causing the validation loss to rebound.

\Cref{fig:switch} shows this dip and rebound after the 50k switch under cosine LR (panel a). The excess squared norm shrinks, and relative updates rise toward those of \textsc{AdamW} (panels c,d). With the switch at 150k, much less weight decay remains and most of the norm excess survives and relative updates stay close to those of uninterrupted \textsc{AdEMAMix}. Both runs finish at a higher validation loss than uninterrupted training.

After the 50k switch, the bound on the modeled rotational coefficients also dips below its uninterrupted counterpart and then rebounds, echoing the measured loss response (inset in \Cref{fig:switch}b and \Cref{app:scheduling-switch}). \Cref{fig:switch-cosine-all,fig:switch-constant} show all four switch times under cosine and constant LR.

\begin{takeaway}[unbreakable]
Transferring the norm trajectory transfers much of \textsc{AdEMAMix}'s constant-rate advantage, while fixing the parameter radii substantially reduces the benefit of slow memory. After the slow channel is removed, its effect persists through the accumulated norm and weakens as that norm decays.
\end{takeaway}

\section{Conclusion}
\label{sec:discussion}

In this paper we have uncovered how slow momentum can contribute little to the current update while substantially changing the parameter norm through correlations across steps. The role of weight decay, instead, is to control how long these contributions persist. The norm growth arising from these interactions cools rotational motion, creating an endogenous schedule even when the learning rate is explicitly set to a constant value. Our interventions show that this mechanism accounts for a large part of \textsc{AdEMAMix}'s advantage over \textsc{AdamW}.

Our evidence shows that optimizer memory cannot be disentangled from learning-rate scheduling, especially for scale-invariant parameters. More broadly, considering the implicit schedules induced by optimization dynamics can help us understand and compare the performance of different optimizers across hyperparameter settings and learning-rate schedules.

\clearpage

\section*{Acknowledgments}
Antonio Orvieto acknowledges the financial support of the Hector Foundation.

\section*{AI Use Statement}
\label{sec:ai-use-statement}

We used generative AI tools to help develop theoretical models, formulate
mathematical claims, write proofs, implement methods, interpret results,
prepare figures, and polish the manuscript.
We reviewed all AI-assisted work, verified the proofs by hand, and tested the code.

\appendix
\crefalias{section}{appendix}
\crefalias{subsection}{subappendix}
\crefalias{subsubsection}{subsubappendix}
\etocdepthtag.toc{appendix}
\etocsettagdepth{main}{none}
\etocsettagdepth{appendix}{subsection}
\clearpage
\etocsettocstyle{\section*{Appendix Contents}}{}
\tableofcontents
\clearpage
\section{Related Work}
\label{sec:related-work}

\paragraph{Multi-Timescale Optimizer Memory.}
\citet{ademamix2025} bring the idea of multi-timescale optimizer memory \citep{lucas2018aggregated} to large-scale language model pretraining, with \textsc{AdEMAMix}. They motivate their optimizer heuristically: maintaining the optimizer's sensitivity to recent gradients with a fast momentum buffer while also retaining a much longer-horizon memory of past gradients with an independently weighted slow momentum buffer. For improved practical stability, they also rely on scheduling both the slow-momentum's timescale and its independent weighting over the entire training horizon. The strong empirical performance of \textsc{AdEMAMix} has been confirmed by other large-scale pretraining studies \citep{semenov2025benchmark,hernandezCano2026apertus}. Recent work has connected this construction to accelerated stochastic optimization, showing that \textsc{AdEMAMix} closely resembles accelerated stochastic gradient descent (SGD) variants, in which the weight on the fresh gradient is decoupled from the momentum timescale \citep{morwani2025connections}. A similar perspective inspires \citet{ferbach2026adana} to propose \textsc{Adana} as a closely related variant of \textsc{AdEMAMix}, augmented with logarithmic-time schedules for momenta and weight decay hyperparameters. These works interpret persistent memory through stochastic acceleration, stability, and scaling. Our work identifies a different mechanism: multi-timescale memory interacts with the geometry of normalized networks, shaping the realized rotational dynamics of the parameters.

\paragraph{Geometry of Scale-Invariant Optimization.}
Normalization layers make many parameter groups exactly or approximately scale-invariant, meaning that rescaling the norm of the parameters leaves the represented function and loss unchanged. This induces an intrinsic spherical geometry in which parameter directions alone determine the represented function. Early work established that the norm plays a key role in optimization, as it inversely modulates the rotational motion of the parameters \citep{vanLaarhoven2017,hoffer2018norm,liArora2020}: larger norms generally correspond to smaller rotational steps. \citet{roburin2022spherical} formalize first-order optimization through its induced motion on the hypersphere, deriving effective learning rates and directions for methods including momentum SGD and \textsc{Adam}, and showing how radial terms and accumulated momentum enter these quantities. \citet{wan2021smd} analyze the corresponding spherical motion dynamics under SGD and momentum SGD with weight decay, using the per-step rotation angle as a direct measure of directional motion. We build on this geometric foundation to study how scale invariance affects the rotational dynamics induced by multi-timescale optimizer memory.

\paragraph{Radial Dynamics and Rotational Equilibrium.}
The effect of optimizer memory on the parameter norm has been studied in several recent works. In the absence of weight decay \citet{heo2021adamp} showed that momentum reuses past updates to produce additional norm growth in scale-invariant parameters, and proposed \textsc{AdamP} to project out the radial component of the momentum update to prevent the associated slowdown of directional motion. With weight decay, \citet{kosson2024rotational} instead showed that stochastic updates and radial contraction can settle into a \emph{rotational equilibrium}, deriving equilibrium norms and angular update rates for several optimizers. For \textsc{AdamW}, their noise-dominated random-walk analysis predicts an equilibrium radius independent of the first-moment hyperparameter $\beta^{(f)}$ in the short-memory regime. Our analysis uncovers the mechanism underlying this cancellation and shows that an independently weighted second momentum timescale can have a much larger effect on the norm than its instantaneous power suggests. Concurrent with our work, \citet{wen2026hyperball} derived a related autocorrelation-weighted stationary radius expression under an idealized isotropic model, together with the corresponding angular step size. They use this analysis to motivate eliminating the radial dynamics by explicitly fixing parameter and update norms and optimizing on a fixed-radius sphere with the \textsc{Hyperball} optimizer. Our perspective instead treats the radial degree of freedom as a dynamical channel to be explained rather than removed.

Our work connects multi-timescale memory with the geometry of normalized networks, linking the stochastic-acceleration perspective to the dynamics of scale-invariant networks. We provide a mechanistic account of how \textsc{AdEMAMix}'s slow memory affects training: a memory channel can contribute little to the update applied at the current step while its correlated history leaves a much larger, persistent effect on the parameter norm. The resulting norm growth gradually anneals angular motion, transforming the learning-rate schedule into an optimizer- and memory-dependent rotational schedule. Through controlled interventions we show that this norm channel accounts for a large part of \textsc{AdEMAMix}'s advantage over \textsc{AdamW}.

\clearpage
\section{Scale-Invariant Geometry and Rotation Measurements}
\label{app:geometry}

\subsection{Gradient Tangency and Inverse-Radius Scaling}
\label{app:geometry-gradient}

Let $L$ be differentiable and scale-invariant on a nonzero parameter group:
\begin{equation}
    L(\kappa\bx)=L(\bx)\qquad(\kappa>0\text{ admissible}).
    \label{eq:scale-invariance}
\end{equation}
We record the short tangency argument from \citet[App.~B]{kosson2024rotational} and the gradient
scaling used later.

\begin{proof}[Gradient tangency and inverse-radius scaling]
Stretching the weight along its own direction cannot change the loss, so the loss has no slope in
that direction. Concretely, for every $h>-1$ the point $\bx+h\bx=(1+h)\bx$ is a positive rescaling
of $\bx$, so the invariance \eqref{eq:scale-invariance} gives $L(\bx+h\bx)=L(\bx)$: the loss is
constant along the whole ray through $\bx$. Its directional derivative along $\bx$ therefore
vanishes at $\bx$, and by the chain rule that derivative is the inner product we want,
\begin{equation}
    0=\frac{\dd}{\dd h}\,L(\bx+h\bx)\Big|_{h=0}
    =\ip{\nabla L(\bx)}{\bx},
    \label{eq:nqm-norm-orthogonality}
\end{equation}
Differentiating \eqref{eq:scale-invariance} in $\bx$ instead gives $\kappa\nabla L(\kappa\bx)=\nabla L(\bx)$, hence
\begin{equation}
    \nabla L(\kappa\bx)=\kappa^{-1}\nabla L(\bx).
    \label{eq:grad-homogeneity-minus-one}
\end{equation}
The raw gradient of a scale-invariant loss is homogeneous of degree $-1$ in the radius
\citep[Lem.~2.2]{arora2019}.
\end{proof}

\subsection{Exact Rotation and Relative-Update Proxies}
\label{app:geometry-rotation}

Recall the relative updates from \Cref{sec:si-geometry}:
\begin{equation}
    \qtot_t:=\frac{\lr_t\norm{\bu_t}}{\norm{\bx_t}},\qquad
    \qperp_t:=\frac{\lr_t\norm{\bu_{t,\perp}}}{\norm{\bx_t}},\qquad
    \qpar_t:=\frac{\lr_tu_{t,\parallel}}{\norm{\bx_t}}.
    \label{eq:eff-rate-rotation}
\end{equation}
The radial and tangential components are orthogonal, giving
\begin{equation}
    (\qtot_t)^2=(\qpar_t)^2+(\qperp_t)^2.
    \label{eq:qtot-pythagoras}
\end{equation}
The exact turning angle is
\begin{equation}
    \dphi_t=\arccos\frac{\ip{\bx_t}{\bx_{t+1}}}{\norm{\bx_t}\norm{\bx_{t+1}}}.
    \label{eq:eff-rate-angle}
\end{equation}

We derive the exact relation \eqref{eq:eff-rate-exact-angle} between $\dphi_t$ and the relative updates, followed by its small-update limit \eqref{eq:eff-rate-smallangle}. The scaled-radius comparison is illustrated in \Cref{fig:two-rate-geometry-main}.

\paragraph{The Exact Relation.}
\emph{Step 1 (decompose the update).} The update in the radial/tangential frame of $\bx_t$ is
$\bu_t=u_{t,\parallel}\,\xhat_t+\bu_{t,\perp}$.

\emph{Step 2 (the next weight in that frame).} Substitution into one decoupled step
$\bx_{t+1}=\delta_t\bx_t-\lr_t\bu_t$, with $\delta_t=1-\lr_t\lam_t$ and $\bx_t=\norm{\bx_t}\xhat_t$,
gives
\[
    \bx_{t+1}=\big(\delta_t\norm{\bx_t}-\lr_t u_{t,\parallel}\big)\,\xhat_t\;-\;\lr_t\,\bu_{t,\perp}.
\]
The bracket is the radial coordinate; $-\lr_t\bu_{t,\perp}$, of length $\lr_t\norm{\bu_{t,\perp}}$, is
the tangential coordinate.

\emph{Step 3 (read the angle off the two coordinates).} $\bx_t$ points purely along $\xhat_t$, so the
angle between $\bx_t$ and $\bx_{t+1}$ is the angle, in the plane those two directions span, of the
vector with radial coordinate $\delta_t\norm{\bx_t}-\lr_t u_{t,\parallel}$ and tangential coordinate
$\lr_t\norm{\bu_{t,\perp}}$. Whenever the radial coordinate is positive,
$\delta_t\norm{\bx_t}-\lr_t u_{t,\parallel}>0$, that angle is
\[
    \dphi_t=\arctan\frac{\lr_t\norm{\bu_{t,\perp}}}{\delta_t\norm{\bx_t}-\lr_t u_{t,\parallel}}.
\]

\emph{Step 4 (normalize by the radius).} Dividing numerator and denominator by $\norm{\bx_t}$ rewrites
them in the relative updates of \eqref{eq:eff-rate-rotation}, giving the exact relation
\begin{equation}
    \tan\dphi_t=\frac{\qperp_t}{\delta_t-\qpar_t},\qquad
    \dphi_t=\arctan\frac{\qperp_t}{\delta_t-\qpar_t}
    \qquad(\delta_t-\qpar_t>0).
    \label{eq:eff-rate-exact-angle}
\end{equation}
The condition $\delta_t-\qpar_t>0$ makes the radial coordinate positive and holds in the small-update regime used here.

\paragraph{The Small-Update Limit.}
Suppose the relative update is small, $\qtot_t\ll1$, and the shrink defect is small,
$1-\delta_t=\lr_t\lam_t\ll1$. Three steps take \eqref{eq:eff-rate-exact-angle} to
\eqref{eq:eff-rate-smallangle}.

\emph{Step 1 (the denominator is close to one).} The identity $\delta_t=1-\lr_t\lam_t$ and the
Pythagorean bound $|\qpar_t|\le\qtot_t$ from \eqref{eq:qtot-pythagoras} show that the denominator of
\eqref{eq:eff-rate-exact-angle} differs from one by at most the two small quantities assumed above,
\[
    \delta_t-\qpar_t=1-\lr_t\lam_t-\qpar_t=1-O(\lr_t\lam_t+\qtot_t).
\]

\emph{Step 2 (so the arctangent's argument is small).} Its numerator obeys $\qperp_t\le\qtot_t\ll1$
by \eqref{eq:qtot-pythagoras} as well, and Step~1 keeps its denominator near one, so the whole
argument $z_t=\qperp_t/(\delta_t-\qpar_t)$ is small, and dividing by that denominator gives
$z_t=\qperp_t\,[1+O(\lr_t\lam_t+\qtot_t)]$.

\emph{Step 3 (expand the arctangent about zero).} For small $z$ the Taylor series
$\arctan z=z-z^3/3+\cdots$ reads $\arctan z=z\,[1+O(z^2)]$. Applying it to $z_t$ and substituting
Step~2 gives \eqref{eq:eff-rate-smallangle}, the second form using
$\qperp_t=\qtot_t\norm{\bu_{t,\perp}}/\norm{\bu_t}$, which is \eqref{eq:eff-rate-rotation} with the
common factor $\lr_t/\norm{\bx_t}$ divided out.

The correction bracket $1+O(\lr_t\lam_t+\qtot_t)$ collects the two effects the steps introduced.
Step~1's denominator contributes the shrink defect $\lr_t\lam_t=1-\delta_t$ together with the radial
slice $\qpar_t$; Step~3's expansion contributes a further term of order $(\qperp_t)^2$, which the
bracket already covers.
Explicitly, the approximation derived above is
\begin{equation}
    \dphi_t=\qperp_t[1+O(\lr_t\lam_t+\qtot_t)]
    =\qtot_t\frac{\norm{\bu_{t,\perp}}}{\norm{\bu_t}}
      [1+O(\lr_t\lam_t+\qtot_t)].
    \label{eq:eff-rate-smallangle}
\end{equation}
The second form uses a nonzero update; for $\bu_t=0$, both the angle and $\qperp_t$ vanish. The error includes the decay contraction, radial displacement, and cubic arctangent correction.

\citet{kosson2024rotational} use the same exact angle (their Eq.~3) and adopt $\dphi\approx\qtot$ as an operational proxy. Equation~\eqref{eq:eff-rate-smallangle} identifies its tangential fraction and finite-step corrections.

\clearpage
\section{Proofs and Algebra for Memory-Induced Norm Cooling}
\label{app:proofs}

\subsection{The Update Correlation Law}
\label{app:proofs-exact}
For the constant-coefficient stationary arguments below, write
\begin{equation}
    \bx_{t+1}=\delta\bx_t-\lr\bu_t,
    \qquad \delta=1-\lr\lam\in(0,1).
    \label{eq:nqm-stationary-update}
\end{equation}
\begin{proof}[Proof of \Cref{lem:norm-growth}]
Expansion of the squared norm gives
\begin{equation}
    \begin{aligned}
    \norm{\bx_{t+1}}^2
    &=\norm{\delta_t \bx_t-\lr_t \bu_t}^2
    \\
    &=\delta_t^2\norm{\bx_t}^2
    -2\delta_t\lr_t\ip{\bu_t}{\bx_t}
    +\lr_t^2\norm{\bu_t}^2 .
    \end{aligned}
    \label{eq:nqm-norm-recursion}
\end{equation}
Symmetry of the inner product identifies \eqref{eq:nqm-norm-growth-compact}.
\end{proof}

\paragraph{Finiteness of the Weight-Decay-Weighted Functional.}
Symmetry of the real inner product gives
\begin{equation}
    C_u(-h)=C_u(h).
    \label{eq:applied-update-correlation-symmetry}
\end{equation}
The equilibrium sum converges under the same finite-power assumption. Cauchy--Schwarz, stationarity, and the geometric series give
\begin{equation}
    \begin{aligned}
    \lvert C_u(h)\rvert
    &\leq
    \sqrt{\E\norm{\bu_t}^2}
    \sqrt{\E\norm{\bu_{t-h}}^2}
    =C_u(0),
    \\
    \sum_{h\geq 1}\delta^h\lvert C_u(h)\rvert
    &\leq
    C_u(0)\sum_{h\geq 1}\delta^h
    =C_u(0)\frac{\delta}{1-\delta}
    <\infty.
    \end{aligned}
    \label{eq:applied-update-correlation-bound}
\end{equation}
In particular,
\begin{equation}
    \sum_{h\geq 1}\delta^h\lvert C_u(h)\rvert<\infty.
    \label{eq:applied-update-correlation-summability}
\end{equation}
This convergence concerns a scalar series of correlations and makes the functional below finite.
It does not require the correlations themselves to decay to zero: the geometric weight-decay factors
already make the series absolutely summable. The proof below uses finite unrollings, while the
preceding absolute-convergence bound controls the scalar correlation limit.

\begin{proof}[Proof of \Cref{thm:equilibrium-norm}]
Take expectations in the pathwise identity \eqref{eq:nqm-norm-growth-compact}. All terms are
finite: the squared terms by assumption and the cross term by Cauchy--Schwarz. Radius stationarity
and update stationarity give
\begin{equation*}
    (1-\delta^2)\bar R^2
    =-2\delta\lr\E\ip{\bx_t}{\bu_t}
      +\lr^2 C_u(0).
\end{equation*}
It remains to evaluate the radial cross moment. Iterating
\eqref{eq:nqm-stationary-update} backward through \(r\) steps gives the finite unrolling
\begin{equation*}
    \bx_t
    =\delta^r\bx_{t-r}
      -\lr\sum_{h=1}^{r}\delta^{h-1}\bu_{t-h}.
\end{equation*}
Taking its inner product with \(\bu_t\) and then expectation yields, at every finite \(r\),
\begin{equation*}
    \E\ip{\bx_t}{\bu_t}
    =\delta^r\E\ip{\bx_{t-r}}{\bu_t}
      -\lr\sum_{h=1}^{r}\delta^{h-1}
       \E\ip{\bu_{t-h}}{\bu_t}.
\end{equation*}
By Cauchy--Schwarz, radius stationarity, finite update power, and \(0<\delta<1\), the remote-state
term satisfies
\begin{equation*}
    \left\lvert
      \delta^r\E\ip{\bx_{t-r}}{\bu_t}
    \right\rvert
    \leq
    \delta^r
    \sqrt{\E\norm{\bx_{t-r}}^2}
    \sqrt{\E\norm{\bu_t}^2}
    =\delta^r\bar R\sqrt{C_u(0)}
    \xrightarrow[r\to\infty]{}0.
\end{equation*}
Symmetry of the real inner product and the definition of \(C_u(h)\) identify
\(\E\ip{\bu_{t-h}}{\bu_t}=C_u(h)\). Moreover,
\eqref{eq:applied-update-correlation-bound} and \(\delta>0\) give
\begin{equation*}
    \sum_{h\geq 1}\delta^{h-1}\lvert C_u(h)\rvert
    =\frac{1}{\delta}\sum_{h\geq 1}\delta^h\lvert C_u(h)\rvert
    <\infty.
\end{equation*}
Taking \(r\to\infty\) in the finite cross-moment identity therefore gives
\begin{equation*}
    \begin{aligned}
    \E\ip{\bx_t}{\bu_t}
    &=-\lr\sum_{h\geq 1}\delta^{h-1}C_u(h).
    \end{aligned}
\end{equation*}
This corresponds to the stationary projection coefficient of \citet[Lemma~4.4]{wen2026hyperball}, here for arbitrary update correlations.
Substitution into the stationary squared-radius balance yields
\begin{equation}
    \begin{aligned}
    (1-\delta^2)\bar R^2
    &=\lr^2 C_u(0)
      +2\lr^2\sum_{h\geq 1}\delta^h C_u(h)
    \\
    &=\lr^2\left[
      C_u(0)+2\sum_{h\geq 1}\delta^h C_u(h)
      \right]
    \\
    &=\lr^2\mathcal C_\delta[\bu].
    \end{aligned}
    \label{eq:nqm-norm-balance-exact}
\end{equation}
Since \(1-\delta^2>0\), division gives the squared-radius identity
\begin{equation*}
    \bar R^2
    =\frac{\lr^2\mathcal C_\delta[\bu]}{1-\delta^2}.
\end{equation*}
The radius identity includes the zero-power case $C_u(0)=0$. If $C_u(0)>0$, then $\bar R>0$: otherwise $\bx_t=\bx_{t+1}=0$ almost surely, and the update equation would force $\bu_t=0$ almost surely. Substituting the radius identity into $\qbarR^2=\lr^2 C_u(0)/\bar R^2$ gives the second identity in \eqref{eq:nqm-norm-equilibrium-colored}.
\end{proof}

\paragraph{Cumulative Weight-Decay Contraction and Unrolled Trajectory.}
\begin{definition}{Cumulative Weight-Decay Contraction}{survival-factor}
For $0\leq i\leq j$, set
\begin{equation}
    W_{i,j}:=\prod_{r=i}^{j-1}(1-\lr_r\lam_r),
    \qquad W_{j,j}=1.
    \label{eq:wd_survival}
\end{equation}
The finite update history obeys
\begin{equation}
    \bx_T=W_{0,T}\bx_0-\sum_{t=0}^{T-1}W_{t+1,T}\lr_t\bu_t.
    \label{eq:theta_unrolled}
\end{equation}
\end{definition}
\begin{proof}[Proof of the unrolling \eqref{eq:theta_unrolled} in \Cref{def:survival-factor}]
For compactness write \(\delta_t=1-\lr_t\lam_t\). Iterating the update \eqref{eq:param_update} backward from step \(T\),
\begin{align}
    \bx_T
    &= \delta_{T-1}\bx_{T-1} - \lr_{T-1}\bu_{T-1} \nonumber \\
    &= \delta_{T-1}\big(\delta_{T-2}\bx_{T-2}-\lr_{T-2}\bu_{T-2}\big)-\lr_{T-1}\bu_{T-1} \nonumber \\
    &= \delta_{T-1}\delta_{T-2}\bx_{T-2}
       - \delta_{T-1}\lr_{T-2}\bu_{T-2}
       - \lr_{T-1}\bu_{T-1} \nonumber \\
    &= \delta_{T-1}\delta_{T-2}\delta_{T-3}\bx_{T-3}
       - \delta_{T-1}\delta_{T-2}\lr_{T-3}\bu_{T-3}
       - \delta_{T-1}\lr_{T-2}\bu_{T-2}
       - \lr_{T-1}\bu_{T-1} \nonumber \\
    &= \cdots \nonumber \\
    &= \Bigg[\prod_{r=T-m}^{T-1}\delta_r\Bigg]\bx_{T-m}
       \;-\;
       \sum_{t=T-m}^{T-1}\Bigg[\prod_{r=t+1}^{T-1}\delta_r\Bigg]\lr_t\,\bu_t
       \nonumber \\
    &= \Bigg[\prod_{r=0}^{T-1}(1-\lr_r\lam_r)\Bigg]\bx_0
       \nonumber\\
    &\quad-\sum_{t=0}^{T-1}W_{t+1,T}\lr_t\,\bu_t.
\end{align}
Each backward substitution peels off one more shrink factor \(\delta_r=1-\lr_r\lam_r\); at \(m=T\) the recursion reaches \(\bx_0\), giving the last line. The bracketed product attached to step \(t\) is \(\prod_{r=t+1}^{T-1}(1-\lr_r\lam_r)\), which is \(W_{t+1,T}\) of \eqref{eq:wd_survival} at \(i=t+1\); the leading product \(\prod_{r=0}^{T-1}(1-\lr_r\lam_r)\) is \(W_{0,T}\) at \(i=0\). Collecting them gives \eqref{eq:theta_unrolled}.
\end{proof}

\subsection{Execution Kernels and the Uncorrelated-Input Model}
\label{app:proofs-represented}
\begin{definition}{Execution Kernel}{execution-kernel}
For a first-moment buffer with retentions $\rho_r$, define
\begin{equation}
    E_{t,i}:=(1-\rho_i)\prod_{r=i+1}^{t}\rho_r
    \qquad(0\leq i\leq t).
    \label{eq:E_kernel}
\end{equation}
With zero initialization, $\bmom_{t+1}=\sum_{i=0}^tE_{t,i}\bg_i$.
\end{definition}
\begin{assumption}{Finite Execution-Stream Representation}{tangent-execution-stream}
For inputs $\ba_i$, the zero-start modeled update is
$\widehat{\bu}_t=\sum_{i=0}^tE_{t,i}\ba_i$. For \textsc{AdEMAMix} the aggregate coefficient is
$E_{t,i}=E^{(f)}_{t,i}+\alpha_tE^{(s)}_{t,i}$. Equality with the actual update is an
optimizer-specific premise.
\end{assumption}
\begin{proof}[Proof of the buffer unrolling in \Cref{def:execution-kernel}]
For generic retentions \(\rho_r\), iterate
\(\bmom_{r+1}=\rho_r\bmom_r+(1-\rho_r)\bg_r\). The constant choice
\(\rho_r=\beta^{(f)}\) recovers the \textsc{AdamW} first-moment recurrence:
\begin{align}
    \bmom_{t+1}
    &= \rho_t \bmom_t + (1-\rho_t) \bg_t \nonumber\\
    &= \rho_t\rho_{t-1} \bmom_{t-1} + \rho_t(1-\rho_{t-1}) \bg_{t-1} + (1-\rho_t) \bg_t\nonumber\\
    &= \rho_t\rho_{t-1}\rho_{t-2} \bmom_{t-2} + \rho_t\rho_{t-1}(1-\rho_{t-2})\bg_{t-2} \nonumber\\
    &\quad + \rho_t(1-\rho_{t-1})\bg_{t-1} + (1-\rho_t) \bg_t.
\end{align}
Each substitution replaces the remaining state \(\bmom_{t-j}\) by \(\rho_{t-j-1}\bmom_{t-j-1}+(1-\rho_{t-j-1})\bg_{t-j-1}\), peeling off one more gradient and extending the leading \(\rho\)-product by one factor: if after \(m\) substitutions the state term reads \(\big(\prod_{r=t-m+1}^{t}\rho_r\big)\bmom_{t-m+1}\), substituting \(\bmom_{t-m+1}=\rho_{t-m}\bmom_{t-m}+(1-\rho_{t-m})\bg_{t-m}\) multiplies the product by \(\rho_{t-m}\) and appends the term \(\big(\prod_{r=t-m+1}^{t}\rho_r\big)(1-\rho_{t-m})\bg_{t-m}\), which is exactly the \(i=t-m\) term of the sum below; by induction, after \(m+1\) substitutions, where the oldest exposed gradient is \(\bg_{t-m}\),
\begin{align}
    \bmom_{t+1}
    &=
    \left(\prod_{r=t-m}^{t}\rho_r\right)\bmom_{t-m}
    \;+\;
    \sum_{i=t-m}^{t}
    (1-\rho_i)
    \left(\prod_{r=i+1}^{t}\rho_r\right)\bg_i.
\end{align}
The convention is that an empty product is one, so the newest gradient \(\bg_t\) has coefficient \(1-\rho_t\). Taking \(m=t\) sets the oldest exposed gradient to \(\bg_0\) and the state term to \(\big(\prod_{r=0}^{t}\rho_r\big)\bmom_0\), which vanishes since \(\bmom_0=\bzero\); this leaves \(\bmom_{t+1}=\sum_{i=0}^{t}E_{t,i}\bg_i\) with the execution kernel \(E_{t,i}=(1-\rho_i)\prod_{r=i+1}^{t}\rho_r\) of \eqref{eq:E_kernel}.
\end{proof}
The execution kernel records only when a gradient input reaches future buffers, before learning rate and weight decay are applied. A slow branch (larger \(\beta\)) has both a small newest-step coefficient \((1-\beta)\) and a long left tail of width \(O(\tau(\beta))\), and the startup transient from \(\bmom_0=\bzero\) fades within \(O(\tau(\beta))\) steps.

\paragraph{Optimizer Status of the Representation.}
For raw \textsc{SGD2mW}, the raw-gradient input gives exact representation equality:
\begin{equation*}
    \ba_i=\bg_i,
    \qquad
    \widehat{\bu}_t=\bu_t.
\end{equation*}
For \textsc{AdamW} and \textsc{AdEMAMix}, the injection-time choice
\begin{equation*}
    \ba_i:=\frac{\bg_i}{\sqrt{\bv_{i+1}}+\eps}
\end{equation*}
freezes each divisor when the input enters the buffer and defines a modeled output. The live optimizers divide the buffered numerator by the current \(\sqrt{\bv_{t+1}}+\eps\), so the modeled and actual updates generally differ.

Finite-history predictions under uncorrelated inputs retain the time-dependent coefficients \(E_{t,i}\) of this representation.

The kernel factors $\Gamma_e(h)$, $P[e]$, and $F_\delta[e]$ are defined in
\Cref{def:survival-weighted-forcing}.

\paragraph{Double-Sum Form of the Stationary Norm Factor.}
The same factor can be written as a sum over every ordered pair of input ages. For
\(0\leq\delta\leq1\), absolute summability of the kernel gives
\begin{equation*}
    \begin{aligned}
    \sum_{k\geq0}\sum_{n\geq0}
    \left\lvert e_ke_n\delta^{\lvert k-n\rvert}\right\rvert
    &\leq
    \sum_{k\geq0}\sum_{n\geq0}
    \lvert e_k\rvert\lvert e_n\rvert
    \\
    &=
    \left(\sum_{k\geq0}\lvert e_k\rvert\right)^2
    <\infty.
    \end{aligned}
\end{equation*}
This absolute convergence licenses splitting the double sum into its diagonal and its two strict
triangles:
\begin{equation*}
    \begin{aligned}
    \sum_{k\geq 0}\sum_{n\geq 0}
    e_k e_n\delta^{\lvert k-n\rvert}
    &=\sum_{n\geq 0}e_n^2
      +2\sum_{h\geq 1}\sum_{n\geq 0}
      e_{n+h}e_n\delta^h
    \\
    &=P[e]+2\sum_{h\geq 1}\delta^h\Gamma_e(h)
    =F_\delta[e].
    \end{aligned}
\end{equation*}
The diagonal pairs have no separation and therefore carry weight one. An off-diagonal pair separated by \(h\) steps carries the weight-decay contraction factor \(\delta^h\).

\paragraph{Stationary Modeled Update Under Uncorrelated Inputs.}
Let $(\ba_t)_{t\in\mathbb Z}$ be a two-sided input sequence satisfying
\Cref{ass:white-noise}, so $\E\ba_t=\bzero$ and
$\E\ip{\ba_t}{\ba_r}=C_a(0)\one_{t=r}$ with $C_a(0)<\infty$.
Let $e$ be a real age kernel with $\sum_{k\ge0}|e_k|<\infty$.
For a random vector $Y$, write $\norm{Y}_{L^2}=(\E\norm{Y}^2)^{1/2}$.
At fixed time $t$, define the finite cutoff
\[
 \widehat{\bu}^{(N)}_t=\sum_{k=0}^Ne_k\ba_{t-k}.
\]
The limit below increases the input-history depth at fixed $t$.

\begin{lemma}{Input Correlations and Modeled Update}{white-forcing}
The cutoffs converge in mean square to a zero-mean, finite-power modeled update
\begin{equation}
 \widehat{\bu}^{\infty}_t=\lim_{N\to\infty}\widehat{\bu}^{(N)}_t
 =\sum_{k\ge0}e_k\ba_{t-k}.
 \label{eq:age-kernel}
\end{equation}
Its temporal correlations satisfy, for $h\ge0$,
\begin{equation}
 C_{\widehat u}(h):=\E\ip{\widehat{\bu}^{\infty}_t}{\widehat{\bu}^{\infty}_{t-h}}
 =C_a(0)\Gamma_e(h).
 \label{eq:white-update-correlation}
\end{equation}
Thus its instantaneous update power is $C_a(0)P[e]$ and its scalar second-order statistics are stationary.
For $0\le\delta<1$,
\begin{equation}
 \mathcal C_\delta[\widehat{\bu}^{\infty}]
 :=C_{\widehat u}(0)+2\sum_{h\ge1}\delta^hC_{\widehat u}(h)
 .
 \label{eq:represented-update-correlation-functional}
\end{equation}
Then
\begin{equation}
 \mathcal C_\delta[\widehat{\bu}^{\infty}]=C_a(0)F_\delta[e].
 \label{eq:white-reduction}
\end{equation}
If the representation equals the actual update, then
\begin{equation}
 \mathcal C_\delta[\bu]=C_a(0)F_\delta[e].
 \label{eq:white-actual-update-bridge}
\end{equation}
\end{lemma}

\begin{proof}[Proof of \Cref{lem:white-forcing}]
\emph{Step 1 (construct the infinite output).}
Absolute summability implies square summability, since
\[
 \sum_{k\ge0}e_k^2\le\left(\sum_{k\ge0}|e_k|\right)^2<\infty.
\]
For $N>K$, expand the squared difference before taking its expectation:
\[
\begin{aligned}
 \E\norm{\widehat{\bu}^{(N)}_t-\widehat{\bu}^{(K)}_t}^2
 &=\sum_{k=K+1}^N\sum_{n=K+1}^Ne_ke_n
   \E\ip{\ba_{t-k}}{\ba_{t-n}}\\
 &=C_a(0)\sum_{k=K+1}^Ne_k^2
 \le C_a(0)\sum_{k>K}e_k^2\longrightarrow0.
\end{aligned}
\]
All sums in the expansion are finite. The assumption of uncorrelated inputs removes every pair with $k\ne n$;
the remaining tail tends to zero by square summability. The cutoffs are therefore
Cauchy in $L^2$, whose completeness gives the limit in \eqref{eq:age-kernel}.
The same expansion from $k=0$ yields
\[
 \norm{\widehat{\bu}^{(N)}_t}_{L^2}^2=C_a(0)\sum_{k=0}^Ne_k^2.
\]
The reverse triangle inequality gives
\[
 \left|\norm{\widehat{\bu}^{(N)}_t}_{L^2}
       -\norm{\widehat{\bu}^{\infty}_t}_{L^2}\right|
 \le\norm{\widehat{\bu}^{(N)}_t-\widehat{\bu}^{\infty}_t}_{L^2}
 \longrightarrow0.
\]
Consequently the limit has finite power
\[
 \E\norm{\widehat{\bu}^{\infty}_t}^2
 =C_a(0)\sum_{k\ge0}e_k^2=C_a(0)P[e].
\]
Each cutoff has zero mean. Cauchy--Schwarz also gives
\[
 \left\|\E\widehat{\bu}^{\infty}_t\right\|
 =\left\|\E(\widehat{\bu}^{\infty}_t-\widehat{\bu}^{(N)}_t)\right\|
 \le\E\norm{\widehat{\bu}^{\infty}_t-\widehat{\bu}^{(N)}_t}
 \le\norm{\widehat{\bu}^{\infty}_t-\widehat{\bu}^{(N)}_t}_{L^2}
 \longrightarrow0,
\]
so the modeled update has zero mean.

\emph{Step 2 (identify repeated input steps).}
Fix $h\ge0$ and take $N\ge h$. Bilinearity gives
\[
\begin{aligned}
 \E\ip{\widehat{\bu}^{(N)}_t}{\widehat{\bu}^{(N)}_{t-h}}
 &=\sum_{k=0}^N\sum_{n=0}^Ne_ke_n
   \E\ip{\ba_{t-k}}{\ba_{t-h-n}}\\
 &=C_a(0)\sum_{n=0}^{N-h}e_{n+h}e_n.
\end{aligned}
\]
A pair contributes exactly when $t-k=t-h-n$, or $k=n+h$. Both cutoff
indices must be at most $N$, which leaves $0\le n\le N-h$.
Cauchy--Schwarz for the coefficient sequences shows that
\[
 \sum_{n\ge0}|e_{n+h}e_n|
 \le\left(\sum_{n\ge0}e_{n+h}^2\right)^{1/2}
     \left(\sum_{n\ge0}e_n^2\right)^{1/2}
 \le P[e]<\infty.
\]
The finite sum therefore tends to $\Gamma_e(h)$.

To pass the left-hand side to its limit, add and subtract
$\E\ip{\widehat{\bu}^{\infty}_t}{\widehat{\bu}^{(N)}_{t-h}}$.
Applying Cauchy--Schwarz to the two differences gives
\[
\begin{aligned}
 &\left|\E\ip{\widehat{\bu}^{(N)}_t}{\widehat{\bu}^{(N)}_{t-h}}
       -\E\ip{\widehat{\bu}^{\infty}_t}{\widehat{\bu}^{\infty}_{t-h}}\right|\\
 &\quad\le
 \norm{\widehat{\bu}^{(N)}_t-\widehat{\bu}^{\infty}_t}_{L^2}
 \norm{\widehat{\bu}^{(N)}_{t-h}}_{L^2}
 +\norm{\widehat{\bu}^{\infty}_t}_{L^2}
 \norm{\widehat{\bu}^{(N)}_{t-h}-\widehat{\bu}^{\infty}_{t-h}}_{L^2}
 \longrightarrow0.
\end{aligned}
\]
The difference norms vanish by Step 1, and the other norms are bounded by
$\sqrt{C_a(0)P[e]}$. This proves \eqref{eq:white-update-correlation}.
The mean is zero and the correlation depends only on $h$. Negative step separations follow
by symmetry of the real inner product, so these are stationary scalar
second-order statistics.

\emph{Step 3 (sum correlations weighted by weight decay).}
The coefficient bound above gives $|\Gamma_e(h)|\le P[e]$. For $0\le\delta<1$,
\[
 \sum_{h\ge1}\delta^h|C_{\widehat u}(h)|
 \le C_a(0)P[e]\sum_{h\ge1}\delta^h
 =C_a(0)P[e]\frac{\delta}{1-\delta}<\infty.
\]
The correlation series is therefore absolutely convergent. Substituting the proved
correlation formula term by term gives
\[
\begin{aligned}
 \mathcal C_\delta[\widehat{\bu}^{\infty}]
 &=C_a(0)\Gamma_e(0)+2C_a(0)\sum_{h\ge1}\delta^h\Gamma_e(h)\\
 &=C_a(0)\left[P[e]+2\sum_{h\ge1}\delta^h\Gamma_e(h)\right]
 =C_a(0)F_\delta[e].
\end{aligned}
\]
If $\widehat{\bu}^{\infty}_t=\bu_t$, their inner products across different steps agree at
every step separation and hence their absolutely convergent weighted correlation sums agree. This gives
\eqref{eq:white-actual-update-bridge}.
\end{proof}

\begin{proof}[Proof of \Cref{prop:white-equilibrium}]
The uncorrelated-input identity and its value at $h=0$ supply
\begin{equation*}
    \mathcal C_\delta[\widehat{\bu}^{\infty}]=C_a(0)F_\delta[e],
    \qquad
    C_{\widehat u}(0)=C_a(0)P[e].
\end{equation*}
Applying \Cref{thm:equilibrium-norm} to the assumed modeled response and substituting the first
identity gives the stationary balance
\begin{equation*}
    \begin{aligned}
    (1-\delta^2)\bar R^2
    &=\lr^2
      \mathcal C_\delta[\widehat{\bu}^{\infty}]
    \\
    &=\lr^2C_a(0)F_\delta[e].
    \end{aligned}
\end{equation*}
Since \(1-\delta^2>0\), division by this factor gives
\begin{equation*}
    \bar R^2
    =\frac{\lr^2C_a(0)F_\delta[e]}{1-\delta^2},
\end{equation*}
which is \eqref{eq:white-radius-ratio-main}.

On the nondegenerate branch,
\begin{equation*}
    C_{\widehat u}(0)=C_a(0)P[e]>0.
\end{equation*}
If \(\bar R^2=0\), then \(\bx_t=\bx_{t+1}=\bzero\) almost surely. The modeled-response
recursion would give
\begin{equation*}
    \bzero
    =\bx_{t+1}-\delta\bx_t
    =-\lr\widehat{\bu}^{\infty}_t
    \qquad\text{almost surely},
\end{equation*}
contradicting \(C_{\widehat u}(0)>0\). Therefore \(\bar R^2>0\). Rearranging the exact radius
identity now gives
\begin{equation*}
    F_\delta[e]
    =\frac{(1-\delta^2)\bar R^2}{\lr^2C_a(0)}
    >0.
\end{equation*}
Applying \Cref{thm:equilibrium-norm} to the modeled
response and substituting its instantaneous power and total norm contribution gives
\begin{equation*}
    \qbarR^2
    =(1-\delta^2)
    \frac{C_a(0)P[e]}{C_a(0)F_\delta[e]}
    =(1-\delta^2)\frac{P[e]}{F_\delta[e]}.
\end{equation*}
When the representation equals the actual update, the promotion identities are
\begin{equation*}
    \mathcal C_\delta[\bu]
    =\mathcal C_\delta[\widehat{\bu}^{\infty}],
    \qquad
    C_u(0)=C_{\widehat u}(0),
\end{equation*}
so the same radius-stationary pair satisfies the actual-update recursion and the modeled
expressions coincide with the exact actual-update conclusions.
\end{proof}

\paragraph{Extension to a Nonzero Input Mean.}
\label{app:proofs-nonzero-mean}
Let $\E\ba_t=\bm\mu$ be constant, and suppose the centered inputs $\ba_t-\bm\mu$ have time-independent power $\sigma_a^2<\infty$ and zero inner-product correlations at distinct times. For a real absolutely summable kernel and $0<\delta<1$, the modeled update has
\begin{equation}
\begin{aligned}
 C_{\widehat u}(0)&=\sigma_a^2P[e]
       +\norm{\bm\mu}^2\left(\sum_{k\geq0}e_k\right)^2,\\
 \mathcal C_\delta[\widehat{\bu}]&=\sigma_a^2F_\delta[e]
       +\frac{1+\delta}{1-\delta}\norm{\bm\mu}^2
        \left(\sum_{k\geq0}e_k\right)^2.
\end{aligned}
\label{eq:nonzero-mean-factors}
\end{equation}
\begin{proof}
Decompose $\widehat{\bu}_t=\sum_{k\geq0}e_k(\ba_{t-k}-\bm\mu)
+\bm\mu\sum_{k\geq0}e_k$. The first term has zero mean and, by \Cref{lem:white-forcing}, correlation $\sigma_a^2\Gamma_e(h)$. The deterministic second term therefore adds $\norm{\bm\mu}^2(\sum_k e_k)^2$ to the correlation between any two updates. Its contribution to the functional is multiplied by $1+2\sum_{h\geq1}\delta^h=(1+\delta)/(1-\delta)$, giving \eqref{eq:nonzero-mean-factors}.
\end{proof}
For a radius-stationary response with finite second moment, substitution into \Cref{thm:equilibrium-norm} gives the corresponding radius and relative-update scale.

\subsection{Single-Timescale Memory}
\label{app:proofs-single}
\begin{proof}[Proof of \Cref{cor:single-buffer-cancellation}]
Since \(0\leq\beta<1\), the geometric series converges and the taps have unit gain:
\begin{equation*}
    \sum_{k\geq0}e_k^{\mathrm{AdamW}}
    =(1-\beta)\sum_{k\geq0}\beta^k
    =(1-\beta)\frac{1}{1-\beta}
    =1.
\end{equation*}
The instantaneous-power factor is another geometric series:
\begin{equation*}
    \begin{aligned}
        P[e^{\mathrm{AdamW}}]
        &=\sum_{k\geq0}(e_k^{\mathrm{AdamW}})^2
        =(1-\beta)^2\sum_{k\geq0}\beta^{2k}
        \\
        &=(1-\beta)^2\frac{1}{1-\beta^2}
        =\frac{1-\beta}{1+\beta}.
    \end{aligned}
\end{equation*}
For every step separation \(h\geq0\), the kernel overlap is
\begin{equation*}
    \begin{aligned}
        \Gamma_{e^{\mathrm{AdamW}}}(h)
        &=\sum_{n\geq0}e_n^{\mathrm{AdamW}}e_{n+h}^{\mathrm{AdamW}}
        \\
        &=(1-\beta)^2\sum_{n\geq0}\beta^n\beta^{n+h}
        =(1-\beta)^2\beta^h\sum_{n\geq0}\beta^{2n}
        \\
        &=\frac{1-\beta}{1+\beta}\,\beta^h
        =P[e^{\mathrm{AdamW}}]\beta^h.
    \end{aligned}
\end{equation*}
Substituting this overlap into the definition of \(F_\delta[e^{\mathrm{AdamW}}]\), and summing the remaining
geometric series, gives
\begin{equation*}
    \begin{aligned}
        F_\delta[e^{\mathrm{AdamW}}]
        &=P[e^{\mathrm{AdamW}}]+2\sum_{h\geq1}\delta^h\Gamma_{e^{\mathrm{AdamW}}}(h)
        \\
        &=P[e^{\mathrm{AdamW}}]\left(1+2\sum_{h\geq1}(\beta\delta)^h\right)
        \\
        &=P[e^{\mathrm{AdamW}}]\left(1+\frac{2\beta\delta}{1-\beta\delta}\right)
        =P[e^{\mathrm{AdamW}}]\frac{1+\beta\delta}{1-\beta\delta}
        \\
        &=\frac{1-\beta}{1+\beta}
          \frac{1+\beta\delta}{1-\beta\delta},
    \end{aligned}
\end{equation*}
which proves \eqref{eq:nqm-norm-single-buffer-forcing}.

The ratio $P[e^{\mathrm{AdamW}}]/F_\delta[e^{\mathrm{AdamW}}]$ cancels the common factor:
\begin{equation*}
    \frac{P[e^{\mathrm{AdamW}}]}{F_\delta[e^{\mathrm{AdamW}}]}
    =\frac{P[e^{\mathrm{AdamW}}]}
           {P[e^{\mathrm{AdamW}}](1+\beta\delta)/(1-\beta\delta)}
    =\frac{1-\beta\delta}{1+\beta\delta}.
\end{equation*}

Differentiating the power directly gives
\begin{equation*}
    \frac{\dd}{\dd\beta}\frac{1-\beta}{1+\beta}
    =\frac{-(1+\beta)-(1-\beta)}{(1+\beta)^2}
    =-\frac{2}{(1+\beta)^2}<0.
\end{equation*}

For $F_\delta[e^{\mathrm{AdamW}}]$, logarithmic differentiation of the positive closed form gives
\begin{equation*}
    \begin{aligned}
        \frac{1}{F_\delta[e^{\mathrm{AdamW}}]}\frac{\dd F_\delta[e^{\mathrm{AdamW}}]}{\dd\beta}
        &=-\frac{1}{1-\beta}-\frac{1}{1+\beta}
          +\frac{\delta}{1+\beta\delta}
          +\frac{\delta}{1-\beta\delta}
        \\
        &=-\frac{2}{1-\beta^2}
          +\frac{2\delta}{1-\beta^2\delta^2}
        \\
        &=-\frac{2(1-\delta)(1+\beta^2\delta)}
                {(1-\beta^2)(1-\beta^2\delta^2)}.
    \end{aligned}
\end{equation*}
Multiplying by the closed form of \(F_\delta[e^{\mathrm{AdamW}}]\), and cancelling
\(1-\beta^2=(1-\beta)(1+\beta)\) and
\(1-\beta^2\delta^2=(1-\beta\delta)(1+\beta\delta)\), gives
\begin{equation*}
    \frac{\dd F_\delta[e^{\mathrm{AdamW}}]}{\dd\beta}
    =-\frac{2(1-\delta)(1+\beta^2\delta)}
            {(1+\beta)^2(1-\beta\delta)^2}.
\end{equation*}
Finally, differentiating the ratio gives
\begin{equation*}
    \frac{\dd}{\dd\beta}\frac{1-\beta\delta}{1+\beta\delta}
    =\frac{-\delta(1+\beta\delta)-\delta(1-\beta\delta)}
           {(1+\beta\delta)^2}
    =-\frac{2\delta}{(1+\beta\delta)^2}.
\end{equation*}
The displayed signs follow from the stated parameter domains.
\end{proof}

At $0<\delta<1$, increasing the sole buffer's retention lowers both $F_\delta[e^{\mathrm{AdamW}}]$ and
the ratio $P[e^{\mathrm{AdamW}}]/F_\delta[e^{\mathrm{AdamW}}]$. Under the zero-mean, uncorrelated-input response assumptions, this lowers the
modeled radius at equal input power while cooling the stationary moment ratio.
The cancellation found in short-memory analyses follows directly from the closed form:
if $1-\delta\ll1-\beta$, then $1-\beta\delta\simeq1-\beta$ and
$1+\beta\delta\simeq1+\beta$, so $F_\delta[e^{\mathrm{AdamW}}]\simeq1$.
Pure scalar gain multiplies both $P[e^{\mathrm{AdamW}}]$ and $F_\delta[e^{\mathrm{AdamW}}]$ by the same squared factor,
leaving their ratio unchanged. Adding a distinct slow pole changes the temporal
shape as well as the gain.

\subsection{Two-Timescale \textsc{AdEMAMix} Kernel Algebra}
\label{app:proofs-two}
\subsubsection{Exact Pairwise Contribution}

The two-channel kernel is
\begin{equation}
    e_k^{\mathrm{AMix}}=(1-\beta^{(f)})(\beta^{(f)})^k+\alpha(1-\beta^{(s)})(\beta^{(s)})^k.
    \label{eq:nqm-mixed-age-kernel}
\end{equation}

The definitions in \Cref{sec:memory-correlation} and the ordered-pair identity proved above give
\begin{equation*}
    P[e^{\mathrm{AMix}}]=\sum_{k\geq0}(e_k^{\mathrm{AMix}})^2,
    \qquad
    F_\delta[e^{\mathrm{AMix}}]=\sum_{k,n\geq0}e_k^{\mathrm{AMix}}e_n^{\mathrm{AMix}}\delta^{|k-n|}.
\end{equation*}
With \(0^0=1\), setting \(\delta=0\) in the ordered-pair sum retains exactly the diagonal
\(k=n\). Hence
\begin{equation*}
    P[e^{\mathrm{AMix}}]=\left.F_\delta[e^{\mathrm{AMix}}]\right|_{\delta=0}.
\end{equation*}

Insert the two-channel kernel \eqref{eq:nqm-mixed-age-kernel} into the two ordered slots of
\(F_\delta[e^{\mathrm{AMix}}]\). Expanding the product before combining either mixed term gives
\begin{equation*}
    \begin{aligned}
    F_\delta[e^{\mathrm{AMix}}]
    &=\sum_{k,n\geq0}
      \Bigl[(1-\beta^{(f)})(\beta^{(f)})^k
      +\alpha(1-\beta^{(s)})(\beta^{(s)})^k\Bigr]
      \cdot\delta^{|k-n|}
    \\
    &\qquad\cdot
      \Bigl[(1-\beta^{(f)})(\beta^{(f)})^n
      +\alpha(1-\beta^{(s)})(\beta^{(s)})^n\Bigr]
    \\
    &=\sum_{k,n\geq0}
      (1-\beta^{(f)})(\beta^{(f)})^k
      \cdot(1-\beta^{(f)})(\beta^{(f)})^n
      \cdot\delta^{|k-n|}
    \\
    &\quad+\alpha\sum_{k,n\geq0}
      (1-\beta^{(f)})(\beta^{(f)})^k
      \cdot(1-\beta^{(s)})(\beta^{(s)})^n
      \cdot\delta^{|k-n|}
    \\
    &\quad+\alpha\sum_{k,n\geq0}
      (1-\beta^{(s)})(\beta^{(s)})^k
      \cdot(1-\beta^{(f)})(\beta^{(f)})^n
      \cdot\delta^{|k-n|}
    \\
    &\quad+\alpha^2\sum_{k,n\geq0}
      (1-\beta^{(s)})(\beta^{(s)})^k
      \cdot(1-\beta^{(s)})(\beta^{(s)})^n
      \cdot\delta^{|k-n|}.
    \end{aligned}
\end{equation*}
Here \(k\) and \(n\) are two execution ages of the same input increment, and
\(\delta^{|k-n|}\) weights their separation by weight-decay contraction. The
four ordered sums are the fast--fast, fast--slow, slow--fast, and slow--slow contributions. Their
diagonal terms, summed over all four branch combinations, give instantaneous power. Their unequal-age
terms add temporal correlations weighted by weight decay to the instantaneous-power factor.

Each of the four sums has the same generic form: it pairs one normalized geometric age profile
with another. The following helper evaluates that pairing for any two poles.

\begin{lemma}{Exact Weighted Pairing of Two EMA Poles}{two-decay-forcing}
For $\beta,\beta'\in[0,1)$ and $\delta\in[0,1]$, define
$\Phi_\delta(\beta,\beta'):=\sum_{k,n\geq0}(1-\beta)\beta^k
(1-\beta')\beta'^{\,n}\delta^{|k-n|}$. It is positive and symmetric, with
\begin{equation}
    \Phi_\delta(\beta,\beta')
    =\frac{(1-\beta)(1-\beta')(1-\delta^2\beta\beta')}
           {(1-\beta\beta')(1-\delta\beta)(1-\delta\beta')}.
    \label{eq:nqm-branch-forcing-geometric}
\end{equation}
\end{lemma}
\begin{proof}[Proof of \Cref{lem:two-decay-forcing}]
\emph{Step 1 (convergence and symmetry).}
Since \(0\leq\delta^{|k-n|}\leq1\), the defining double series satisfies
\begin{equation*}
    \begin{aligned}
        &\sum_{k,n\geq0}
        \left|
        (1-\beta)\beta^k
        (1-\beta')\beta'^{\,n}
        \delta^{|k-n|}
        \right|
        \\
        &\qquad\leq
        (1-\beta)\sum_{k\geq0}\beta^k
        (1-\beta')\sum_{n\geq0}\beta'^{\,n}
        =1.
    \end{aligned}
\end{equation*}
It therefore converges absolutely and may be split and reindexed. Exchanging \(k\) with \(n\) and
\(\beta\) with \(\beta'\) leaves the sum unchanged, which proves symmetry.

\emph{Step 2 (the diagonal).}
On \(k=n\), the weight-decay contraction factor equals one, so the diagonal contribution is
\begin{equation*}
    \begin{aligned}
        \sum_{n\geq0}
        (1-\beta)(1-\beta')(\beta\beta')^n
        &=(1-\beta)(1-\beta')
          \frac{1}{1-\beta\beta'}.
    \end{aligned}
\end{equation*}

\emph{Step 3 (the triangle \(k>n\)).}
Write \(k=n+h\) with \(h\geq1\). Absolute convergence justifies this reindexing, and hence
\begin{equation*}
    \begin{aligned}
        &\sum_{n\geq0}\sum_{h\geq1}
        (1-\beta)\beta^{n+h}
        (1-\beta')\beta'^{\,n}
        \delta^h
        \\
        &\qquad=(1-\beta)(1-\beta')
        \left(\sum_{n\geq0}(\beta\beta')^n\right)
        \left(\sum_{h\geq1}(\beta\delta)^h\right)
        \\
        &\qquad=
        \frac{(1-\beta)(1-\beta')}{1-\beta\beta'}
        \frac{\beta\delta}{1-\beta\delta}.
    \end{aligned}
\end{equation*}

\emph{Step 4 (the triangle \(n>k\)).}
Writing \(n=k+h\) in the same way gives
\begin{equation*}
    \begin{aligned}
        &\sum_{k\geq0}\sum_{h\geq1}
        (1-\beta)\beta^k
        (1-\beta')\beta'^{\,k+h}
        \delta^h
        \\
        &\qquad=
        \frac{(1-\beta)(1-\beta')}{1-\beta\beta'}
        \frac{\beta'\delta}{1-\beta'\delta}.
    \end{aligned}
\end{equation*}

\emph{Step 5 (common denominator).}
Adding the diagonal and the two triangles yields
\begin{equation*}
    \Phi_\delta(\beta,\beta')
    =\frac{(1-\beta)(1-\beta')}{1-\beta\beta'}
      \left(
        1+\frac{\beta\delta}{1-\beta\delta}
         +\frac{\beta'\delta}{1-\beta'\delta}
      \right).
\end{equation*}
The numerator after putting the bracket over a common denominator is
\begin{equation*}
    \begin{aligned}
        &(1-\beta\delta)(1-\beta'\delta)
        +\beta\delta(1-\beta'\delta)
        +\beta'\delta(1-\beta\delta)
        \\
        &\qquad=1-\delta^2\beta\beta'.
    \end{aligned}
\end{equation*}
Substitution proves \eqref{eq:nqm-branch-forcing-geometric}. Every factor in that expression is
positive on the stated parameter range, so the pairing is strictly positive.

\end{proof}

Two immediate parameter evaluations will be used below. At \(\delta=0\), only equal execution
ages remain paired:
\begin{equation*}
    \Phi_0(\beta,\beta')
    =\frac{(1-\beta)(1-\beta')}{1-\beta\beta'}.
\end{equation*}
At \(\delta=1\), every ordered age pair receives full credit, and the closed form gives
\begin{equation*}
    \begin{aligned}
        \Phi_1(\beta,\beta')
        &=\frac{(1-\beta)(1-\beta')(1-\beta\beta')}
                {(1-\beta\beta')(1-\beta)(1-\beta')}
        =1.
    \end{aligned}
\end{equation*}

Apply the proved identity to the four ordered sums in the preceding expansion. In their displayed
order, they are
\begin{equation*}
    \Phi_\delta(\beta^{(f)},\beta^{(f)}),
    \qquad
    \alpha\Phi_\delta(\beta^{(f)},\beta^{(s)}),
    \qquad
    \alpha\Phi_\delta(\beta^{(s)},\beta^{(f)}),
    \qquad
    \alpha^2\Phi_\delta(\beta^{(s)},\beta^{(s)}).
\end{equation*}
The proved symmetry merges the two mixed terms. Since
\(P[e^{\mathrm{AMix}}]=\left.F_\delta[e^{\mathrm{AMix}}]\right|_{\delta=0}\), the two macroscopic quantities are therefore
\begin{align}
    P[e^{\mathrm{AMix}}]
    &=\Phi_0(\beta^{(f)},\beta^{(f)})
      +2\alpha\Phi_0(\beta^{(f)},\beta^{(s)})
      +\alpha^2\Phi_0(\beta^{(s)},\beta^{(s)}),
    \label{eq:nqm-two-channel-power}
    \\
    F_\delta[e^{\mathrm{AMix}}]
    &=\Phi_\delta(\beta^{(f)},\beta^{(f)})
      +2\alpha\Phi_\delta(\beta^{(f)},\beta^{(s)})
      +\alpha^2\Phi_\delta(\beta^{(s)},\beta^{(s)}).
    \label{eq:nqm-branch-forcing-decomposition}
\end{align}
Expanding the two kernel slots produces three contributions. The fast--fast pairing is the baseline;
the two symmetric fast--slow pairings form the first-order \(\alpha\) contribution; and the slow--slow
pairing enters at order \(\alpha^2\) and controls large amplitudes. Power \(P[e^{\mathrm{AMix}}]\) evaluates each
pairing at \(\delta=0\), so it credits equal-age executions. The full factor \(F_\delta[e^{\mathrm{AMix}}]\) additionally
includes positive relative-age separations weighted by \(\delta^{|k-n|}\). The comparison below therefore
depends on all three pairings.
Equations~\eqref{eq:nqm-two-channel-power} and
\eqref{eq:nqm-branch-forcing-decomposition} are exact deterministic kernel identities.

\begin{proof}[Proof of \Cref{thm:ademamix-lever}]
\emph{Step 1 (both kernel factors increase).}
The closed form \eqref{eq:nqm-branch-forcing-geometric} is strictly positive for every admissible
pair of poles, including at \(\delta=0\). Differentiating the two exact quadratics gives
\begin{equation*}
    \begin{aligned}
        \frac{\dd P[e^{\mathrm{AMix}}]}{\dd\alpha}
        &=2\Phi_0(\beta^{(f)},\beta^{(s)})
          +2\alpha\Phi_0(\beta^{(s)},\beta^{(s)})>0,
        \\
        \frac{\dd F_\delta[e^{\mathrm{AMix}}]}{\dd\alpha}
        &=2\Phi_\delta(\beta^{(f)},\beta^{(s)})
          +2\alpha\Phi_\delta(\beta^{(s)},\beta^{(s)})>0.
    \end{aligned}
\end{equation*}
This proves that both $P[e^{\mathrm{AMix}}]$ and $F_\delta[e^{\mathrm{AMix}}]$ increase with the slow amplitude.

\emph{Step 2 (three pairwise comparisons).}
For \(0<\delta<1\), dividing the closed form by its value at \(\delta=0\) gives
\begin{equation*}
    \frac{\Phi_\delta(\beta,\beta')}{\Phi_0(\beta,\beta')}
    =\frac{1-\delta^2\beta\beta'}
           {(1-\delta\beta)(1-\delta\beta')}.
\end{equation*}
For the fast--fast and fast--slow pairings, direct subtraction gives
\begin{equation*}
    \begin{aligned}
        &\Phi_0(\beta^{(f)},\beta^{(s)})\Phi_\delta(\beta^{(f)},\beta^{(f)})
        -\Phi_0(\beta^{(f)},\beta^{(f)})\Phi_\delta(\beta^{(f)},\beta^{(s)})
        \\
        &\quad=
        \Phi_0(\beta^{(f)},\beta^{(f)})\Phi_0(\beta^{(f)},\beta^{(s)})
        \left[
          \frac{1+\delta\beta^{(f)}}{1-\delta\beta^{(f)}}
          -\frac{1-\delta^2\beta^{(f)}\beta^{(s)}}
                 {(1-\delta\beta^{(f)})(1-\delta\beta^{(s)})}
        \right]
        \\
        &\quad=
        -\frac{\delta(\beta^{(s)}-\beta^{(f)})}
               {(1-\delta\beta^{(f)})(1-\delta\beta^{(s)})}
        \Phi_0(\beta^{(f)},\beta^{(f)})\Phi_0(\beta^{(f)},\beta^{(s)}).
    \end{aligned}
\end{equation*}
Comparing the two diagonal pairings gives
\begin{equation*}
    \begin{aligned}
        &\Phi_0(\beta^{(s)},\beta^{(s)})\Phi_\delta(\beta^{(f)},\beta^{(f)})
        -\Phi_0(\beta^{(f)},\beta^{(f)})\Phi_\delta(\beta^{(s)},\beta^{(s)})
        \\
        &\quad=
        \Phi_0(\beta^{(f)},\beta^{(f)})\Phi_0(\beta^{(s)},\beta^{(s)})
        \left[
          \frac{1+\delta\beta^{(f)}}{1-\delta\beta^{(f)}}
          -\frac{1+\delta\beta^{(s)}}{1-\delta\beta^{(s)}}
        \right]
        \\
        &\quad=
        -\frac{2\delta(\beta^{(s)}-\beta^{(f)})}
               {(1-\delta\beta^{(f)})(1-\delta\beta^{(s)})}
        \Phi_0(\beta^{(f)},\beta^{(f)})\Phi_0(\beta^{(s)},\beta^{(s)}).
    \end{aligned}
\end{equation*}
Finally, the fast--slow and slow--slow comparison is
\begin{equation*}
    \begin{aligned}
        &\Phi_0(\beta^{(s)},\beta^{(s)})\Phi_\delta(\beta^{(f)},\beta^{(s)})
        -\Phi_0(\beta^{(f)},\beta^{(s)})\Phi_\delta(\beta^{(s)},\beta^{(s)})
        \\
        &\quad=
        \Phi_0(\beta^{(f)},\beta^{(s)})\Phi_0(\beta^{(s)},\beta^{(s)})
        \left[
          \frac{1-\delta^2\beta^{(f)}\beta^{(s)}}
               {(1-\delta\beta^{(f)})(1-\delta\beta^{(s)})}
          -\frac{1+\delta\beta^{(s)}}{1-\delta\beta^{(s)}}
        \right]
        \\
        &\quad=
        -\frac{\delta(\beta^{(s)}-\beta^{(f)})}
               {(1-\delta\beta^{(f)})(1-\delta\beta^{(s)})}
        \Phi_0(\beta^{(f)},\beta^{(s)})\Phi_0(\beta^{(s)},\beta^{(s)}).
    \end{aligned}
\end{equation*}
Every denominator and every \(\Phi_0\) factor in these expressions is positive. Their signs are
therefore controlled only by \(\delta(\beta^{(s)}-\beta^{(f)})\).

\emph{Step 3 (the quotient derivative).}
Using the derivatives from Step 1, expand the quotient numerator in powers of $\alpha$:
\begin{equation*}
\begin{aligned}
 &\frac12\left(\frac{\dd P[e^{\mathrm{AMix}}]}{\dd\alpha}F_\delta[e^{\mathrm{AMix}}]
      -P[e^{\mathrm{AMix}}]\frac{\dd F_\delta[e^{\mathrm{AMix}}]}{\dd\alpha}\right)\\
 &=\Phi_0(\beta^{(f)},\beta^{(s)})\Phi_\delta(\beta^{(f)},\beta^{(f)})
   -\Phi_0(\beta^{(f)},\beta^{(f)})\Phi_\delta(\beta^{(f)},\beta^{(s)})\\
 &\quad+\alpha\bigl[\Phi_0(\beta^{(s)},\beta^{(s)})\Phi_\delta(\beta^{(f)},\beta^{(f)})
   -\Phi_0(\beta^{(f)},\beta^{(f)})\Phi_\delta(\beta^{(s)},\beta^{(s)})\bigr]\\
 &\quad+\alpha^2\bigl[\Phi_0(\beta^{(s)},\beta^{(s)})\Phi_\delta(\beta^{(f)},\beta^{(s)})
   -\Phi_0(\beta^{(f)},\beta^{(s)})\Phi_\delta(\beta^{(s)},\beta^{(s)})\bigr].
\end{aligned}
\end{equation*}
Substituting the three factorizations from Step 2 and dividing by
\(F_\delta[e^{\mathrm{AMix}}]^2\) yields
\begin{equation*}
    \begin{aligned}
        \frac{\dd}{\dd\alpha}\frac{P[e^{\mathrm{AMix}}]}{F_\delta[e^{\mathrm{AMix}}]}
        &=
        -\frac{2\delta(\beta^{(s)}-\beta^{(f)})}
               {F_\delta[e^{\mathrm{AMix}}]^2
                (1-\delta\beta^{(f)})(1-\delta\beta^{(s)})}
        \\
        &\quad\cdot
        \Bigl[
          \Phi_0(\beta^{(f)},\beta^{(f)})\Phi_0(\beta^{(f)},\beta^{(s)})
        \\
        &\hspace{12mm}
          +2\alpha\Phi_0(\beta^{(f)},\beta^{(f)})\Phi_0(\beta^{(s)},\beta^{(s)})
        \\
        &\hspace{12mm}
          +\alpha^2\Phi_0(\beta^{(f)},\beta^{(s)})\Phi_0(\beta^{(s)},\beta^{(s)})
        \Bigr].
    \end{aligned}
\end{equation*}
Under the ordered-pole assumptions, every factor outside the leading minus sign is positive. This
proves the strict ratio decrease and completes the deterministic claim.
\end{proof}

The sign factor isolates the boundary and reversed-order cases. At \(\delta=0\), distinct execution ages receive
zero weight and the ratio is constant in \(\alpha\). At coincident poles, the two branches have the
same age profile, so varying \(\alpha\) is pure scalar gain and the ratio is again constant.
Reversing the pole order reverses the sign of the ratio derivative.

\begin{assumption}{Equal Input Power}{matched-source-power}
Across the compared kernels or amplitudes, the input power $C_a(0)$ is the same
positive constant and is independent of the modeled radius.
\end{assumption}

\paragraph{Modeled-Response Reading.}
Suppose now that, for every amplitude in the comparison, the input satisfies
\Cref{ass:white-noise}, has positive power, and drives a finite-second-moment radius-stationary
response. Substituting the exact quadratics into \eqref{eq:white-radius-ratio-main} gives
\begin{equation}
    \left.\qbarR^2(\alpha)\right|_{\mathrm{white\ model}}
    =(1-\delta^2)
    \frac{
      \begin{aligned}
        &\Phi_0(\beta^{(f)},\beta^{(f)})
        +2\alpha\Phi_0(\beta^{(f)},\beta^{(s)})\\
        &\quad+\alpha^2\Phi_0(\beta^{(s)},\beta^{(s)})
      \end{aligned}
    }{
      \begin{aligned}
        &\Phi_\delta(\beta^{(f)},\beta^{(f)})
        +2\alpha\Phi_\delta(\beta^{(f)},\beta^{(s)})\\
        &\quad+\alpha^2\Phi_\delta(\beta^{(s)},\beta^{(s)})
      \end{aligned}
    }.
    \label{eq:nqm-two-channel-qbar}
\end{equation}
Because \(1-\delta^2>0\), differentiating this identity and using the theorem gives
\begin{equation}
    \frac{\dd}{\dd\alpha}
    \left.\qbarR^2(\alpha)\right|_{\mathrm{white\ model}}
    =(1-\delta^2)
    \frac{\dd}{\dd\alpha}\frac{P[e^{\mathrm{AMix}}]}{F_\delta[e^{\mathrm{AMix}}]}<0.
    \label{eq:nqm-two-channel-qbar-derivative}
\end{equation}
This moment-ratio decrease needs no comparison of input powers. Under the additional matched-power
premise of \Cref{ass:matched-source-power}, the radius identity gives
\begin{equation*}
    \frac{\dd}{\dd\alpha}
    \left.\bar R^2(\alpha)\right|_{\mathrm{white\ model}}
    =\frac{\lr^2C_a(0)}{1-\delta^2}
      \frac{\dd F_\delta[e^{\mathrm{AMix}}]}{\dd\alpha}>0.
\end{equation*}
The radius derivative uses matched input power; the moment-ratio derivative does not.

\begin{corollary}{Two-Channel Cooling with a Constant Input Mean}{two-channel-nonzero-mean}
Fix $\lr>0$, $0<\delta<1$, and $0\leq\beta^{(f)}<\beta^{(s)}<1$.
Let the inputs have constant mean $\bm\mu$ and centered power $\sigma_a^2$, with
zero centered inner-product correlations at distinct times, as in
\eqref{eq:nonzero-mean-factors}. Hold $\bm\mu$ and $\sigma_a^2$ fixed as $\alpha\geq0$
varies, with $\sigma_a^2+\norm{\bm\mu}^2>0$, and assume that each modeled response
has a finite stationary squared radius. For the two-channel kernel $e=e^{\mathrm{AMix}}$,
\begin{equation}
\begin{aligned}
 \bar R^2(\alpha)&=\frac{\lr^2}{1-\delta^2}
 \left[\sigma_a^2F_\delta[e]
       +\frac{1+\delta}{1-\delta}\norm{\bm\mu}^2(1+\alpha)^2\right],\\
 \qbarR^2(\alpha)&=(1-\delta^2)
 \frac{\sigma_a^2P[e]+\norm{\bm\mu}^2(1+\alpha)^2}
      {\sigma_a^2F_\delta[e]+\frac{1+\delta}{1-\delta}\norm{\bm\mu}^2(1+\alpha)^2}.
\end{aligned}
\label{eq:nqm-two-channel-nonzero-mean}
\end{equation}
The modeled squared radius strictly increases with $\alpha$. The relative-update
scale strictly decreases when $\sigma_a^2>0$; for a pure mean input,
$\sigma_a^2=0$, it equals $\qbarR=1-\delta$ independently of $\alpha$.
\end{corollary}

\begin{proof}
\emph{Step 1: Derive the identities in \eqref{eq:nqm-two-channel-nonzero-mean}.}
The response formulas depend on how the two buffers transmit the constant input
mean. Each EMA has unit total weight, so the two-channel kernel satisfies
\[
\begin{aligned}
 \sum_{k\geq0}e_k
 &=(1-\beta^{(f)})\sum_{k\geq0}(\beta^{(f)})^k
   +\alpha(1-\beta^{(s)})\sum_{k\geq0}(\beta^{(s)})^k\\
 &=1+\alpha.
\end{aligned}
\]
The update mean is therefore $(1+\alpha)\bm\mu$. To determine its contribution
to $\mathcal C_\delta[\widehat{\bu}]$, note that its squared norm
$\norm{\bm\mu}^2(1+\alpha)^2$ contributes to the correlation between updates
at any two times. The decay weights sum to
\[
 1+2\sum_{h\geq1}\delta^h
 =1+\frac{2\delta}{1-\delta}
 =\frac{1+\delta}{1-\delta}.
\]
Consequently, \eqref{eq:nonzero-mean-factors} becomes
\[
\begin{aligned}
 C_{\widehat u}(0)
 &=\sigma_a^2P[e]+\norm{\bm\mu}^2(1+\alpha)^2,\\
 \mathcal C_\delta[\widehat{\bu}]
 &=\sigma_a^2F_\delta[e]
   +\frac{1+\delta}{1-\delta}\norm{\bm\mu}^2(1+\alpha)^2.
\end{aligned}
\]
Substitution into \Cref{thm:equilibrium-norm} gives both identities in
\eqref{eq:nqm-two-channel-nonzero-mean}.

\emph{Step 2: Show that $\bar R^2$ strictly increases with $\alpha$.}
We differentiate its two contributions while the input statistics remain fixed:
\[
 \frac{\dd\bar R^2}{\dd\alpha}
 =\frac{\lr^2}{1-\delta^2}
 \left[\sigma_a^2\frac{\dd F_\delta[e]}{\dd\alpha}
       +2\frac{1+\delta}{1-\delta}\norm{\bm\mu}^2(1+\alpha)\right]>0.
\]
Here $\dd F_\delta[e]/\dd\alpha>0$ by \Cref{thm:ademamix-lever}, and at
least one of $\sigma_a^2$ and $\norm{\bm\mu}^2$ is positive.

\emph{Step 3: Evaluate $\qbarR$ when $\sigma_a^2=0$.}
In this case $\norm{\bm\mu}^2>0$, so the common mean-power factor cancels
from the relative-update formula:
\[
 \qbarR^2
 =(1-\delta^2)
 \frac{\norm{\bm\mu}^2(1+\alpha)^2}
      {\frac{1+\delta}{1-\delta}\norm{\bm\mu}^2(1+\alpha)^2}
 =(1-\delta^2)\frac{1-\delta}{1+\delta}
 =(1-\delta)^2.
\]
Since $0<\delta<1$, this gives $\qbarR=1-\delta$.

\emph{Step 4: Show that $\qbarR^2$ strictly decreases when $\sigma_a^2>0$.}
We differentiate the expression for $\qbarR^2$ in
\eqref{eq:nqm-two-channel-nonzero-mean} and determine the sign of each
contribution. The quotient rule, followed by collecting terms, gives
\begin{equation}
\begin{aligned}
 \frac{\dd\qbarR^2}{\dd\alpha}
 &=(1-\delta^2)
   \frac{\sigma_a^2F_\delta[e]}
        {\sigma_a^2F_\delta[e]
           +\frac{1+\delta}{1-\delta}\norm{\bm\mu}^2(1+\alpha)^2}
   \frac{\dd}{\dd\alpha}\frac{P[e]}{F_\delta[e]}\\
 &\quad+
   \frac{(1-\delta^2)\sigma_a^2\norm{\bm\mu}^2(1+\alpha)^4}
        {\left[\sigma_a^2F_\delta[e]
           +\frac{1+\delta}{1-\delta}\norm{\bm\mu}^2(1+\alpha)^2\right]^2}
   \left(\frac{1+\delta}{1-\delta}\frac{P[e]}{F_\delta[e]}-1\right)
   \frac{\dd}{\dd\alpha}\frac{F_\delta[e]}{(1+\alpha)^2}.
\end{aligned}
\label{eq:nqm-mean-direct-derivative}
\end{equation}
The contribution containing
$\frac{\dd}{\dd\alpha}(P[e]/F_\delta[e])$ is strictly negative by
\Cref{thm:ademamix-lever}, since its multiplier is positive for
$\sigma_a^2>0$. To show that the contribution proportional to
$\norm{\bm\mu}^2$ is nonpositive, we establish the signs of its remaining
factors in (a) and (b).

\emph{(a) Show that $\frac{1+\delta}{1-\delta}P[e]/F_\delta[e]-1\geq0$.}
This makes the expression in parentheses in
\eqref{eq:nqm-mean-direct-derivative} nonnegative. We bound $\Gamma_e(h)$
by $P[e]$ using Cauchy--Schwarz:
\[
\begin{aligned}
 |\Gamma_e(h)|
 &=\left|\sum_{n\geq0}e_ne_{n+h}\right|\\
 &\leq\left(\sum_{n\geq0}e_n^2\right)^{1/2}
       \left(\sum_{n\geq0}e_{n+h}^2\right)^{1/2}
 \leq P[e],
\end{aligned}
\]
because $\sum_{n\geq0}e_{n+h}^2$ omits the first $h$ nonnegative terms of
$\sum_{n\geq0}e_n^2$. Substitution into the definition of $F_\delta[e]$ gives
$F_\delta[e]\leq\frac{1+\delta}{1-\delta}P[e]$, and hence
\begin{equation}
 \frac{1+\delta}{1-\delta}\frac{P[e]}{F_\delta[e]}-1\geq0.
\label{eq:nqm-mean-correlation-bound}
\end{equation}

\emph{(b) Show that $\frac{\dd}{\dd\alpha}\big(F_\delta[e]/(1+\alpha)^2\big)<0$.}
Together with (a), this will make the contribution proportional to
$\norm{\bm\mu}^2$ in \eqref{eq:nqm-mean-direct-derivative} nonpositive.
To determine this derivative, expand $F_\delta[e]$ using
\eqref{eq:nqm-branch-forcing-decomposition}. We need to compare its three
pairings as a fast buffer is replaced by a slow one. Subtracting the closed
forms in \eqref{eq:nqm-branch-forcing-geometric} yields
\[
\begin{aligned}
 &\frac{\Phi_\delta(\beta^{(f)},\beta^{(f)})
             -\Phi_\delta(\beta^{(f)},\beta^{(s)})}
            {\Phi_0(\beta^{(f)},\beta^{(f)})}
 =\frac{\Phi_\delta(\beta^{(f)},\beta^{(s)})
             -\Phi_\delta(\beta^{(s)},\beta^{(s)})}
            {\Phi_0(\beta^{(s)},\beta^{(s)})}\\
 &\qquad=
 \frac{(\beta^{(s)}-\beta^{(f)})(1-\delta)
                  (1+\delta\beta^{(f)}\beta^{(s)})}
      {(1-\beta^{(f)}\beta^{(s)})
       (1-\delta\beta^{(f)})(1-\delta\beta^{(s)})}>0.
\end{aligned}
\]
Every factor in the last expression is positive under the stated assumptions.
Thus
$\Phi_\delta(\beta^{(f)},\beta^{(f)})
 >\Phi_\delta(\beta^{(f)},\beta^{(s)})
 >\Phi_\delta(\beta^{(s)},\beta^{(s)})$.
Differentiating the quadratic in
\eqref{eq:nqm-branch-forcing-decomposition} now gives
\begin{equation}
\begin{aligned}
 \frac{\dd}{\dd\alpha}\frac{F_\delta[e]}{(1+\alpha)^2}
 &=\frac{2}{(1+\alpha)^3}\Big[
       \Phi_\delta(\beta^{(f)},\beta^{(s)})
         -\Phi_\delta(\beta^{(f)},\beta^{(f)})\\
 &\qquad+\alpha\big(
       \Phi_\delta(\beta^{(s)},\beta^{(s)})
         -\Phi_\delta(\beta^{(f)},\beta^{(s)})\big)\Big]<0.
\end{aligned}
\label{eq:nqm-mean-normalized-forcing}
\end{equation}
Both differences are negative, proving (b).

By (a) and (b), the contribution proportional to $\norm{\bm\mu}^2$ in
\eqref{eq:nqm-mean-direct-derivative} is nonpositive. The contribution
containing $\frac{\dd}{\dd\alpha}(P[e]/F_\delta[e])$ is strictly negative.
Therefore $\dd\qbarR^2/\dd\alpha<0$. Since $\qbarR>0$,
$\qbarR$ itself also strictly decreases.
\end{proof}

\subsection{Weight-Decay Contraction Limits Useful Memory}
\label{app:proofs-limits}
We hold the slow amplitude finite and ask what happens as the slow pole approaches
one. Its total mass remains $\alpha$, but each tap becomes small and the memory
extends beyond the weight-decay timescale $\tw=1/(1-\delta)$.

\begin{proposition}{Extreme Reach Returns to Fast-Only}{finite-survival-window}
Fix $0<\delta<1$, $0\leq\beta^{(f)}<1$, and $0<\alpha<\infty$. As $\beta^{(s)}\to1^-$,
\begin{equation}
 \begin{aligned}
    P[e^{\mathrm{AMix}}]&\to\Phi_0(\beta^{(f)},\beta^{(f)}),\qquad
    F_\delta[e^{\mathrm{AMix}}]\to\Phi_\delta(\beta^{(f)},\beta^{(f)}),\\
    \frac{P[e^{\mathrm{AMix}}]}{F_\delta[e^{\mathrm{AMix}}]}&\to
    \frac{\Phi_0(\beta^{(f)},\beta^{(f)})}
         {\Phi_\delta(\beta^{(f)},\beta^{(f)})}.
 \end{aligned}
    \label{eq:nqm-extreme-reach-return}
\end{equation}
The ratio equals fast-only at coincident poles, falls for every strictly ordered finite pole,
and returns to fast-only at extreme reach; continuity therefore gives at least one interior
maximum of its reduction.
\end{proposition}
\begin{proof}[Proof of \Cref{prop:finite-survival-window}]
\emph{Step 1 (the mixed terms vanish).}
At \(\delta=0\),
\begin{equation*}
    \Phi_0(\beta^{(f)},\beta^{(s)})
    =\frac{(1-\beta^{(f)})(1-\beta^{(s)})}
           {1-\beta^{(f)}\beta^{(s)}}
    \longrightarrow0
    \qquad\text{as }\beta^{(s)}\to1^-.
\end{equation*}
For \(0<\delta<1\), the exact mixed pairing is
\begin{equation*}
    \Phi_\delta(\beta^{(f)},\beta^{(s)})
    =\frac{(1-\beta^{(f)})(1-\beta^{(s)})
           (1-\delta^2\beta^{(f)}\beta^{(s)})}
          {(1-\beta^{(f)}\beta^{(s)})
           (1-\delta\beta^{(f)})
           (1-\delta\beta^{(s)})}.
\end{equation*}
Its numerator contains \(1-\beta^{(s)}\), while its denominator tends to
\begin{equation*}
    (1-\beta^{(f)})(1-\delta\beta^{(f)})(1-\delta)>0.
\end{equation*}
Therefore \(\Phi_\delta(\beta^{(f)},\beta^{(s)})\to0\).

\emph{Step 2 (the slow--slow terms vanish).}
The diagonal formulas give
\begin{equation*}
    \Phi_0(\beta^{(s)},\beta^{(s)})
    =\frac{1-\beta^{(s)}}{1+\beta^{(s)}}
    \longrightarrow0
\end{equation*}
and
\begin{equation*}
    \Phi_\delta(\beta^{(s)},\beta^{(s)})
    =\frac{1-\beta^{(s)}}{1+\beta^{(s)}}
     \frac{1+\delta\beta^{(s)}}{1-\delta\beta^{(s)}}
    \longrightarrow0.
\end{equation*}
The second limit uses \(1-\delta>0\).

\emph{Step 3 (return to the \(\alpha=0\) fast-branch functionals).}
Substituting the four limits from Steps 1 and 2 into
\eqref{eq:nqm-two-channel-power} and
\eqref{eq:nqm-branch-forcing-decomposition} leaves only their fast--fast terms. Since
\(\Phi_\delta(\beta^{(f)},\beta^{(f)})>0\), passage to the quotient is valid. This proves
\eqref{eq:nqm-extreme-reach-return}.

\emph{Step 4 (the interior maximum).}
At \(\beta^{(s)}=\beta^{(f)}\), both quadratics acquire the same factor:
\begin{equation*}
    P[e^{\mathrm{AMix}}]=(1+\alpha)^2\Phi_0(\beta^{(f)},\beta^{(f)}),
    \qquad
    F_\delta[e^{\mathrm{AMix}}]=(1+\alpha)^2\Phi_\delta(\beta^{(f)},\beta^{(f)}).
\end{equation*}
Their ratio is therefore the fast-only value. For every
\(\beta^{(f)}<\beta^{(s)}<1\), \Cref{thm:ademamix-lever} shows that increasing \(\alpha\) from zero to the
fixed positive value strictly decreases the ratio. The closed form for every pairing is continuous
in \(\beta^{(s)}\) on \([\beta^{(f)},1)\), and Step 3 supplies a continuous extension at \(1\). The decrease
below the fast-only value is the kernel-ratio reduction. It is continuous, vanishes at
\(\beta^{(s)}=\beta^{(f)}\) and in the limit
\(\beta^{(s)}\to1^-\), and is positive between them. The extreme-value theorem places at least one of
its maxima in the interior.
\end{proof}

For every slow pole in this comparison, suppose that the input has zero mean and is
uncorrelated across steps, and that its modeled update has a finite-second-moment radius-stationary
response. Then \(\qbarR^2=(1-\delta^2)P[e^{\mathrm{AMix}}]/F_\delta[e^{\mathrm{AMix}}]\), so the same interior window gives a
modeled moment-ratio decrease relative to fast-only.

\clearpage
\section{Controlled Validation of the Mechanism}
\label{app:controlled}

The tangent-noise random walk gives an exact fixed-power test of the memory model. The scale-invariant MLP tests its norm predictions with autograd gradients and a live RMS preconditioner.

\subsection{Noisy-Sphere Testbed and Kernel Tests}
\label{app:controlled-sphere}

\paragraph{Testbed and Inputs.}
At each step, draw $\bxi_t\sim\mathcal N(\bzero,\sigma^2\bI)$ independently of the past and project it onto the tangent space at $\xhat_t=\bx_t/\norm{\bx_t}$:
\begin{equation}
 \ba_t=(\bI-\xhat_t\xhat_t^\top)\bxi_t.
 \label{eq:nqm-norm-noise-projection}
\end{equation}
Conditional on the past, $\ba_t$ has zero mean, so inputs at distinct times are uncorrelated. The projection has rank $d-1$, giving constant input power
\begin{equation}
 C_a(0)=\E[\norm{\ba_t}^2\mid\bx_t]
 =\sigma^2\operatorname{tr}(\bI-\xhat_t\xhat_t^\top)
 =(d-1)\sigma^2,
 \label{eq:nqm-norm-projected-power}
\end{equation}
independent of the radius. The experiment uses these noise inputs alone.

\paragraph{Optimizer and Reporting.}
We run \textsc{SGD2mW}, the raw two-EMA counterpart of the \textsc{AdEMAMix} update in \Cref{sec:ademamix} with decoupled weight
decay and no coordinatewise RMS normalization,
\[
    \bx_{t+1}=(1-\lr\lam)\bx_t
    -\lr\big(\bmom^{(f)}_{t+1}+\alpha\bmom^{(s)}_{t+1}\big).
\]
The representation of \Cref{ass:tangent-execution-stream} is then exact with
\(\ba_i=(\bI-\xhat_i\xhat_i^\top)\bxi_i\), and \eqref{eq:nqm-norm-noise-projection} makes each input tangent at its own
weight. Define the weight-decay timescale $\tw=1/(\lr\lam)$. The default fixed-power run uses \(d=256\), \(\sigma=1\), \(\lr=0.01\),
\(\lam=0.03\), \(\beta^{(f)}=0.9\), \(T=12\tw\), and \(\tw=1/(\lr\lam)=3333\).

From a unit-radius start, the squared radius climbs before settling. Each constant-configuration replica is
averaged over the final \(20\%\) of optimizer steps, then five replica estimates are combined.

\paragraph{Radial Counterfactuals.}
To identify the radial mechanism across the \(\alpha\) sweep, panel (a) of \Cref{fig:sphere} compares
three update rules. The \textbf{full} update is defined above. The \textbf{power-only} control
preserves its instantaneous length but redraws its direction with an independent uniform unit vector
\(\bw_t\),
\[
    \bx_{t+1}=(1-\lr\lam)\bx_t-\lr\norm{\bu_t}\bw_t,
\]
destroying the persistent directional alignment. The \textbf{radial-off} control preserves decoupled
decay and projects only the optimizer update,
\begin{equation}
    \bx_{t+1}=(1-\lr\lam)\bx_t
    -\lr\big(\bI-\xhat_t\xhat_t^\top\big)\bu_t .
    \label{eq:nqm-norm-update-projection}
\end{equation}
The projection removes the radial cross term and radial update power, retaining tangential chord growth.

Both interventions produce smaller settled radii than the full update; power only stays above radial off.

\paragraph{Moment Ratio and Pure-Gain Control.}
The slow-channel radius increase is accompanied by a lower moment
ratio. At \(\beta^{(s)}=0.99\), where the slow reach is short relative to the weight-decay timescale, the derived \(\qbarR\) and its
empirical tail estimate fall as \(\alpha\) grows and follow the exact model curve with weight decay
\eqref{eq:nqm-two-channel-qbar} under the ordered pole conditions of
\eqref{eq:nqm-two-channel-qbar-derivative}. The same-pole $\beta^{(s)}=\beta^{(f)}$ control stays flat. Pure gain multiplies both $P[e]$ and
$F_\delta[e]$ by the same squared factor, leaving their ratio unchanged. The reduction with an
ordered slow pole comes from its longer memory.

\subsection{Scale-Invariant MLP Bridge}
\label{app:controlled-mlp}

\paragraph{Architecture and Scale Invariance.}
A small MNIST classifier is trained in PyTorch with ordinary autograd gradients. The hidden blocks
are bias-free linear layers followed by non-affine, epsilon-free root mean square layer normalization (RMSNorm) and a rectified linear unit (ReLU), and a fixed
readout scale \(s_{\mathrm{out}}\) sets the logit magnitude:
\[
    \begin{aligned}
    \bz_1&=\operatorname{ReLU}\!\left(\operatorname{RMSNorm}(\bTheta_1\bz_{\mathrm{in}})\right),\\
    \bz_2&=\operatorname{ReLU}\!\left(\operatorname{RMSNorm}(\bTheta_2\bz_1)\right),\\
    f(\bz_{\mathrm{in}})&=s_{\mathrm{out}}\,\bTheta_3\bz_2.
    \end{aligned}
\]
For a nonzero row vector \(\bv\in\mathbb R^{n}\), the normalization used here is
\[
    \operatorname{RMSNorm}(\bv)
    =\frac{\bv}{\sqrt{n^{-1}\norm{\bv}^2}},
\]
with a zero-row convention at \(\bv=\bzero\). Here \(\bz_{\mathrm{in}}\) is the flattened MNIST input,
\(\bz_1,\bz_2\) are hidden activations, and \(\bTheta_1,\bTheta_2,\bTheta_3\)
are the weight matrices. RMSNorm divides out positive scale, so the
loss is invariant to positive scalar rescaling of either nonzero hidden matrix:
\(L(\kappa\bTheta_\ell)=L(\bTheta_\ell)\) for \(\kappa>0\) and \(\ell\in\{1,2\}\)
(\Cref{sec:si-geometry}). The optimizer denominator also uses \(\eps=0\), which preserves
the same positive-rescaling symmetry exactly. The readout scale is fixed at \(s_{\mathrm{out}}=4\) and never
trained.

\paragraph{Hyperparameters.}
We use bias-free linear maps, hidden width \(128\), minibatches of \(32\), and \(3000\) steps over five
seeds, at constant \(\lr=3\times10^{-3}\), \(\lam=0.3\) (so \(\tw\approx1111\) and the horizon is
\(\approx2.7\,\tw\)), \(\beta^{(f)}=0.9\), shared RMS preconditioner with \(\beta_2=0.99\) and \(\eps=0\),
and \(\beta^{(s)}=0.99\). Each seed draws its own \(12{,}000\)-example training subset from the
\(60\)k MNIST pool, so the dispersion bands reflect data as well as initialization and minibatch order. Time-series bands and endpoint error bars are \(\pm1\sigma\) across seeds.

\paragraph{Norm Response.}
We compare hidden-matrix squared norms across $\alpha\in\{0,1,2,5,10\}$.
The finite-radius recursion of \Cref{app:scheduling-forward} starts from the observed
radius. The $\alpha=0$ endpoint calibrates one input-power scale, held fixed for
every other amplitude and all plotted times. The model reproduces the ordering of
the norm trajectories and their settled tails, while agreement is weaker during
warmup. This tests the norm effect with live preconditioning.

\begin{figure*}[t]
  \centering
  \begin{subfigure}[t]{.48\textwidth}
  \includegraphics[width=\linewidth]{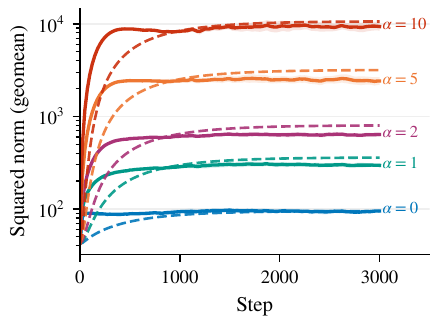}\caption{}
  \end{subfigure}\hfill
  \begin{subfigure}[t]{.48\textwidth}
  \includegraphics[width=\linewidth]{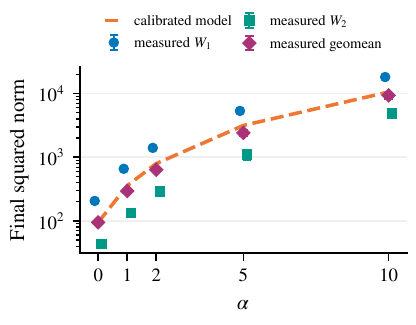}\caption{}
  \end{subfigure}
  \caption{Squared norms in the scale-invariant MNIST MLP, over five seeds. (a) Geometric mean across the two hidden matrices: solid curves show measurements with one-standard-deviation bands on the log scale, and dashed curves show the finite-history model. (b) Final norms of each matrix and their geometric mean, with one-standard-deviation bars. One input-power scale is calibrated at $\alpha=0$ and reused for all amplitudes.}
  \label{fig:scale-invariant-mlp}
\end{figure*}

\clearpage
\section{Rotational-Scheduling Details}
\label{app:scheduling}

Throughout this appendix, we use the one-based coefficient indexing $s_1,\ldots,s_T$ of the finite-horizon bound \citep{graldi2026rotational}. The main text instead indexes coefficients by the zero-based optimizer step $t$, so its $s_t$ corresponds to $s_{t+1}$ here.

\subsection{The Last-Iterate Bound}
\label{app:scheduling-criterion}

Following \citet[Thm.~2.3]{graldi2026rotational}, let $L$ be
differentiable and geodesically convex on a fixed open spherical cap of angular radius
$0<\thetacap\leq\pi/2$, with a minimizer $\xhat^\star$ inside the cap. Let all iterates through $\xhat_{T+1}$ stay in the cap,
and let a fresh tangent oracle have conditional mean $\nabla_{\Sph}L(\xhat_t)$ and satisfy $\E\norm{\gtilde_t}^2\leq G^2$ for a common deterministic $G>0$. The coefficients $s_1,\ldots,s_T$ are positive, deterministic, and fixed
independently of the oracle draws. Their normalized tangent step is
\begin{equation}
    \xhat_{t+1}=\frac{\xhat_t-s_t\gtilde_t}{\sqrt{1+s_t^2\norm{\gtilde_t}^2}},
    \qquad 1\leq t\leq T.
    \label{eq:finite-horizon-tangent-step}
\end{equation}
The last-iterate bound at $\xhat_T$ uses $T-1$
executed updates and an auxiliary final coefficient $s_T$.

Write $M=\sum_{t=1}^Ts_t$, $\omega_t=s_t/M$, and
$D=\arccos\ip{\xhat_1}{\xhat^\star}$. The bound is
\begin{equation}
    \begin{aligned}
    \E[L(\xhat_T)-L(\xhat^\star)]
    &\leq\frac{D^2}{2M}+\frac{(1+\thetacap)G^2}{2}M C_T(\omega),\\
    C_T(\omega)&:=\omega_T+
      \sum_{r=1}^{T-1}\frac{\omega_r^2}{\sum_{t=r+1}^T\omega_t}.
    \end{aligned}
    \label{eq:shape-score}
\end{equation}
At a fixed normalized profile, optimizing its overall coefficient sum gives \citep[Prop.~2.4]{graldi2026rotational}
\[
 \min_{M>0}\left[\frac{D^2}{2M}
       +\frac{(1+\thetacap)G^2}{2}M C_T(\omega)\right]
 =DG\sqrt{(1+\thetacap)C_T(\omega)}.
\]
With common $D$, $G$, and $\thetacap$, ratios of these optimized bounds are therefore ratios of $\sqrt{C_T}$. The use of the bound for LLM scheduling is motivated by the empirical comparisons in \citet[Sec.~4]{graldi2026rotational}.

\subsection{Forward Construction: From Optimizer Memory to a Rotational Coefficient}
\label{app:scheduling-forward}

\paragraph{The Inherited Input Model.}
Let the learning rates \(\lr_t>0\), decay coefficients \(\lam_t\geq0\), and execution
coefficients \(E_{t,i}\) be deterministic. Set
\begin{equation}
 \delta_t:=1-\lr_t\lam_t\in(0,1],\qquad
 \widehat{\bu}_t=\sum_{i=0}^tE_{t,i}\ba_i,\qquad
 \bx_{t+1}^{\mathrm{rep}}=\delta_t\bx_t^{\mathrm{rep}}-\lr_t\widehat{\bu}_t.
 \label{eq:represented-process}
\end{equation}
The parameter starts from fixed \(\bx_0:=\bx_0^{\mathrm{rep}}\ne\bzero\), and the optimizer
buffers start from zero. Use the zero-mean, uncorrelated inputs of \Cref{ass:white-noise}, with positive power:
\begin{equation}
 \E\ba_i=\bzero,\qquad
 \E\ip{\ba_i}{\ba_\ell}
 =\begin{cases}C_a(0),&i=\ell,\\0,&i\ne\ell,\end{cases}
 \qquad C_a(0)>0.
 \label{eq:represented-source}
\end{equation}
These input moments determine the finite recursion \eqref{eq:finite-radius-recursion}, used for the MLP overlay in \Cref{app:controlled-mlp} and the scheduling simulation below.

The input power $C_a(0)$ is fixed as the radius changes and matched across compared policies. This is the linear model of \Cref{sec:memory-correlation}; the scheduling surrogate below supplies fresh tangent directions separately from the remembered updates used to compute the radius.

\paragraph{Momentum Coefficients and Weight Decay.}
Specializing the EMA coefficients of \Cref{def:execution-kernel} to the zero-start two-buffer policies gives
\begin{equation}
 E_{t,i}=(1-\beta^{(f)})(\beta^{(f)})^{\,t-i}
 +\alpha_t(1-\beta^{(s)}_{i})\prod_{\ell=i+1}^{t}\beta^{(s)}_{\ell},
 \qquad 0\leq i\leq t.
 \label{eq:execution-coefficients}
\end{equation}
The first term is the fast buffer. The second is the slow buffer, multiplied by its
\emph{execution-time} amplitude \(\alpha_t\), rather than the amplitude when the input was recorded. Setting
\(\alpha_t=0\) gives single momentum. With constant slow coefficients the second term is
\(\alpha(1-\beta^{(s)})(\beta^{(s)})^{t-i}\). An empty product is one. These simulations use raw EMAs of the fixed-power inputs.

After an update is applied, weight decay contracts its displacement. We use the cumulative contraction $W_{i,t}$ defined in \Cref{def:survival-factor}. An update applied at $r$ enters $\bx_{r+1}^{\mathrm{rep}}$, so its displacement at step $t$ is multiplied by $W_{r+1,t}$.

\paragraph{Overlap and Instantaneous Power.}
Two modeled updates can correlate because the same input occurs in both momentum histories. Define their finite kernel inner product and its
diagonal by
\begin{equation}
 \Gamma_{t,r}:=\sum_{i=0}^{\min(t,r)}E_{t,i}E_{r,i},
 \qquad P_t:=\Gamma_{t,t}=\sum_{i=0}^tE_{t,i}^2.
\end{equation}
To see their meaning, expand the inner product before taking its expectation:
\begin{equation}
 \begin{aligned}
 \E\ip{\widehat{\bu}_t}{\widehat{\bu}_r}
 &=\sum_{i=0}^t\sum_{\ell=0}^r
 E_{t,i}E_{r,\ell}\E\ip{\ba_i}{\ba_\ell}\\
 &=C_a(0)\sum_{i=0}^{\min(t,r)}E_{t,i}E_{r,i}
 =C_a(0)\Gamma_{t,r}.
 \end{aligned}
 \label{eq:finite-update-correlation}
\end{equation}
In particular,
\begin{equation}
 \E\norm{\widehat{\bu}_t}^2=C_a(0)P_t.
 \label{eq:finite-update-power}
\end{equation}
We require \(P_t>0\) at every transferred step; the EMA policies above satisfy this condition.

\paragraph{Temporal Correlations in the Norm Recursion.}
Instantaneous power alone does not determine the radius. Define
\begin{equation}
 R_t:=\sqrt{\E\norm{\bx_t^{\mathrm{rep}}}^2}.
 \label{eq:represented-radius}
\end{equation}
To identify the missing historical contribution, first unroll the parameter recursion:
\begin{equation}
 \bx_t^{\mathrm{rep}}
 =W_{0,t}\bx_0-\sum_{r=0}^{t-1}W_{r+1,t}\lr_r\widehat{\bu}_r.
 \label{eq:finite-parameter-unrolling}
\end{equation}
The fixed initialization has zero expected overlap with the centered update. Pairing the
remaining terms with \(\widehat{\bu}_t\) and using
\eqref{eq:finite-update-correlation} therefore gives
\begin{equation}
 \E\ip{\bx_t^{\mathrm{rep}}}{\widehat{\bu}_t}
 =-C_a(0)\sum_{r=0}^{t-1}W_{r+1,t}\lr_r\Gamma_{t,r}.
 \label{eq:finite-radial-overlap}
\end{equation}
The sign has a physical meaning. Earlier updates displaced the parameter by
\(-\lr_r\widehat{\bu}_r\). Positive temporal correlation aligns the current update with earlier updates, and hence against the displacement they left in the parameter. Because the applied step
is \(-\lr_t\widehat{\bu}_t\), this negative parameter--update overlap does outward radial
work on average.

Next expand one squared parameter step, as in \eqref{eq:nqm-norm-growth-compact}:
\begin{equation}
 R_{t+1}^2=\delta_t^2R_t^2
 -2\delta_t\lr_t\E\ip{\bx_t^{\mathrm{rep}}}{\widehat{\bu}_t}
 +\lr_t^2\E\norm{\widehat{\bu}_t}^2.
 \label{eq:finite-radius-balance}
\end{equation}
Combining the instantaneous-power and temporal-correlation terms gives
\begin{equation}
 F_t:=P_t+2\sum_{r=0}^{t-1}
 \frac{\delta_t\lr_r}{\lr_t}W_{r+1,t}\Gamma_{t,r}.
 \label{eq:finite-forcing}
\end{equation}
Here $P_t$ is the instantaneous-power factor, and the sum accounts for temporal correlations weighted by learning rates and weight decay. The exact finite
radius recursion is consequently
\begin{equation}
 R_{t+1}^2=\delta_t^2R_t^2+\lr_t^2C_a(0)F_t.
 \label{eq:finite-radius-recursion}
\end{equation}
Under a learning-rate schedule, \(F_t\) includes historical rate ratios as well as kernel
overlap and decay.

For the nonnegative EMA coefficients in \eqref{eq:execution-coefficients},
\(\Gamma_{t,r}\geq0\) and \(F_t\geq P_t>0\). The additional squared-radius contribution is then nonnegative.

\paragraph{Accumulating the Norm Contributions.}
Iterating \eqref{eq:finite-radius-recursion} gives the conceptual form of the radius history:
\begin{equation}
 R_t^2=W_{0,t}^2\norm{\bx_0}^2+
 C_a(0)\sum_{r=0}^{t-1}W_{r+1,t}^2\lr_r^2F_r.
 \label{eq:finite-radius-unrolled}
\end{equation}
The first term is the contracted initialization. Each later squared-radius contribution is contracted by subsequent weight decay.
This is why the radius can keep changing after the instantaneous-power factor has almost settled.
For the nonnegative EMA policies, $F_r>0$, so the same identity gives
$R_t^2\ge W_{0,t}^2\norm{\bx_0}^2>0$. Thus the coefficient below has a positive radius denominator.

\paragraph{The Finite Relative-Update Scale.}
The finite-time analogue of $\qbarR$ is $\lr_t\sqrt{C_a(0)P_t}/R_t$, with the update power and squared radius averaged before taking their root ratio.

\subsubsection{From Modeled Radius to Rotational Coefficients}
To isolate the scalar scheduling effect of memory, we supply the deterministic
radius \(R_t\) and replace the remembered update direction by the fresh tangent
oracle of \Cref{app:scheduling-criterion}, scaled by \(\sqrt{P_t}\).
The oracle has its own second-moment bound $G^2$, as specified in \Cref{app:scheduling-criterion}.

At optimizer step \(t\), form the proposal along the surrogate direction
\(\xhat_{t+1}\) and factor its positive contracted radius:
\begin{equation}
 \begin{aligned}
 &\delta_tR_t\xhat_{t+1}-\lr_t\sqrt{P_t}\gtilde_{t+1}\\
 &\qquad=\delta_tR_t
 \left(\xhat_{t+1}-\frac{\lr_t\sqrt{P_t}}{\delta_tR_t}\gtilde_{t+1}\right).
 \end{aligned}
 \label{eq:factored-tangent-proposal}
\end{equation}
Unit normalization cancels \(\delta_tR_t>0\). Because the oracle is tangent,
the remaining proposal has the normalized form \eqref{eq:finite-horizon-tangent-step}
with coefficient
\begin{equation}
 s_{t+1}=\frac{\lr_t\sqrt{P_t}}{\delta_tR_t}.
 \label{eq:transferred-schedule-app}
\end{equation}
The single inverse-radius factor comes from
normalizing the displacement. A raw scale-invariant gradient contributes an
additional inverse-radius factor through \eqref{eq:grad-homogeneity-minus-one}.

\subsection{Evaluating the Bound for Constant-Rate Training}
\label{app:scheduling-application}

For \Cref{fig:scheduling-application}, we generate the coefficients of \eqref{eq:transferred-schedule-app} over $T=189{,}000$ updates, starting from $R_0=10^{-3}$ and zero momentum buffers. The input power is $C_a(0)=1$, weight decay is $\lambda=0.1$, and fast retention is $\beta^{(f)}=0.85$. Both models use the LLM's 3,000-update LR warmup followed by a constant rate $8\cdot10^{-4}$. The single-buffer model has $\alpha_t=0$; the two-buffer model uses the full-horizon slow-memory schedules of \Cref{app:llm-protocol}, with targets $\alpha=12$ and $\beta^{(s)}=0.9999$.

We retain the radius and buffer history accumulated during warmup, then normalize the $N=186{,}000$ post-warmup coefficients by their sum. For each profile $\omega$, the optimal overall scale gives
\[
 s_t^\star=\frac{D\omega_t}{G\sqrt{(1+\thetacap)C_N(\omega)}}.
\]
Panel (a) follows \citet[App.~C.4]{graldi2026rotational} and removes the common problem-dependent scale from these coefficients:
\[
 \frac{G\sqrt{1+\thetacap}}{D}s_t^\star
 =\frac{\omega_t}{\sqrt{C_N(\omega)}}.
\]
Panel (b) evaluates \eqref{eq:shape-score} on successive prefixes of the same $s_t^\star$, keeping the final-horizon scale fixed. Dividing the bound by $DG\sqrt{1+\thetacap}$ gives the displayed normalized value, whose endpoint is $\sqrt{C_N(\omega)}$.

\subsection{Bound Trajectories after Slow-Memory Removal}
\label{app:scheduling-switch}

At each cut, we set $\alpha_t=0$ and continue from the existing weights and buffers. We evaluate \eqref{eq:shape-score} on each full post-warmup prefix at its modeled scale and divide by $DG\sqrt{1+\thetacap}$, following \citet[App.~C.4]{graldi2026rotational}.

\paragraph{Cosine LR.}
For \Cref{fig:switch}b and the complete comparison in \Cref{fig:switch-cosine-all}b, we use cosine LR with peak $8\cdot10^{-4}$ and the zero-mean input model of \Cref{app:scheduling-forward}. AdamW uses only the fast buffer. All curves share $\thetacap=\pi/4$, $D=\thetacap/2$, and $G=1$. The main figure magnifies the 50k transient; the complete comparison includes an inset for each cut. Each inset shows the percentage difference between the cut and uninterrupted bounds at the same update.

\paragraph{Constant LR.}
For \Cref{fig:switch-constant}b, we use the same zero-mean input model with constant LR $10^{-4}$ after warmup, $D=G=1$, and $\thetacap=\pi/4$. The bound briefly dips below the uninterrupted curve after each switch; after early cuts, it rises again later in training. 

\subsection{Additional Comparisons with Tuned Learning Rates}
\label{app:scheduling-controlled}

We evaluate the right-hand side of \eqref{eq:shape-score} on the deterministic
coefficients supplied by the finite-time model. The computation uses
$T=189{,}000$, $R_0=10^{-3}$, $\lambda=0.1$, $C_a(0)=G^2=1$,
$\thetacap=(1-10^{-9})\pi/4$, and $D=\thetacap/2$.

All policies use zero-start raw EMAs with $\beta^{(f)}=0.85$. Single momentum has
$\alpha_t=0$. Fixed two-timescale memory has $\alpha_t=12$ and
$\beta^{(s)}_{t}=0.9999$. Scheduled two-timescale memory uses
$\alpha_t=12(t+1)/T$ and interpolates $\beta^{(s)}_{t}$ from $0.85$ to $0.9999$
at fraction $(t+1)/T$ linearly in EMA half-life. For each peak rate, the learning rate is either constant or follows cosine decay to one percent of its peak over the full horizon.
For each policy and rate family, the search evaluates 81 logarithmically spaced
peaks in $[10^{-10},10^{-2}]$, followed by 41 logarithmically spaced peaks between
the coarse neighbors of the best coarse candidate. Every candidate regenerates
its own radius history and coefficients. Selection minimizes the final bound
value, including the auxiliary final coefficient in \eqref{eq:shape-score}.

Under constant rates, both fixed and scheduled slow memory lower the best tested score. Under cosine, both raise it. These finite-grid comparisons support the qualitative expectation that endogenous cooling is more useful when explicit learning-rate decay is absent.

\begin{figure}[htbp]
\centering
\includegraphics[width=.3\linewidth]{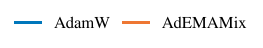}\par\smallskip
\begin{subfigure}[t]{.4\linewidth}
\includegraphics[width=\linewidth]{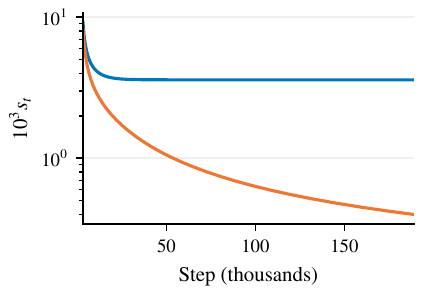}\caption{Constant model}
\end{subfigure}\hfill
\begin{subfigure}[t]{.4\linewidth}
\includegraphics[width=\linewidth]{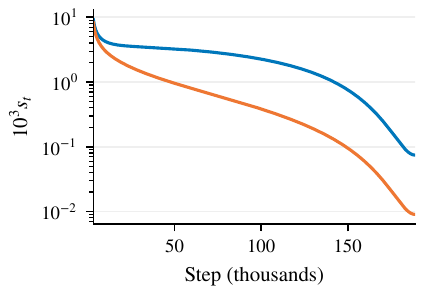}\caption{Cosine model}
\end{subfigure}
\par\smallskip
\begin{subfigure}[t]{.4\linewidth}
\includegraphics[width=\linewidth]{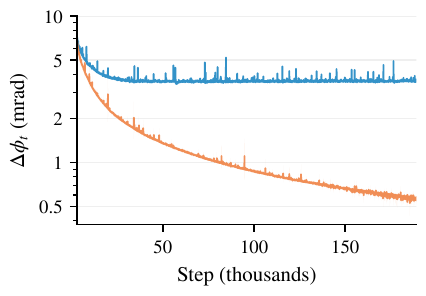}\caption{Constant LLM}
\end{subfigure}\hfill
\begin{subfigure}[t]{.4\linewidth}
\includegraphics[width=\linewidth]{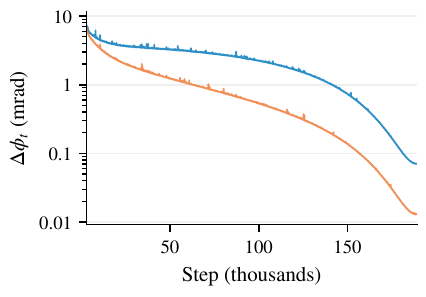}\caption{Cosine LLM}
\end{subfigure}
\caption{Modeled rotational coefficients (a,b) and measured LLM angles (c,d) under constant (a,c) and cosine (b,d) LR. The rows use separate vertical scales to compare the temporal profiles. Settings are in \Cref{app:scheduling-curve-settings}.}
\label{fig:generated}
\end{figure}

\paragraph{Settings for the Model--LLM Comparison.}
\label{app:scheduling-curve-settings}
\Cref{fig:generated} uses the same finite model with $R_0=10^{-3}$, $C_a(0)=1$, $\lambda=0.1$, $\beta^{(f)}=0.85$, and the LLM's scheduled slow coefficients. Both optimizer models use peak LR $8\cdot10^{-4}$, with the 3,000-step warmup and constant or cosine schedule of \Cref{app:llm-protocol}. Curves start at step 3050. Input power is not fitted to the LLM measurements.

\clearpage
\section{LLM Experimental Details and Supplementary Results}
\label{app:llm-experiments}

\subsection{Training, Tuning, and Measurement Protocol}
\label{app:llm-protocol}

\paragraph{Model Architecture and Pretraining Data.}
We use the pretraining codebase of \citet{semenov2025benchmark} for our LLM experiments.
All pretraining experiments train a dense, decoder-only Llama-style transformer consisting of 12 transformer layers, hidden dimension $d = 768$, 12 attention heads, non-affine RMSNorm with regularizer $\epsilon = 10^{-5}$, bias-free linear projections, and tied input/output embeddings over a 50,304-token vocabulary. This configuration totals $\approx 124$M parameters, of which $\approx 46$M belong to non-embedding projection weights. Pretraining is performed on the FineWeb dataset tokenized at sequence length 1024 with a batch size of 64 sequences (65,536 tokens per step) across $T = 189{,}000$ steps, processing $\approx 12.4$B tokens ($\approx 100$ tokens per parameter), corresponding to an overtrained, small-batch regime. Within each seed, compared methods share initialization and minibatch order.

\paragraph{Common Optimizer Configuration.}
Both \textsc{AdamW} and \textsc{AdEMAMix} share baseline retentions $\beta^{(f)} = 0.85$ and $\beta_2 = 0.999$, decoupled weight decay $\lambda = 0.1$, and a fixed global gradient-norm clipping threshold of 0.5. The learning rate begins with a 3,000-step linear warmup from $\eta / 100$. Standard \textsc{Adam} bias corrections $(1 - (\beta^{(f)})^t)^{-1}$ and $(1 - \beta_2^t)^{-1}$ are applied to the fast first-moment and second-moment buffers, while the slow first-moment buffer in \textsc{AdEMAMix} enters uncorrected (raw EMA). In the baseline \textsc{AdEMAMix} configuration, the slow-channel targets are $\alpha = 12$ and $\beta^{(s)} = 0.9999$, warmed over the training horizon as specified below.

\paragraph{Targeted Normalized Matrix Bucket.}
Norm interventions target five hidden-matrix classes in each of the 12 transformer blocks: the fused query-key-value projection (\texttt{qkv}), attention output (\texttt{attn\_out}), and the feed-forward projections (\texttt{mlp\_gate}, \texttt{mlp\_up}, \texttt{mlp\_down}). This defines a set of $12 \times 5 = 60$ weight-decayed matrices. Tied embeddings, unembeddings, RMSNorm affine gains, and any other no-decay parameters are excluded. Every ablation operates per-tensor independently across these 60 matrices, while all non-targeted parameters follow their native optimizer update without modification.

\paragraph{Learning-Rate Schedules.}
With warmup length $t_w = 3000$, horizon $T = 189{,}000$, and peak rate $\eta$, the warmup is
\begin{equation}
\eta_t = \eta \left[ \frac{t}{t_w} + \left(1 - \frac{t}{t_w}\right) \frac{1}{100} \right], \qquad 0 \le t < t_w.
\label{eq:lr-warmup}
\end{equation}
After warmup, the cosine schedule decays from its peak to one percent of the peak at step $T$. The flat schedule holds $\eta_t = \eta$ for $t \ge t_w$.

\paragraph{\textsc{AdEMAMix} Schedules.}
The slow-channel amplitude rises linearly from zero over the full horizon $T_\alpha = T$,
\begin{equation}
\alpha_t = \alpha \frac{t}{T_\alpha}, \qquad 0 \le t \le T_\alpha.
\label{eq:alpha-schedule}
\end{equation}
The slow retention $\beta^{(s)}_t$ is interpolated linearly in half-life. For $0 < \beta < 1$ and $\nu > -1$, set
\begin{equation}
\mathrm{hl}(\beta) := \frac{\ln \frac{1}{2}}{\ln \beta} - 1, \qquad \mathrm{hl}^{-1}(\nu) = \left(\frac{1}{2}\right)^{1/(\nu+1)},
\label{eq:half-life-def}
\end{equation}
so that $\beta^{\mathrm{hl}(\beta)+1} = \frac{1}{2}$; the schedule is
\begin{equation}
\beta^{(s)}_t = \mathrm{hl}^{-1}\left( \left(1 - \frac{t}{T_\beta}\right) \mathrm{hl}(\beta_{\mathrm{start}}) + \frac{t}{T_\beta} \mathrm{hl}(\beta^{(s)}) \right), \qquad 0 \le t \le T_\beta,
\label{eq:beta3-schedule}
\end{equation}
with $T_\beta = T$ and $\beta_{\mathrm{start}} = \beta^{(f)} = 0.85$. At $t = 0$ the slow buffer therefore has the fast buffer's timescale, and its half-life grows linearly to that of $\beta^{(s)} = 0.9999$. Here $t$ denotes the scheduler's post-increment step number. The first optimizer update uses $t = 1$ in the implementation; with zero-based update indices, substitute $t+1$ in the two warmup formulas. The difference is one step of the 189,000-step warmup.

\begin{figure}[!htbp]
\centering
\begin{subfigure}[t]{.48\linewidth}
\includegraphics[width=\linewidth]{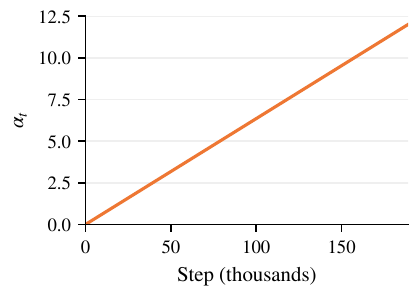}\caption{}
\end{subfigure}\hfill
\begin{subfigure}[t]{.48\linewidth}
\includegraphics[width=\linewidth]{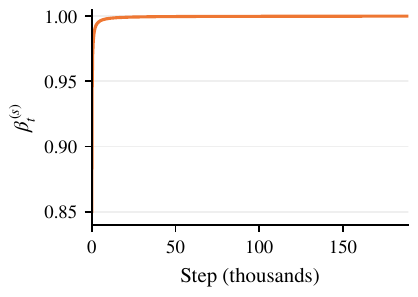}\caption{}
\end{subfigure}
\caption{\textsc{AdEMAMix} schedules for (a) slow amplitude and (b) slow retention. Retention is interpolated in EMA half-life.}
\label{fig:ademamix-schedules}
\end{figure}

\paragraph{Hyperparameter Tuning Sweeps.}
\Cref{tab:llm-search-grids} separates joint searches from conditional sweeps and fixed comparisons. Bold entries mark the selected configurations; each search uses one seed and the final validation checkpoint. Selected settings are held fixed across repeated runs.

\begin{table}[htbp]
\centering
\caption{Search grids for cosine schedules. Bold entries form the selected configuration in each optimizer family. Conditional sweeps fix the parameters indicated in the table; extensions are additional checks.}
\label{tab:llm-search-grids}
\small
\renewcommand{\arraystretch}{1.15}
\begin{tabularx}{\linewidth}{@{}>{\raggedright\arraybackslash}p{0.25\linewidth}>{\raggedright\arraybackslash}X@{}}
\toprule
Search or comparison & Grid and selected setting \\
\midrule
\textsc{AdamW}: joint & $\beta^{(f)}\in\{\mathbf{0.85},0.9,0.95\}$; $\beta_2\in\{0.99,\mathbf{0.999},0.9999\}$; peak $\eta\in\{1,3,5,\mathbf{8}\}\cdot10^{-4}$, at $\lambda=0.1$ \\
\textsc{AdamW}: decay & $\lambda\in\{0.005,0.01,\mathbf{0.1},0.5\}$; selected $(\beta^{(f)},\beta_2,\eta)$ fixed \\
\textsc{AdEMAMix}: slow channel & $\alpha\in\{4,6,8,10,\mathbf{12},15\}$; $\beta^{(s)}\in\{0.99,0.999,\mathbf{0.9999},0.99999\}$; $\beta^{(f)}=0.85$, $\beta_2=0.999$, peak $\eta=8\cdot10^{-4}$ and $\lambda=0.1$ fixed \\
\textsc{AdEMAMix}: decay & $\lambda\in\{0.005,0.01,\mathbf{0.1},0.5\}$; selected optimizer settings fixed \\
\textsc{AdamH}, joint & $\beta^{(f)}\in\{0.8,0.85,\mathbf{0.9}\}$; $\beta_2\in\{0.99,\mathbf{0.999}\}$; $\eta_H\in\{\mathbf{0.005},0.008,0.013,0.02\}$ \\
\textsc{AdEMAMixH}, amplitude & $\alpha\in\{1,\mathbf{2},3,4,6,8,10,12,15\}$ at $\beta^{(s)}=0.9999$ and fixed \textsc{AdamH} settings \\
\textsc{AdamW}/\textsc{AdEMAMix}, higher-rate checks & Peak $\eta\in\{2,5\}\cdot10^{-3}$; selected momentum and decay settings fixed \\
\textsc{AdamH}, low-rate extension & $\eta_H\in\{0.0008,0.0013,0.002,0.003\}$ \\
\bottomrule
\end{tabularx}
\end{table}

\Cref{tab:llm-three-seeds} reports final losses at the selected configurations. The \textsc{AdEMAMixH} amplitude comparison kept $\eta_H=0.005$ fixed.

\begin{table}[htbp]
\centering
\caption{Final-checkpoint loss on the full validation set at update 189,000. SD is in millinats ($10^{-3}$ nats per token).}
\label{tab:llm-three-seeds}
\small
\begin{tabular}{llrr}
\toprule
Schedule & Optimizer & Mean loss & SD (millinats) \\
\midrule
Constant & \textsc{AdamW} & $3.283$ & $0.5$ \\
Constant & \textsc{AdEMAMix} & $3.209$ & $2.2$ \\
Constant & \textsc{AdamW} with \textsc{AdEMAMix} norms & $3.221$ & $1.9$ \\
Constant & \textsc{AdEMAMix} with \textsc{AdamW} norms & $3.273$ & $1.7$ \\
Cosine & \textsc{AdamW} & $3.154$ & $2.1$ \\
Cosine & \textsc{AdEMAMix} & $3.124$ & $2.9$ \\
Cosine & \textsc{AdamH} & $3.140$ & $0.6$ \\
Cosine & \textsc{AdEMAMixH}, $\alpha=2$ & $3.131$ & $0.5$ \\
\bottomrule
\end{tabular}
\end{table}

\begin{figure}[htbp]
\centering
\includegraphics[width=.55\linewidth]{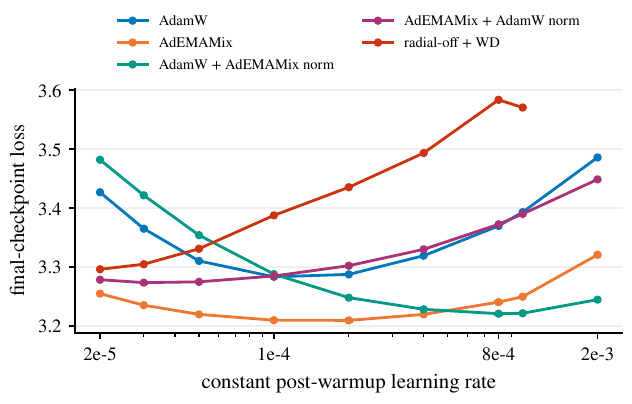}
\caption{Final validation loss across constant learning rates. Points without error bars represent individual runs. \Cref{fig:teaser}b shows the runs at their selected rates. The radial-off run at $2\cdot10^{-3}$ diverged and is omitted.}
\label{fig:lr-ladder}
\end{figure}

The constant-rate comparisons use the common grid in \Cref{tab:llm-constant-grid}, after the 3,000-step warmup, with $\lambda=0.1$, $\beta^{(f)}=0.85$, and $\beta_2=0.999$ fixed. \Cref{fig:lr-ladder} reports the final-checkpoint losses of every configuration on this grid; radial removal diverged at $2\cdot10^{-3}$.

The paired improvements are $74.0\pm1.8$ millinats under constant LR and $30.0\pm1.4$ under cosine (mean $\pm$ sample SD).

\begin{table}[htbp]
\centering
\caption{Common constant-rate search grid and selected rate for each configuration (bold).}
\label{tab:llm-constant-grid}
\small
\begin{tabular}{@{}lc@{}}
\toprule
Common grid & $\{2,3,5\}\cdot10^{-5}\ \cup\ \{1,2,4,8\}\cdot10^{-4}\ \cup\ \{1,2\}\cdot10^{-3}$ \\
\midrule
Configuration & Selected rate \\
\midrule
\textsc{AdamW} & $\boldsymbol{10^{-4}}$ \\
\textsc{AdEMAMix} & $\boldsymbol{2\cdot10^{-4}}$ \\
\textsc{AdamW} with \textsc{AdEMAMix} norms & $\boldsymbol{8\cdot10^{-4}}$ \\
\textsc{AdEMAMix} with \textsc{AdamW} norms & $\boldsymbol{3\cdot10^{-5}}$ \\
Radial-off & $\boldsymbol{2\cdot10^{-5}}$ \\
\bottomrule
\end{tabular}
\end{table}

\subsection{Norm, Rotation, and Instantaneous-Power Measurements}
\label{app:llm-measurements}

\begin{figure}[htbp]
\centering
\includegraphics[width=.50\linewidth]{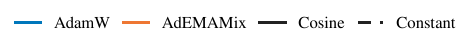}\par\smallskip
\begin{subfigure}[t]{.40\linewidth}
\includegraphics[width=\linewidth]{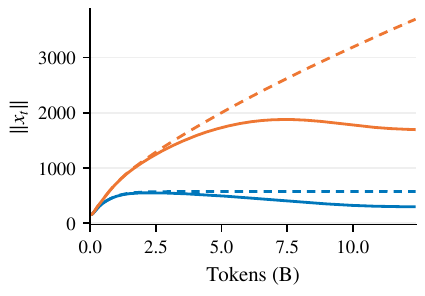}\caption{}
\end{subfigure}\hspace{.05\linewidth}
\begin{subfigure}[t]{.40\linewidth}
\includegraphics[width=\linewidth]{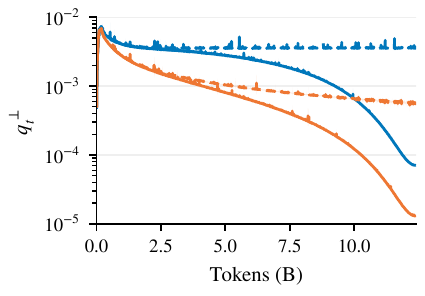}\caption{}
\end{subfigure}
\caption{Aggregate matrix norm (a) and tangential relative update (b) for 124M Llama at peak rate $8\cdot10^{-4}$. Constant-rate \textsc{AdamW} uses one run.}
\label{fig:puzzle}
\end{figure}

The aggregate norm in \Cref{fig:puzzle} is $\norm{\bx_t}=(\sum_m\norm{\bx_{m,t}}_F^2)^{1/2}$. Norms and inner products are summed across the targeted matrices before computing the relative updates of \Cref{app:geometry-rotation}. At matched learning rate, their endpoint ratios obey
\begin{equation}
 \frac{\norm{\bu_{t,\perp}^{\mathrm{AMix}}}}{\norm{\bu_{t,\perp}^{\mathrm{AdamW}}}}
 =\frac{q_t^{\perp,\mathrm{AMix}}}{q_t^{\perp,\mathrm{AdamW}}}
  \frac{\norm{\bx_t^{\mathrm{AMix}}}}{\norm{\bx_t^{\mathrm{AdamW}}}}.
\end{equation}
For individual recorder runs, the cosine comparison gives $0.184\sqrt{32.6}\simeq1.05$: similar tangential displacement lengths accompany the much larger \textsc{AdEMAMix} radius. The constant-rate comparison has a squared-norm ratio of $41.6$ at the last update.

The power shares use separate coordinate-recording runs, one per optimizer and schedule, with the same settings as the norm recordings. Four panels of 128 coordinates per matrix estimate the three terms in \eqref{eq:present-power-split}, aggregated across the 60 matrices over the final $30\%$ of training. The recorded branches use the common live divisor, the fast-buffer bias correction, and the scheduled slow amplitude; the shares already include the factors of $\alpha_t$ in \eqref{eq:present-power-split}.

The same panels also split the survived update displacement $\sum_{t=0}^{T-1}W_{t+1,T}\lr_t\bu_t$ of \eqref{eq:theta_unrolled} at $T=189{,}000$. Substituting $\bu_t=\bu_t^{(f)}+\alpha_t\bu_t^{(s)}$ and expanding its squared norm gives fast, overlap, and slow terms in the same order as \eqref{eq:present-power-split}; each update now carries its weight-decay survival factor. \Cref{tab:survived-shares} compares the two splits.

\begin{table}[htbp]
\centering
\caption{Shares of the three branch terms in present update power (final $30\%$ of training) and in the squared survived update displacement at the endpoint. Coordinate-panel estimates summed over the 60 targeted matrices.}
\label{tab:survived-shares}
\small
\begin{tabular}{llrrr}
\toprule
Schedule & Quantity & Fast & Overlap & Slow\\
\midrule
Constant & Present power & $95.0\%$ & $2.0\%$ & $3.0\%$\\
 & Survived displacement & $1.0\%$ & $14.4\%$ & $84.6\%$\\
\midrule
Cosine & Present power & $93.7\%$ & $2.2\%$ & $4.1\%$\\
 & Survived displacement & $1.5\%$ & $21.3\%$ & $77.2\%$\\
\bottomrule
\end{tabular}
\end{table}

\subsection{Intervention Definitions}
\label{app:llm-interventions}

\paragraph{Post-Step Norm Transfer.}
For each targeted matrix $m$, the post-step transfer is
\begin{equation}
\tilde{\bx}_{m,t+1} = \|\bx_{m,t+1}^{\mathrm{donor}}\| \frac{\bx_{m,t+1}^{\mathrm{opt}}}{\|\bx_{m,t+1}^{\mathrm{opt}}\|}.
\label{eq:norm-transfer}
\end{equation}
Here $\bx_{m,t+1}^{\mathrm{opt}}$ is the recipient's native step, including decay. The donor supplies a dense post-step norm record at the same LR schedule, seed, and data order. Rescaling changes the matrix norm while retaining its direction and optimizer buffers. The logged normalized norm is $\norm{\bx_{m,t}}_F/\norm{\bx_{m,0}}_F$.

\paragraph{Decay-Preserving Radial Removal.}
Writing $\bu_t = u_{t,\parallel} \xhat_t + \bu_{t,\perp}$, the native update rule and the decay-preserving radial-off rule are
\begin{equation}
\bx_{t+1}^{\mathrm{native}} = \delta_t \bx_t - \eta_t (u_{t,\parallel} \xhat_t + \bu_{t,\perp}), \qquad \bx_{t+1}^{\mathrm{off}} = \delta_t \bx_t - \eta_t \bu_{t,\perp}.
\label{eq:radial-rules}
\end{equation}
Their pathwise squared-radius balances satisfy
\begin{align}
\|\bx_{t+1}^{\mathrm{native}}\|^2 &= \delta_t^2 \|\bx_t\|^2 - 2 \delta_t \eta_t \|\bx_t\| u_{t,\parallel} + \eta_t^2 u_{t,\parallel}^2 + \eta_t^2 \|\bu_{t,\perp}\|^2, \label{eq:native-balance} \\
\|\bx_{t+1}^{\mathrm{off}}\|^2 &= \delta_t^2 \|\bx_t\|^2 + \eta_t^2 \|\bu_{t,\perp}\|^2. \label{eq:radial-off-balance}
\end{align}
The wrapper projects the optimizer update before applying decoupled weight decay. The projection leaves the optimizer buffers untouched.

\begin{figure}[!htbp]
\centering
\includegraphics[width=.70\linewidth]{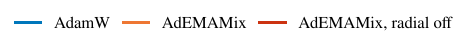}\par\smallskip
\begin{subfigure}[t]{.44\linewidth}
\includegraphics[width=\linewidth]{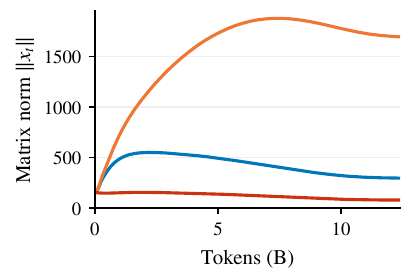}\caption{}
\end{subfigure}\hfill
\begin{subfigure}[t]{.44\linewidth}
\includegraphics[width=\linewidth]{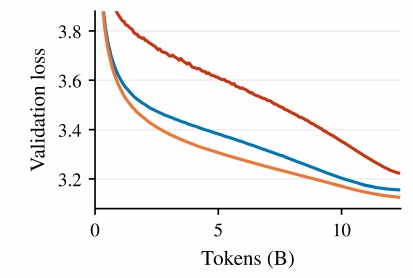}\caption{}
\end{subfigure}
\caption{Radial removal under cosine LR with peak $8\cdot10^{-4}$. (a) Matrix norms. (b) Validation loss.}
\label{fig:radial-removal}
\end{figure}

The projection suppresses norm growth, and the run finishes with higher validation loss than \textsc{AdamW} at the same peak LR (\Cref{fig:radial-removal}).

\FloatBarrier

\paragraph{Slow-Channel Switch.}
We set $\alpha_t=0$ from the cut onward, retaining the parameters and optimizer buffers. The slow buffer continues updating but contributes nothing to the step. Cuts at $50\mathrm{k},100\mathrm{k},150\mathrm{k},170\mathrm{k}$ are tested under cosine peak LR $8\cdot10^{-4}$ and constant LR $10^{-4}$, each with a matched independent \textsc{AdamW} baseline. If post-cut squared-radius growth contributions agreed with the baseline, the squared-radius excess would contract by $\prod_{r=t_{\mathrm{cut}}}^{t-1}(1-\eta_r\lambda)^2$. The measured trajectory also depends on changes in update power and radial alignment.

\paragraph{\textsc{Hyperball} Fixed-Radius Control.}
The \textsc{Hyperball} optimizer \citep{wen2026hyperball} eliminates the radial degree of freedom by constraining each of the 60 targeted matrices to a fixed-radius hypersphere defined by its initial Frobenius norm $r_m^H = \|\bx_{m,0}\|_F$. For each constrained matrix $m$, the update is defined by
\begin{align}
\bz_{m,t+1} &= \bx_{m,t} - \eta_H r_m^H \frac{\bu_{m,t}}{\|\bu_{m,t}\|_F}, \label{eq:hyperball-z} \\
\bx_{m,t+1} &= r_m^H \frac{\bz_{m,t+1}}{\|\bz_{m,t+1}\|_F}. \label{eq:hyperball-x}
\end{align}
These steps normalize the proposal and retract it to the initial Frobenius radius. The step size $\eta_H$ follows the cosine schedule. Constrained matrices receive no weight decay; the remaining parameters use their native \textsc{AdamW} or \textsc{AdEMAMix} update. At the selected $\alpha=2$, the paired improvement over \textsc{AdamH} is $9.13\pm0.04$ millinats. Dividing its mean by the free-radius mean improvement gives $9.13/30.0\simeq0.30$, leaving about one third of the slow-memory benefit.

\begin{figure}[htbp]
\centering
\includegraphics[width=.4\linewidth]{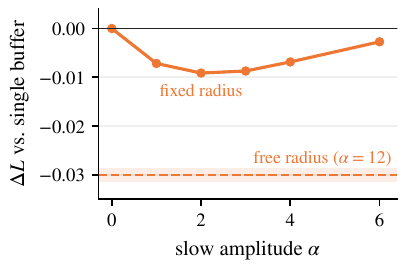}
\caption{Slow-memory amplitude under fixed radius: validation loss relative to \textsc{AdamH}, compared with the free-radius improvement (dashed).}
\label{fig:hyperball-amplitude}
\end{figure}

\subsection{Full Curves for the Main Performance Experiments}
\label{app:llm-full-curves}

\Cref{fig:full-training-curves} shows selected absolute validation-loss trajectories at two constant and two cosine rates, spanning the operating points discussed in \Cref{sec:llm}. \Cref{fig:lr-ladder} shows the full constant-rate ladder.

\begin{figure}[htbp]
\centering
\includegraphics[width=.48\linewidth]{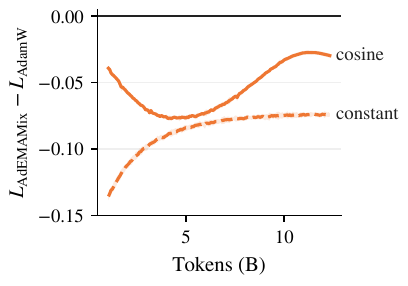}
\caption{\textsc{AdEMAMix} minus \textsc{AdamW} validation loss at each schedule's tuned rate.}
\label{fig:schedule-benefit}
\end{figure}

\begin{figure}[H]
\centering
\includegraphics{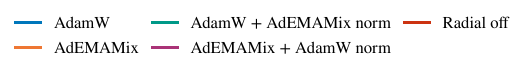}\par\smallskip
\begin{subfigure}[t]{.48\linewidth}
\includegraphics[width=\linewidth]{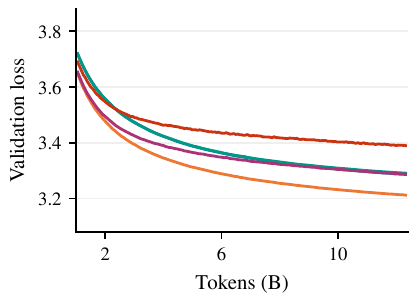}\caption{Constant, $\eta=10^{-4}$}
\end{subfigure}\hfill
\begin{subfigure}[t]{.48\linewidth}
\includegraphics[width=\linewidth]{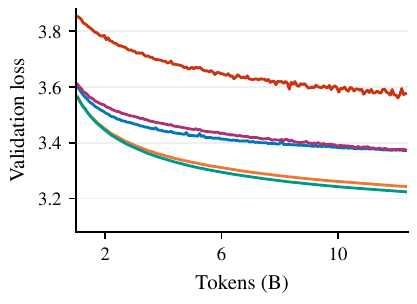}\caption{Constant, $\eta=8\cdot10^{-4}$}
\end{subfigure}
\par\smallskip
\begin{subfigure}[t]{.48\linewidth}
\includegraphics[width=\linewidth]{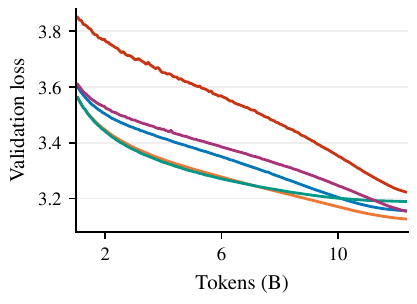}\caption{Cosine, $\eta=8\cdot10^{-4}$}
\end{subfigure}\hfill
\begin{subfigure}[t]{.48\linewidth}
\includegraphics[width=\linewidth]{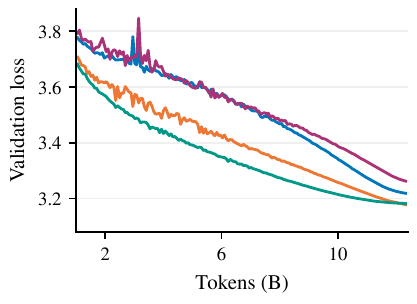}\caption{Cosine, $\eta=5\cdot10^{-3}$}
\end{subfigure}
\caption{Validation loss after the first billion tokens at two constant and two cosine rates. Curves without bands represent individual runs. Panels (a)--(c) also show radial removal.}
\label{fig:full-training-curves}
\end{figure}

\begin{figure}[H]
\centering
\includegraphics[width=\linewidth]{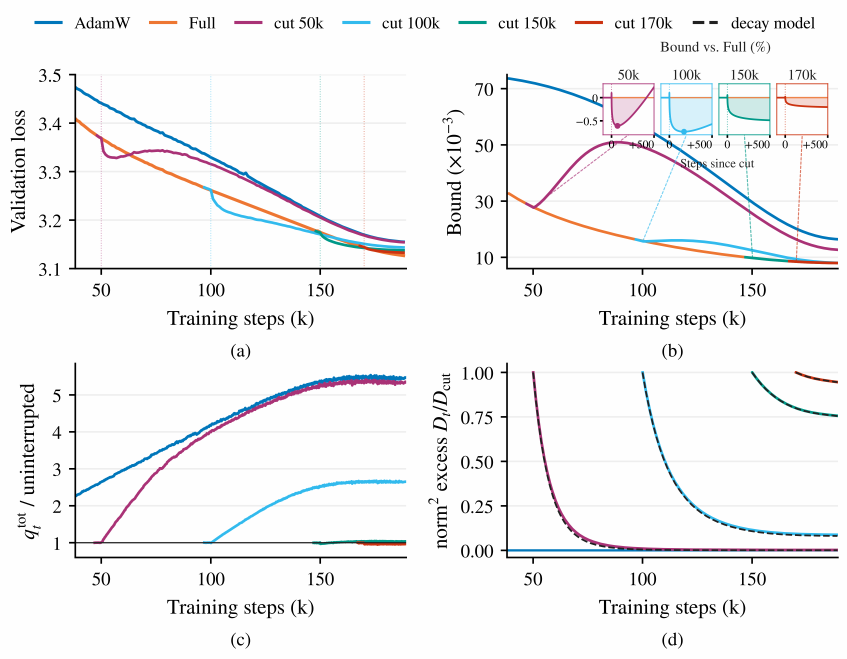}
\caption{Complete cosine-LR comparison for cuts at 50k, 100k, 150k, and 170k updates. Panels are as in \Cref{fig:switch}; each bound inset shows the percentage gap to Full from 100 steps before to 500 steps after its cut, on a shared vertical scale.}
\label{fig:switch-cosine-all}
\end{figure}

\begin{figure}[H]
\centering
\includegraphics[width=.92\linewidth]{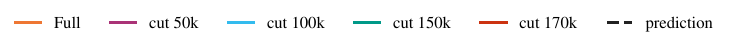}\par\smallskip
\begin{subfigure}[t]{.48\linewidth}
\includegraphics[width=\linewidth]{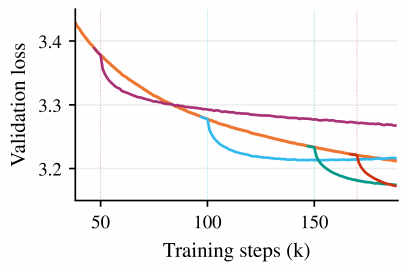}\caption{}
\end{subfigure}\hfill
\begin{subfigure}[t]{.48\linewidth}
\includegraphics[width=\linewidth]{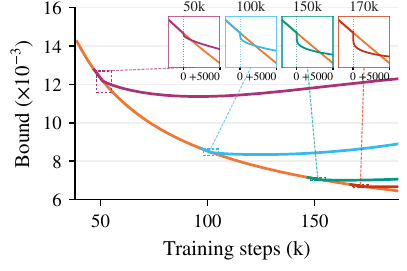}\caption{}
\end{subfigure}
\par\medskip
\begin{subfigure}[t]{.48\linewidth}
\includegraphics[width=\linewidth]{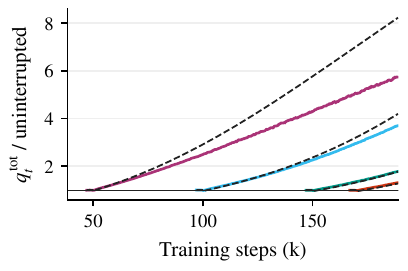}\caption{}
\end{subfigure}\hfill
\begin{subfigure}[t]{.48\linewidth}
\includegraphics[width=\linewidth]{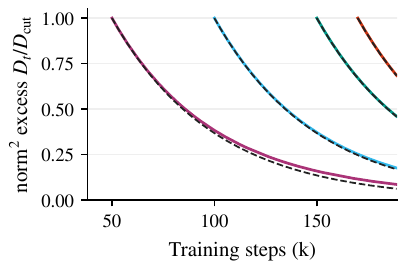}\caption{}
\end{subfigure}
\caption{Removing slow memory at constant LR $10^{-4}$ (individual runs). (a) Validation loss. (b) Normalized bound on modeled rotational coefficients, with zero-mean inputs (\Cref{app:scheduling-switch}). Insets span $2\mathrm{k}$ updates before to $5\mathrm{k}$ after each cut and use separate vertical scales. (c) Total relative update divided by that of uninterrupted \textsc{AdEMAMix}, with the zero-mean model (dashed). (d) Squared-norm excess over \textsc{AdamW}, normalized at the cut, with weight-decay survival (dashed).}
\label{fig:switch-constant}
\end{figure}

\Cref{fig:schedule-benefit} compares the native optimizers under the two LR schedules. Their paired final-loss difference is larger under constant rates; under cosine, the gap narrows as the learning rate decays. With cosine decay, the effect of giving \textsc{AdamW} the \textsc{AdEMAMix} norm trajectory changes sign between the two peak rates shown in \Cref{fig:full-training-curves}c,d.

\begin{figure}[htbp]
\centering
\includegraphics[width=.62\linewidth]{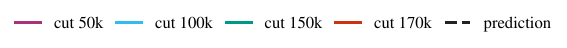}\par\smallskip
\begin{subfigure}[t]{.48\linewidth}\includegraphics[width=\linewidth]{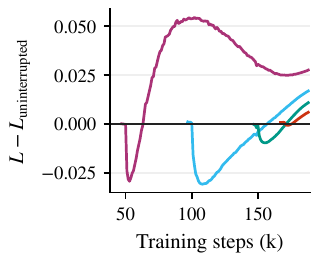}\caption{}\end{subfigure}\hfill
\begin{subfigure}[t]{.48\linewidth}\includegraphics[width=\linewidth]{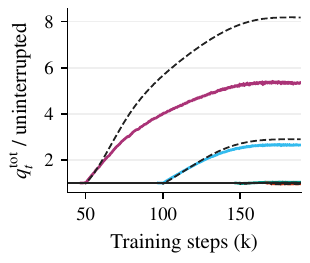}\caption{}\end{subfigure}
\caption{Additional diagnostics for the cosine switches in \Cref{fig:switch}. (a) Loss difference from uninterrupted training. (b) Total relative update divided by that of uninterrupted training, with the finite-time model (dashed).}
\label{fig:switch-cosine}
\end{figure}

\begin{figure}[htbp]
\centering
\includegraphics[width=.62\linewidth]{switch_cosine_legend.pdf}\par\smallskip
\includegraphics[width=.48\linewidth]{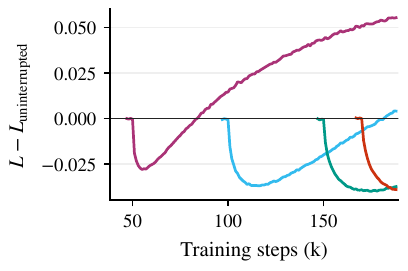}
\caption{Loss difference from uninterrupted training for the constant-LR switch runs of \Cref{fig:switch-constant}.}
\label{fig:switch-difference}
\end{figure}

\FloatBarrier
\subsection{Approximate Scale Invariance of the Matrix Bucket}
\label{app:llm-validity-bucket}
\label{app:llm-validity}

Finite RMSNorm regularization, residual connections, and attention scaling break exact scale invariance. We examine how the loss responds when a matrix is rescaled while its direction is held fixed. Write $\ell_t$ for the average loss on the current minibatch and $\bg_{m,t}$ for its raw gradient with respect to matrix $\bx_{m,t}$. With the other parameters fixed, the chain rule gives
\[
 \left.\frac{\mathrm d}{\mathrm d\kappa}\ell_t(\kappa\bx_{m,t})\right|_{\kappa=1}
 =\ip{\bg_{m,t}}{\bx_{m,t}}.
\]
A $1\%$ increase in norm therefore predicts a first-order loss change of one hundredth of this inner product. Dividing its absolute value by $\norm{\bg_{m,t}}_F\norm{\bx_{m,t}}_F$ gives the fraction of the gradient norm in the radial direction.

\begin{figure}[!htbp]
\centering
\includegraphics{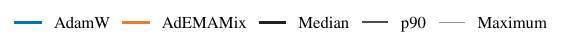}\par\smallskip
\begin{subfigure}[t]{.48\linewidth}
\includegraphics[width=\linewidth]{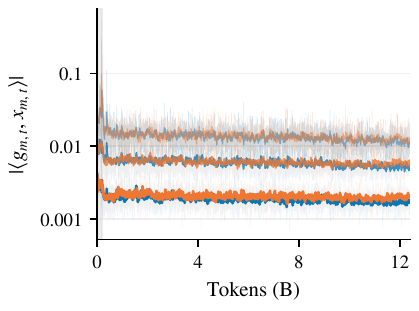}\caption{}
\end{subfigure}\hfill
\begin{subfigure}[t]{.48\linewidth}
\includegraphics[width=\linewidth]{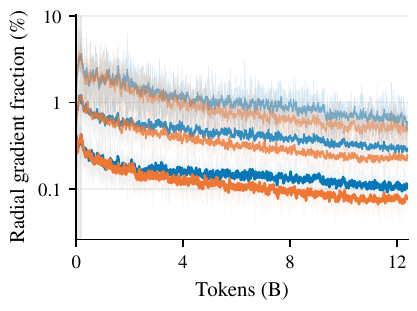}\caption{}
\end{subfigure}
\par\smallskip
\begin{subfigure}[t]{\linewidth}
\centering
\includegraphics[width=.48\linewidth]{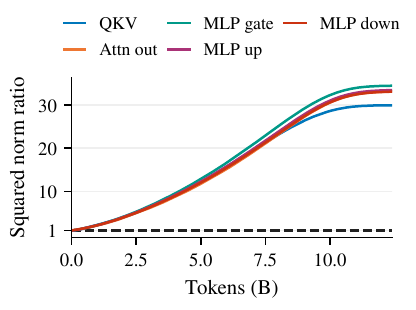}\caption{}
\end{subfigure}
\caption{Local scale sensitivity and norm response. (a) Absolute sensitivity to relative rescaling, $|\ip{\bg_{m,t}}{\bx_{m,t}}|$. (b) Radial fraction of the raw gradient. Panels (a,b) show the median, 90th percentile, and maximum across 60 matrices, sampled every 50 updates in one cosine run per optimizer. An EMA with a 250-update half-life smooths the curves; faint traces show the raw observations, and grey shading marks warmup. (c) \textsc{AdEMAMix}/\textsc{AdamW} squared-norm ratios by matrix class.}
\label{fig:llm-scale-control}
\end{figure}

We measure both quantities before clipping in one additional cosine run per optimizer, following \Cref{app:llm-protocol}. Figure~\ref{fig:llm-scale-control}(a,b) shows small local loss sensitivity in both optimizers after warmup. This sensitivity stays roughly stable while the radial gradient fraction decreases. The largest sensitivity spikes occur near the end of warmup. These observations support approximate scale invariance locally along training. The larger norms under \textsc{AdEMAMix} occur across all matrix classes (panel c).

At updates $3\mathrm{k},50\mathrm{k},100\mathrm{k},150\mathrm{k},170\mathrm{k},189\mathrm{k}$ in each of the four new transplant runs, we evaluated the same 32,768 held-out tokens immediately before and after rescaling. The individual matrix-norm corrections ranged from $-0.04\%$ to $+0.07\%$. Across the 24 comparisons, the largest absolute change in cross-entropy was $2.5\cdot10^{-4}$ and the largest $\mathrm{KL}(p_{\mathrm{before}}\|p_{\mathrm{after}})$ was $3.5\cdot10^{-4}$, both in nats per token.

\clearpage
\subsection{Checking the Preconditioner Approximation}
\label{app:llm-preconditioner}

\textsc{AdEMAMix} divides its momentum buffers by the current $\sqrt{\bv_{t+1}}+\eps$. The model in \eqref{eq:memory-model-main} instead divides each gradient when it is computed, then averages these normalized gradients in the buffers. Each gradient therefore keeps its original divisor when it is used again later. We check how much this changes the accumulated updates.

We reconstruct $\widehat{\bu}_t$ from the recorded normalized gradients, using the same momentum schedules and bias correction as in training. Over each block of 16,384 updates, we measure the percentage difference in squared displacement,
\[
 100\left(
 \frac{\left\|\sum_{t=t_0}^{T-1}W_{t+1,T}\lr_t\widehat{\bu}_t\right\|^2}
      {\left\|\sum_{t=t_0}^{T-1}W_{t+1,T}\lr_t\bu_t\right\|^2}
 -1\right).
\]
Here $W_{t+1,T}$ is the fraction of the displacement at step $t$ that remains after weight decay up to $T$. The buffers keep their full training history; the sums cover only the block.

For each matrix, we average the squared accumulated displacements of four sets of 128 recorded coordinates and multiply by the number of weights in that matrix. We sum these estimates over the 60 targeted matrices before calculating the percentage difference.

\begin{figure}[htbp]
\centering
\includegraphics[width=.48\linewidth]{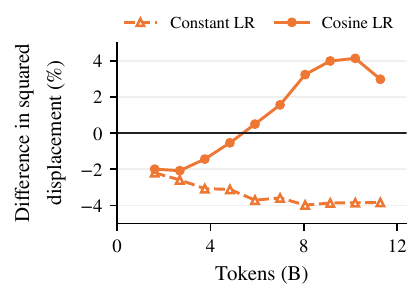}
\caption{Percentage difference in squared accumulated displacement between the approximation and actual updates. Each point summarizes one block, using one \textsc{AdEMAMix} run per LR schedule. Zero means agreement; positive values mean the approximation gives a larger squared displacement.}
\label{fig:llm-preconditioner}
\end{figure}

\Cref{fig:llm-preconditioner} shows the difference over the ten recorded blocks under constant and cosine LR, with the settings of \Cref{app:llm-protocol}. It stays within about $4\%$ across these blocks.

\clearpage

\end{document}